%% file: acl2021.tex
\documentclass[11pt,a4paper]{article}
\usepackage[hyperref]{acl2021}
\usepackage{times}
\usepackage{latexsym}

\input{math_commands.tex}

\usepackage{hyperref}
\usepackage{url}
\usepackage{graphicx,subcaption}
\usepackage{booktabs}
\usepackage{multirow}

\title{Reservoir Transformers}

\author{Sheng Shen$^\dagger$, Alexei Baevski$^\ddagger$, Ari S. Morcos$^\ddagger$, Kurt Keutzer$^\dagger$,\\
\bf{Michael Auli$^\ddagger$, Douwe Kiela$^\ddagger$}\\
 $^\dagger$UC Berkeley; $^\ddagger$Facebook AI Research\\
\texttt{sheng.s@berkeley.edu, dkiela@fb.com}
}

\def\aclpaperid{2831}
\aclfinalcopy

\begin{document}

\maketitle

\begin{abstract}
We demonstrate that transformers obtain impressive performance even when some of the layers are randomly initialized and never updated. Inspired by old and well-established ideas in machine learning, we explore a variety of non-linear ``reservoir'' layers interspersed with regular transformer layers, and show improvements in wall-clock compute time until convergence, as well as overall performance, on various machine translation and (masked) language modelling tasks.
\end{abstract}

\section{Introduction}

Transformers \citep{Vaswani:2017attention} have dominated natural language processing (NLP) in recent years, from large scale machine translation~\citep{Ott:2018scaling} to pre-trained (masked) language modeling ~\citep{Devlin:2018bert,Radford:2019gpt2}, and are becoming more popular in other fields as well, from reinforcement learning~\citep{vinyals2019grandmaster} to speech recognition~\citep{Baevski2019vqwav2vec} and computer vision~\citep{Carion:2020detr}. Their success is enabled in part by ever increasing
computational demands, which has naturally led to an increased interest in improving their efficiency. Scalability gains in transformers could facilitate bigger, deeper networks with longer contexts~\citep{Kitaev:2020reformer,Wang:2020linformer,Beltagy:2020longformer,Kaplan2020scaling,Tay:2020transformersurvey}. Conversely, improved efficiency could reduce environmental costs \citep{Strubell:2019energy} and hopefully help democratize the technology.

In this work, we explore a simple question: if some layers of the transformer are kept frozen---i.e., never updated after random initialization---can we match the performance of fully learned transformers, while being more efficient? Surprisingly, the answer is resoundingly yes; and what is more, we find that freezing layers may actually \emph{improve} performance.

Beyond desirable efficiency gains, random layers are interesting for several additional reasons. Fixed randomly initialized networks \citep{Gallicchio:2020survey} converge to Gaussian processes in the limit of infinite width \citep{Daniely:2016randominit}, have intriguing interpretations in metric learning~\citep{Rosenfeld:2019intriguing,Giryes:2016metriclearning}, and have been shown to provide excellent~``priors'' either for subsequent learning~\citep{Ulyanov:2018deepimageprior} or pruning \citep{Frankle:2018lottery}. Fixed layers allow for efficient low-cost hardware implementations \citep{Schrauwen:2007lsmhardware} and can be characterized using only a random number generator and its seed. This could facilitate distributed training and enables highly efficient deployment to edge devices, since it only requires transmission of a single number. The strong performance of networks with fixed layers also sheds new light on the inner workings of BERT~\citep{Devlin:2018bert}, and layer-wise interpretations of such models \citep{Rogers2020:primer,Tenney:2019bert}. It appears that~``not all layers are created equal''~\citep{Zhang:2019alllayersequal} is true to such an extent that some layers can simply remain random and fixed. 

Random projections have a long history in machine learning. By Cover's theorem \citep{Cover:1965theorem}, any high-dimensional non-linear transformation is more likely to be linearly separable than its lower-or-equal-dimensional input space. By Johnson-Lindenstrauss~\citep{Johnson:1984extensions}, random projections distort Euclidean distances very little under mild assumptions, which is useful e.g. for dimensionality reduction and random indexing \citep{Sahlgren:2005randomindexing}. Fixed random layers in neural networks pre-date deep learning by far~\citep{Gamba:1961papa,Baum:1988jc}. Indeed, random kernel methods have long been influential in machine learning \citep{Rahimi:2008random,Rahimi:2009kitchen}.

One way to think of such layers is as ``reservoirs'' \citep{Lukovsevivcius:2009reservoir}, where a highly non-linear high-dimensional black box representation is provided to a lightweight ``readout'' network, as in echo state networks \citep{Jaeger:2003echostate} and liquid state machines \citep{Maass:2002lsm}. The benefit of such an approach is that the reservoir has fixed parameters and is computationally efficient, as it can be pre-computed and does not (necessarily) require backpropagation.

In NLP, \newcite{Wieting:2019notraining} showed that random sentence encoders present a strong baseline for text classification, with subsequent work showing applications in a variety of tasks from summarization to machine translation~\citep{Enguehard:2019neurallanguagepriors,Garg:2020echostatenmt,Pilault:2020impressive}. To our knowledge, this work is the first to examine this phenomenon in transformers, and the first to recursively alternate reservoirs with subsequent transformer layers acting as readout functions. We introduce ``reservoir transformers'', wherein fixed random reservoir layers are interspersed with regular updateable transformer layers. The goal of this work is to put our understanding of transformer models on a more solid footing by providing empirical evidence of their capabilities even when some of their parameters are fixed. Our contributions are as follows:

\begin{itemize}
    \item We introduce a \emph{area under the convergence curve} metric for measuring performance-efficiency trade-offs, and show that replacing regular transformer layers with reservoir layers leads to improvements.
    \item We show that the addition of reservoir layers leads to improved test set generalization on a variety of tasks in a variety of settings.
    \item We show that pre-trained masked language modelling architectures like BERT and RoBERTa \citep{Liu:2019roberta} can benefit from having some of their layers frozen, both during pre-training as well as when fine-tuning on downstream tasks.
    \item We experiment with different types of reservoir layers, including convolutional and recurrent neural network-based ones.
    \item We show empirical evidence that the backward pass can be skipped \emph{in its entirety} by approximating upstream gradients using an approach we call \emph{backskipping}, which can reduce the training compute further without sacrificing performance.
\end{itemize}

\section{Approach}

This paper is based on a very simple idea. 
Neural networks are trained via backpropagation, which involves consecutive steps of matrix addition and multiplication, i.e.,

\begin{equation*}
    \theta_{t+1} \leftarrow \theta_{t}  - \eta \frac{\partial J}{\partial \theta_t}; \frac{\partial J}{\partial \theta_t} = \frac{\partial J}{\partial L_n}\frac{\partial L_n}{\partial L_{n-1}} \cdots 
    \frac{\partial L_0}{\partial x}
\end{equation*}

for some objective $J$, parameterization $\theta$ and learning rate $\eta$, with the gradient computed via the chain rule,
%
where $L_i$ is the $i$-th layer of the neural network and $x$ is the input. Let $L = \text{Transformer}(X)$ be a single layer in a Transformer network \citep{Vaswani:2017attention}, i.e.,

\begin{align*}
    \begin{split}
    H = \text{MultiHeadSelfAttn}(\text{LayerNorm}(X)) + X\\
    L = \text{FFN}(\text{LayerNorm}(H)) + H
    \end{split}
\end{align*}

Now, during every ``backward pass'', we compute the Jacobian for parameters $\theta^L$ at layer $L$, which are used to update the parameters of $L$, $\theta^L_t$, as well as to compute the next layer's Jacobian, thus back-propagating the gradients. In this work however, for some of the layers, we still backpropagate through them to compute gradients for earlier layers, \emph{but we never apply the parameter update}. As a result, these layers stay fixed at their initialization, saving computational resources.

\subsection{Background}

Naturally, never updating some of the parameters is computationally more efficient, as some matrix addition operations can be skipped in the backward pass, but why is this not detrimental to the performance of the network?

In the early days of neural networks, the bottom layers were often kept fixed as ``associators''~\citep{Block:1962perceptron}, or what \cite{Minsky:2017book} called the Gamba perceptron \citep{Gamba:1961papa,Borsellino:1961papa}. Fixed random networks~\citep{Baum:1988jc,Schmidt:1992pr,Pao:1994nc} have been explored from many angles, including as ``random kitchen sink'' kernel machines~\citep{Rahimi:2008random,Rahimi:2009kitchen}, ``extreme learning machines''~\citep{Huang:2006nc} and reservoir computing~\citep{Jaeger:2003echostate,Maass:2002lsm,Lukovsevivcius:2009reservoir}. In reservoir computing, input data are represented through fixed random  high-dimensional non-linear representations, called ``reservoirs'', which are followed by a regular~(often but not necessarily linear)~``readout'' network to make the final classification decision.

The theoretical justification for these approaches lies in two well-known results in machine learning: Cover's theorem \citep{Cover:1965theorem} on the separability of patterns states that high-dimensional non-linear transformations are more likely to be linearly separable; and the Johnson-Lindenstrauss lemma \citep{Johnson:1984extensions} shows that (most) random projections distort Euclidean distances very little.

Practically, random layers can be seen as a cheap way to increase network depth. There are interesting advantages to this approach. Fixed layers are known to have particularly low-cost hardware requirements and can be easily implemented on high-bandwidth FPGAs with low power consumption \citep{Hadaeghi:2017hardware,Tanaka:2019review}, or on optical devices \citep{Hicke:2013semiconductor}. This might yield interesting possibilities for training in a distributed fashion across multiple devices, as well as for neuromorphic hardware~\citep{Neftci:2017randombackprop}. This approach also facilitates lower-latency deployment of neural networks to edge devices, since weights can be shared simply by sending the seed number, assuming the random number generator is known on both ends.

\subsection{Reservoir Transformers}

This work explores inserting random non-linear transformations, or what we call reservoir layers, into transformer networks. Specifically, we experiment with a variety of reservoir layers:

\begin{itemize}
\item Transformer Reservoir: The standard transformer layer as described above, but with all parameters fixed after initialization, including the self-attention module.
\item FFN Reservoir: A transformer-style fixed feed-forward layer without any self-attention, i.e., \text{FFN}($\text{LayerNorm}(\text{Previous\_layer}))+\text{Previous\_layer}$.
\item BiGRU Reservoir: A fixed bidirectional Gated Recurrent Unit \citep{Cho:2014gru} layer, which is closer in spirit to previous work on reservoir computing, most of which builds on recurrent neural network architectures.
\item CNN Reservoir: A fixed Convolutional Neural Network \citep{Lecun:1998cnn} layer, specifically light dynamical convolution layers \citep{Wu2019pay}, which are known to be competitive with transformers in sequence-to-sequence tasks.
\end{itemize}

We find that all these approaches work well, to a certain extent. For clarity, we focus primarily on the first two reservoir layers, but include a broader comparison in Appendix \ref{appendix:hybrids}.

In each case, contrary to traditional reservoir computing, our reservoir layers are interspersed throughout a regular transformer network, or what we call a reservoir transformer. Since random projections are not learned and might introduce noise, subsequent normal transformer ``readout'' layers might be able to benefit from additional depth while allowing us to recover from any adverse effects of randomness. For example, previous work has shown that ResNets, with all of their parameters fixed except for the scale and shift parameters of batch normalization, can still achieve high performance, simply by scaling and shifting random features \citep{Frankle:2020batchnorm}. Adding some form of noise to the parameters is also known to help convergence and generalization~\citep{Jim1995effects,jim1996analysis,Gulcehre:2016noisy,Noh:2017regularizing}.

\section{Evaluation}

We evaluate the proposed approach on a variety of well-known tasks in natural language processing, namely: machine translation, language modelling and masked language model pre-training.

We set out to do this work with the main objective of examining any potential efficiency gains, i.e. the relationship between compute time and task performance. This is closely related to efforts in Green AI, which are concerned with the trade-offs between compute, data, and performance~\citep{Schwartz:2019greenai}. We propose to measure this trade-off via the \emph{area under the convergence curve} (AUCC): similarly to how the area under the receiver operating characteristic~\citep[AUC-ROC]{Bradley:1997roc} measures a classifier's performance independent of the classification threshold, AUCC measures a model's performance independent of the specific compute budget. Specifically, AUCC is computed as follows:

\begin{equation}
    \int_{t=0}^{\hat{T}} \sum_{x,y\in\mathcal{D}} g_t(f(x),y)
\end{equation}

where $f$ is the network and $g$ is the evaluation metric, measured until convergence time $\hat{T}$, which is the maximum convergence time of all models included in the comparison. Note that time here is wall-clock time, not iterations. By convergence, we mean that validation performance has stopped improving, and hence the convergence curve whose area we measure plots the desired metric over time. Runs are averaged over multiple seeds and reported with standard deviation. We normalize raw AUCC scores by their maximum to ensure a more interpretable $[0-1]$ range. 

One potential downside of this approach is that the AUCC metric could lead to higher scores for a model that converges quickly but to ultimately worse performance, if measured in a small window. This can be solved by making sure that $\hat{T}$ is set sufficiently high. We include the raw validation curves in the appendix to demonstrate that the chosen window sizes are sufficient and the results are not a influenced by this limitation. In addition, we report the number of trainable parameters and the wall-clock training time until maximum performance (plus 95\% and 99\% convergence results in the appendix). Finally, we show test set generalization in each experiment. Overall, this gives us a wide set of axes along which to examine models.

\begin{figure*}[t]
    \centering
    \begin{subfigure}{.33\textwidth}
        \centering
        \includegraphics[width=\textwidth]{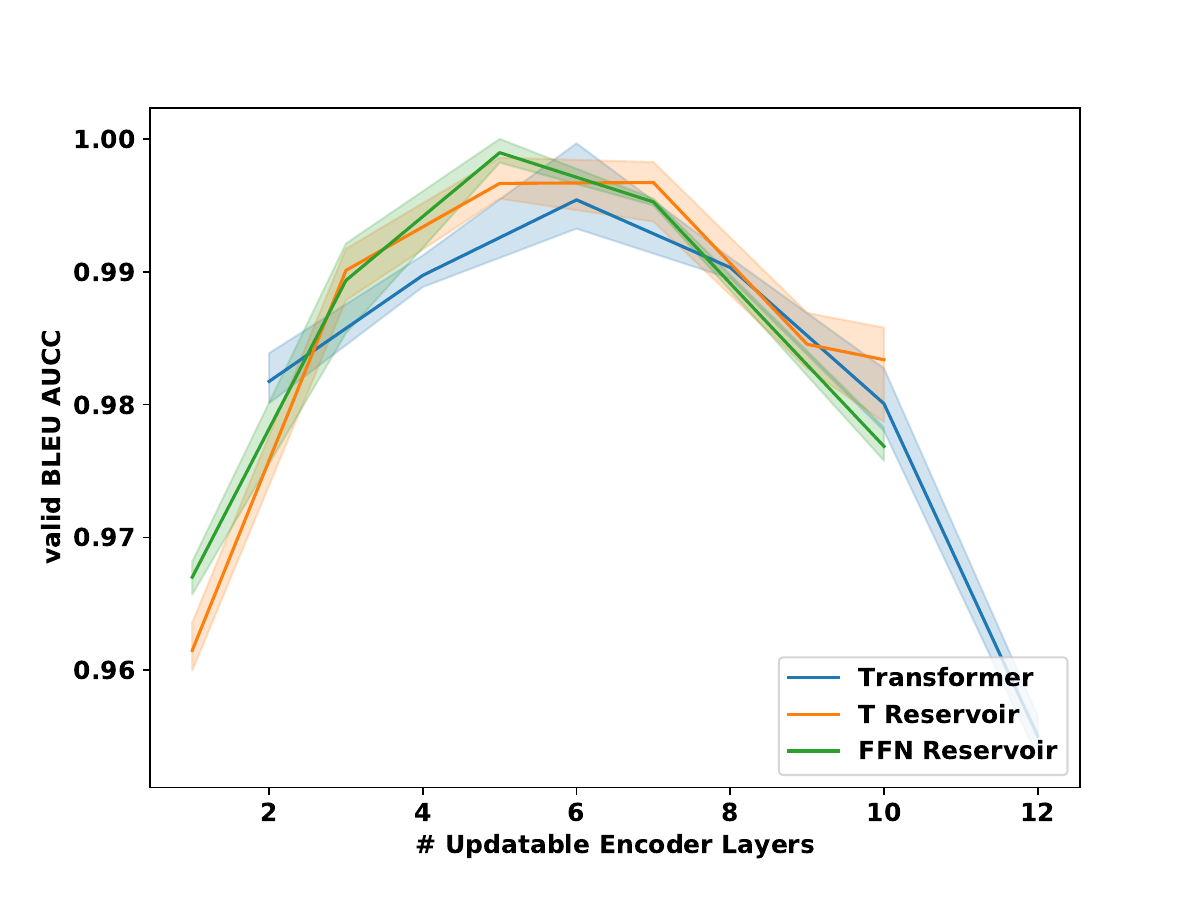}
    \end{subfigure}%
    \begin{subfigure}{.33\textwidth}%
        \centering
        \includegraphics[width=\textwidth]{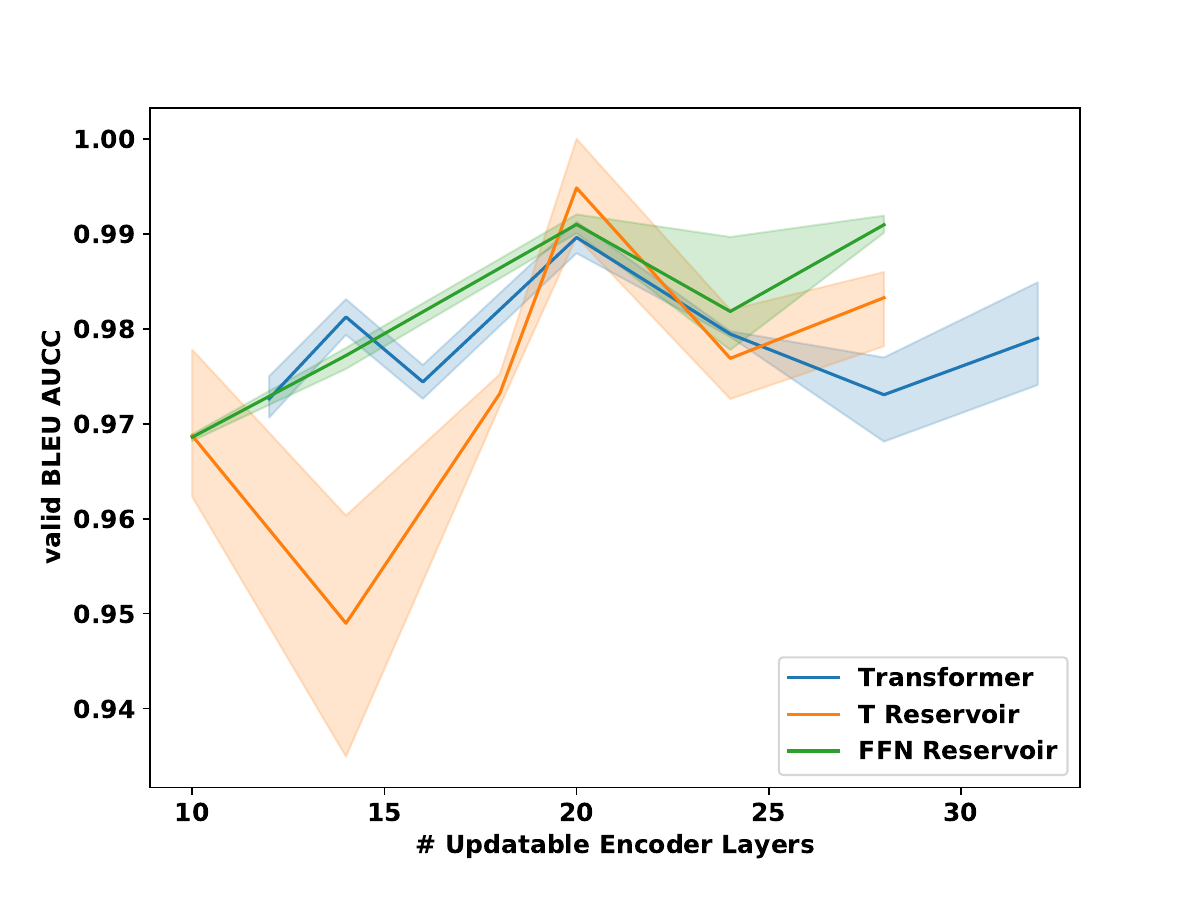}
    \end{subfigure}
    \begin{subfigure}{.33\textwidth}
        \centering
        \includegraphics[width=\textwidth]{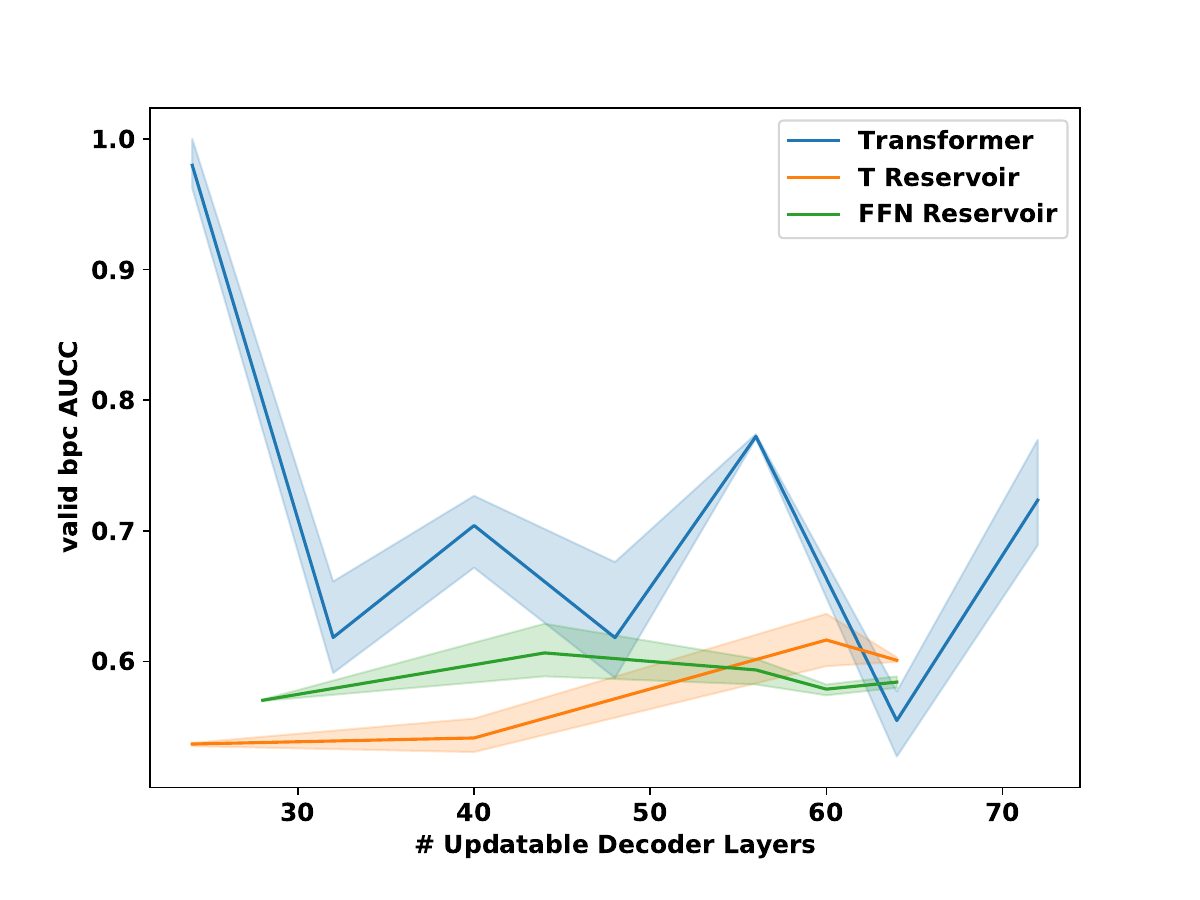}
    \end{subfigure}\\
    \begin{subfigure}{.33\textwidth}
        \centering
        \includegraphics[width=\textwidth]{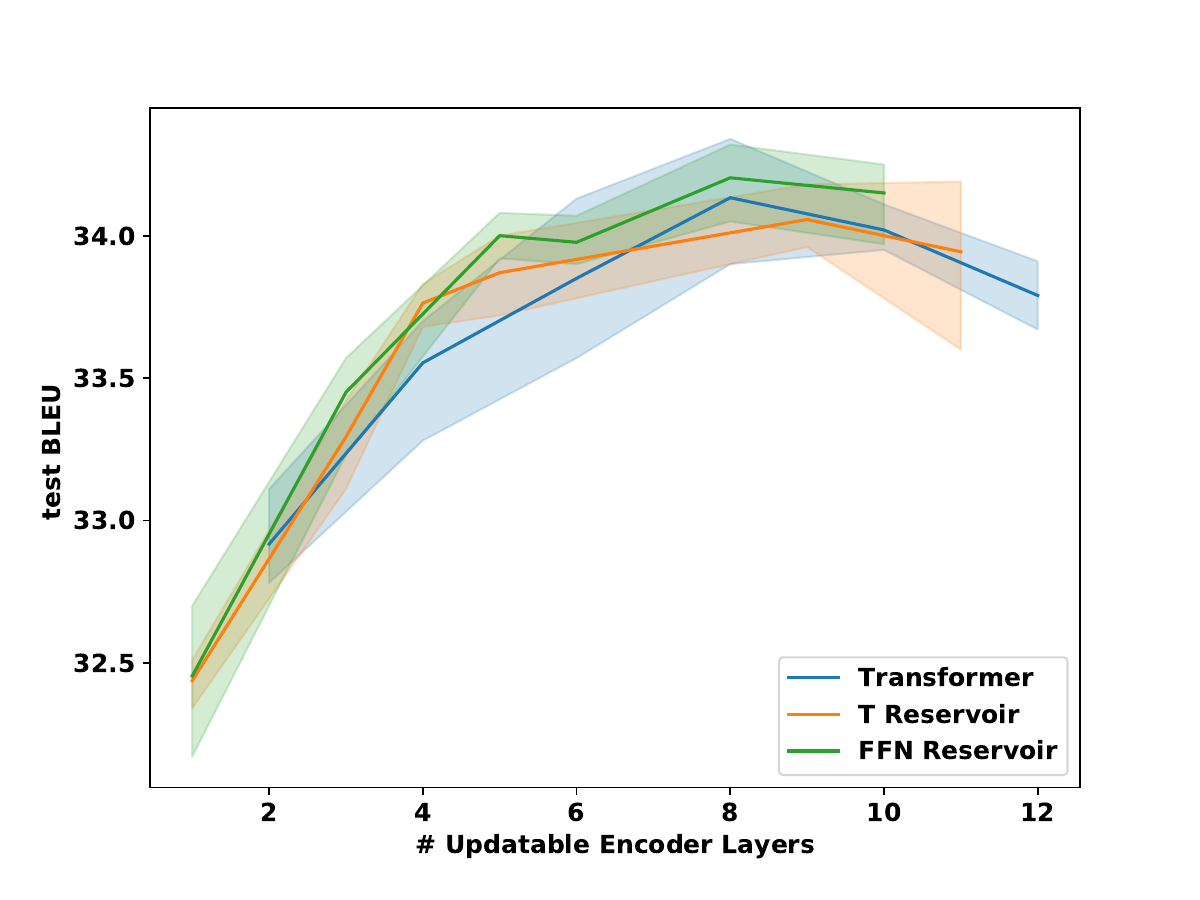}
    \end{subfigure}%
    \begin{subfigure}{.33\textwidth}
        \centering
        \includegraphics[width=\textwidth]{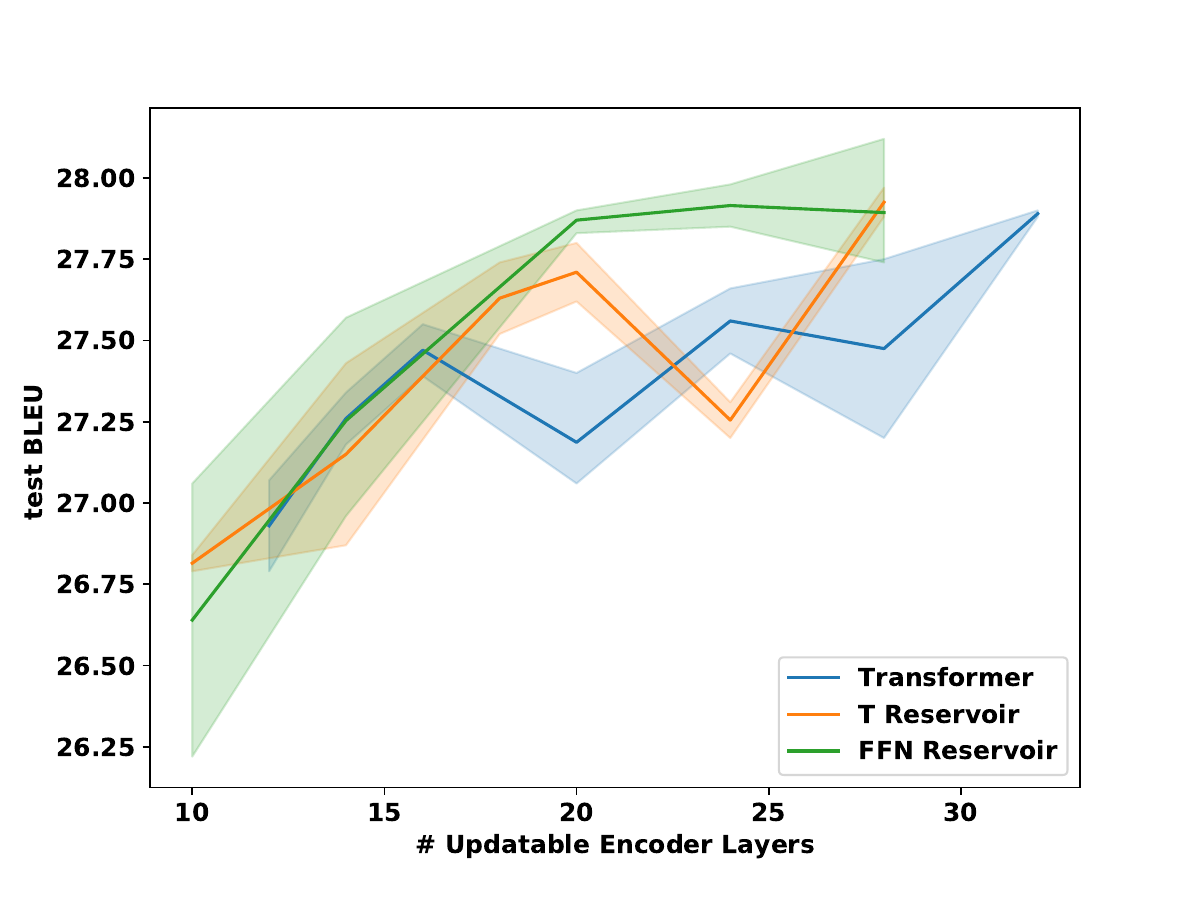}
    \end{subfigure}
    \begin{subfigure}{.33\textwidth}
        \centering
        \includegraphics[width=\textwidth]{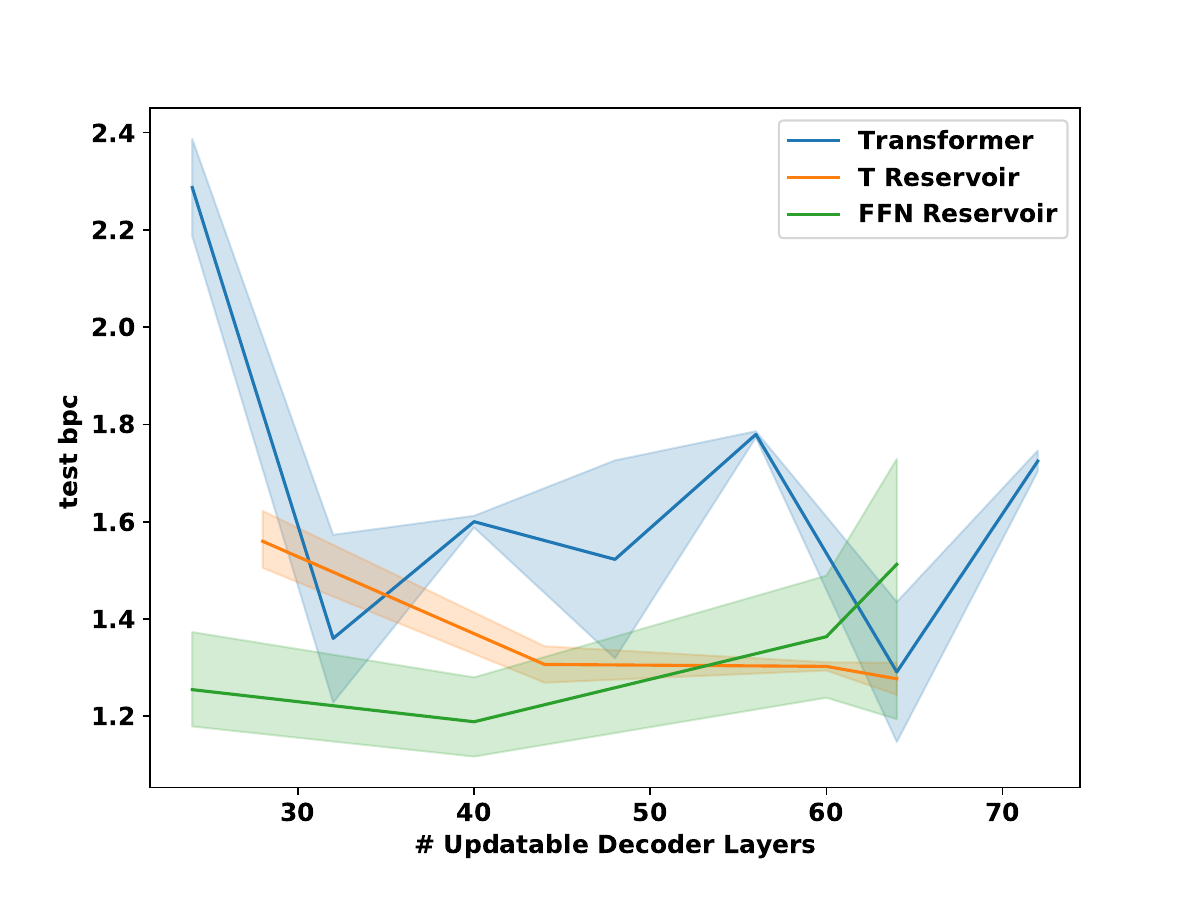}
    \end{subfigure}\\
    \caption{Validation (top) and test (bottom) results for IWSLT (left), WMT (middle) and enwiki8 language modelling (right). IWSLT and WMT are BLEU (high is good); enwiki8 is BPC (low is good). Comparison of regular transformer (blue) and reservoir transformer with FFN (green) or Transformer reservoir (orange) layers added. 
    }
    \label{fig:iwslt}
\end{figure*}


\begin{table*}[t]
    \centering
    \small
    \resizebox{\textwidth}{!}{
    \begin{tabular}{lccccccc}
    \toprule
        \multirow{2}{*}\textbf{Model} & \textbf{\# Layers} & \textbf{Frozen} & \textbf{Max BLEU} & \textbf{Train time}  & \textbf{Ratio} & \textbf{\# Params}  & \textbf{Train Time each}  \\
        &&&&\textbf{until max }(in hours)& & \textbf{Trainable} (Total) &\textbf{epoch} (in seconds)
 \\\midrule
        \multirow{4}{*}{Transformer} 
        & 6 & 0 & 34.52 $\pm$ 0.07 & 2.548 $\pm$ 0.06 & 1 & 26.8M & 122.73 $\pm$ 1.16\\
        & 8 & 0 & 34.59 $\pm$ 0.11 & 2.557 $\pm$ 0.05 & 1 & 31.1M & 142.28 $\pm$ 1.87\\
        & 10 & 0 & 34.56 $\pm$ 0.05 & 3.173 $\pm$ 0.04 & 1 & 35.3M & 161.66 $\pm$ 1.54\\
        & 12 & 0 & 34.29 $\pm$ 0.12 & 3.521 $\pm$ 0.09 & 1 & 39.5M & 172.45 $\pm$ 1.98\\\midrule
        \multirow{4}{*}{T Reservoir} 
        & 6 & 2 & 34.37 $\pm$ 0.12 & 2.422 $\pm$ 0.03 & 0.95 & 22.6M (26.8M) & 120.59 $\pm$ 1.32\\
        & 8 & 2 & 34.80 $\pm$ 0.07 & 2.450 $\pm$ 0.06 & 0.96 & 26.8M (31.1M) & 134.49 $\pm$ 1.76\\
        & 10 & 2 & 34.70 $\pm$ 0.03 & 2.831 $\pm$ 0.05 & 0.89 & 31.1M (35.3M) & 144.42 $\pm$ 1.98\\
        & 12 & 2 & 34.78 $\pm$ 0.04 & 3.476 $\pm$ 0.04 & 0.98 & 35.3M (39.5M) & 159.43 $\pm$ 1.67\\\midrule
        \multirow{4}{*}{FFN Reservoir} 
        & 6 & 2 & 34.43 $\pm$ 0.15 & 2.120 $\pm$ 0.04 & 0.83 & 22.6M (25.8M) & 107.71 $\pm$ 1.73\\
        & 8 & 2 & 34.56 $\pm$ 0.16 & 2.203  $\pm$ 0.06 & 0.86 & 26.8M (29.1M) & 120.07 $\pm$ 1.65\\
        & 10 & 2 & 34.66 $\pm$ 0.02 & 2.493 $\pm$ 0.05 & 0.79 & 31.1M (33.3M) & 130.11 $\pm$ 1.43\\
        & 12 & 2 & 34.76 $\pm$ 0.03 & 3.241  $\pm$ 0.04 & 0.92 & 35.3M (37.5M) & 156.32 $\pm$ 1.87\\\midrule
        \multirow{4}{*}{LayerDrop} 
        & 6 & 2 & 34.59  $\pm$ 0.15 & 2.364 $\pm$ 0.08 & 0.92 & 22.6M (26.8M) & 119.30 $\pm$ 1.36\\
        & 8 & 2 & 34.58 $\pm$ 0.16 & 2.554 $\pm$ 0.05 & 0.99 & 26.8M (31.1M) & 138.62 $\pm$ 1.44\\
        & 10 & 2 & 34.57 $\pm$ 0.07 & 3.404 $\pm$ 0.06 & 1.07 & 31.1M (35.3M) & 140.88 $\pm$ 1.62\\
        & 12 & 2 & 33.65 $\pm$ 0.24 & 3.251 $\pm$ 0.04 & 0.92 & 35.3M (39.5M) & 160.85 $\pm$ 1.49\\
    \bottomrule
    \end{tabular}}
    \caption{Wall-clock time (averaged over multiple runs) saved for IWSLT for different model types and encoder depths. Max BLEU is for validation. Number of layers is for encoder, decoder depth is kept fixed at 2. The ratio is computed compared to the corresponding number of layers in the regular transformer case.}
    \label{tab:iwslt-time}
\end{table*}

\subsection{Experimental Settings}
\label{sec:experimental_details}
We evaluate on IWSLT de-en \citep{Cettolo:2015ReportOT} and WMT en-de \citep{bojar:2014-findings} for machine translation; enwiki8 \citep{LLC:2009long} for language modelling; and experiment with RoBERTa \citep{Liu:2019roberta} in our pretraining experiments. For IWSLT, we follow the pre-processing steps in \newcite{edunov:2018classical}. The train/val/test split is 129k/10k/6.8k sentences. For WMT, we follow pre-process as in \newcite{Ott:2018scaling}, with 4.5M/16.5k/3k sentences in train/val/test. For enwiki8, we follow the pre-processing steps in \newcite{dai:2019-transformer}. The train/val/test split is 1M/54k/56k sentences. For RoBERTa pretraining, we follow the pre-processing steps in \newcite{Liu:2019roberta}.

We use 8 Volta V100 GPUs for WMT and enwik8, 32 V100 GPUs for RoBERTa and a single V100 for IWSLT. The hyperparameters for IWSLT14 and WMT16 were set  to the best-performing values from \newcite{Ott:2018scaling} and \newcite{Kasai:2020deepshallow} respectively. The enwik8 experiment settings followed \newcite{Bachlechner:2020rezero} and the RoBERTa experiments followed \newcite{Liu:2019roberta}.

All the experiments in this paper were run with $3$ random seeds and the mean and standard deviation are reported. For the relatively small IWSLT, the $\hat{T}$ value in the AUCC metric was set to $4$ hours. For the larger WMT, we set it to $20$ hours. For enwiki8, it was $30$ hours; and for the RoBERTa pre-training experiments, it was set to $60$ hours.

The projection weights in random layers were initialized using orthogonal initialization \citep{Saxe:2013orthogonal}, since random orthogonal projections should ideally be maximally information-preserving, and which was found to work well empirically for initializing fixed random representations in previous work \citep{Wieting:2019notraining}. Biases and layer norm parameters were initialized using their respective PyTorch defaults \citep[based on Xavier init;][]{Glorot:2010init}.

We intersperse reservoir layers in alternating fashion starting from the middle. Specifically, we alternate one reservoir layer with one transformer layer, and place the alternating block in the middle. For example: a $7$-layer encoder $\text{LLLLLLL}$ in which we replace three layers with reservoirs becomes $\text{LRLRLRL}$, and with two becomes $\text{LLRLRLL}$. See Appendix \ref{appendix:freezing strategy} for a study comparing this strategy to alternative approaches~(e.g., freezing in the bottom, middle or top).

\begin{table*}[t]
    \centering
    \small
    \resizebox{\textwidth}{!}{
    \begin{tabular}{lccccccc}
    \toprule
        \multirow{2}{*}\textbf{Model} & \textbf{\# Layers} & \textbf{Frozen} & \textbf{Max BLEU} & \textbf{Train time}  & \textbf{Ratio} & \textbf{\# Params}  & \textbf{Train Time each}  \\
        &&&&\textbf{until max }(in hours)& & \textbf{Trainable} (Total) &\textbf{epoch} (in hours)
 \\\midrule
        \multirow{4}{*}{Transformer} 
        & 12 & 0 & 24.46 $\pm$ 0.04 & 15.15 $\pm$ 0.15 & 1 & 75.6M & 0.505 $\pm$ 0.005\\
        & 16 & 0 & 24.52 $\pm$ 0.03 & 16.05 $\pm$ 0.18 & 1 & 88.2M & 0.643 $\pm$ 0.006\\
        & 24 & 0 & 24.69 $\pm$ 0.05 & 17.61 $\pm$ 0.85 & 1 & 113.4M & 0.877 $\pm$ 0.029\\
        & 32 & 0 & 24.83 $\pm$ 0.04 & 18.42 $\pm$ 0.28 & 1 & 138.6M & 1.036 $\pm$ 0.010\\\midrule
        \multirow{4}{*}{T Reservoir} 
        & 12 & 4 & 24.26 $\pm$ 0.08 & 14.11 $\pm$ 0.21 & 0.93 & 72.4M (75.6M) & 0.472 $\pm$ 0.007\\
        & 16 & 4 & 24.50 $\pm$ 0.05 & 15.25 $\pm$ 0.28 & 0.95 & 75.6M (88.2M) & 0.596 $\pm$ 0.009\\
        & 24 & 4 & 25.11 $\pm$ 0.07 & 15.89 $\pm$ 0.74 & 0.90 & 100.8M (113.4M) & 0.776 $\pm$ 0.024\\
        & 32 & 4 & 24.66 $\pm$ 0.04 & 16.38 $\pm$ 0.24 & 0.88 & 126.0M (138.6M) & 0.998 $\pm$ 0.009\\\midrule
        \multirow{4}{*}{FFN Reservoir} 
        & 12 & 4 & 24.42 $\pm$ 0.05 & 14.01 $\pm$ 0.09 & 0.92 & 72.4M (71.4M) & 0.441 $\pm$ 0.003\\
        & 16 & 4 & 24.65 $\pm$ 0.07 & 14.53 $\pm$ 0.17 & 0.91 & 75.6M (83.9M) & 0.524 $\pm$ 0.006\\
        & 24 & 4 & 24.93 $\pm$ 0.04 & 12.62 $\pm$ 1.53 & 0.71 & 100.8M (109.2M) & 0.743 $\pm$ 0.018\\
        & 32 & 4 & 24.98 $\pm$ 0.03 & 13.96 $\pm$ 0.19 & 0.73 & 126.0M (134.4M) & 0.964 $\pm$ 0.007\\\midrule
        \multirow{4}{*}{LayerDrop} 
        & 12 & 4 & 24.27 $\pm$ 0.03 & 14.61 $\pm$ 0.14 & 0.96 & 72.4M (75.6M) & 0.489 $\pm$ 0.006\\
        & 16 & 4 & 24.15 $\pm$ 0.06 & 15.55 $\pm$ 0.54 & 0.97 & 75.6M (88.2M) & 0.597 $\pm$ 0.017\\
        & 24 & 4 & 24.37 $\pm$ 0.05 & 16.25 $\pm$ 0.36 & 0.92 & 100.8M (113.4M) & 0.823 $\pm$ 0.013\\
        & 32 & 4 & 23.84 $\pm$ 0.03 & 15.27 $\pm$ 0.38 & 0.83 & 126.0M (138.6M) & 1.028 $\pm$ 0.012\\
    \bottomrule
    \end{tabular}}
    \caption{Wall-clock time (averaged over multiple runs) saved for WMT for different model types and encoder depths. Decoder depth is kept fixed at 1.}
    \label{tab:wmt-time}
\end{table*}

\section{Experiments}

In what follows, we first show our main result, on a variety of tasks: reservoir transformers mostly have better AUCC metrics; less training time per epoch; less convergence time until the best validation performance is achieved; and even improved test set generalization metrics. As a strong baseline method, we compare to LayerDrop \citep{Fan:2019reducing}. LayerDrop can also be seen as a method that dynamically bypasses parts of the computation during Transformer training in an attempt to improve efficiency, and making it a strong comparison to examine our methods. Then, we examine whether we can minimize the expectation over the gradients of upstream layers in the network such that we do not \emph{at all} have to pass gradients through the reservoir layers, skipping their backward pass.

\subsection{Machine Translation}

Machine translation (MT) is one of the core tasks of NLP. We demonstrate on two well-known MT datasets, IWSLT'14 German-English and WMT'16 English-German, that reservoir transformers obtain a better AUCC. For the raw validation plots over time that were used to calculate the AUCC, please refer to Appendix \ref{appendix:validation_plot}.

Following \newcite{Kasai:2020deepshallow}, the architecture of the network is an N-layer reservoir transformer encoder, followed by a regular shallow one- or two-layer decoder. This design choice has been shown to lead to very good speed and efficiency trade-offs, and serves as a good baseline for our experiments. Moreover, shallow decoders make it easier to decide where to place reservoir layers~(in the encoder) and makes it more straightforward to identify where performance gains come from.


Figure \ref{fig:iwslt} shows the results for IWSLT (left) and WMT (middle). On the y-axis we show validation AUCC for the BLEU metric; on the x-axis we show the number of updatable layers in the encoder. The performance of a regular transformer encoder with 6 layers and a reservoir transformer encoder with 6 layers plus N additional reservoir layers are plotted for the same x-axis value to show the total number of \emph{updated} layers. Plots for the \emph{total} number of layers (updatable plus not-updatable, so essentially shifted versions of the plots) are shown in Appendix \ref{appendix:total_layer}.

WMT is much larger and requires a much deeper encoder, as illustrated by the fact that a certain minimum depth is required for reservoir transformers to achieve a comparable validation AUCC. At test time, reservoir transformers outperform regular transformers for almost all encoder depths. The FFN Reservoir seems to work best in both cases, which is surprising because it does not have any self-attention component at all. This finding shows that self-attention, or the mechanism to summarize context information, should be learned if present. Once the context features have been gathered, a random projection via a fixed FFN module appears to be beneficial.

Table \ref{tab:iwslt-time} and \ref{tab:wmt-time} show the time it took to achieve the maximum validation BLEU score and how that relates to the regular transformer, demonstrating that reservoir transformers consistently converge faster in terms of wall-clock time. We save up to 22\% convergence wall-clock time using reservoir transformers as much with the same number of updateable layers. We save as much as 27\% time until convergence a 24 layer model on WMT, as shown in Table \ref{tab:wmt-time}. One other noticeable point is that we can see that the T Reservoir achieves similar performance to LayerDrop on IWSLT and WMT in terms of wall-clock per epoch and wall-clock time to the best performance. However, on both tasks, FFN Reservoir performs much better than LayerDrop in terms of efficiency per epoch and achieves better/similar performance in less time in each case. As a point of reference, a half hour gain on IWSLT would translate to a gain of several days in the training of bigger transformer models like GPT-3~\citep{Brown:2020gpt3}.

We observe that reservoir transformers consistently perform better than, or are competitive to, regular transformers, both in terms of validation BLEU AUCC as well as test time BLEU, for all examined encoder depths.

\subsection{Language Modelling}

\begin{figure*}[t]
    \centering
    \begin{subfigure}{.4\textwidth}
        \centering
        \includegraphics[width=\textwidth]{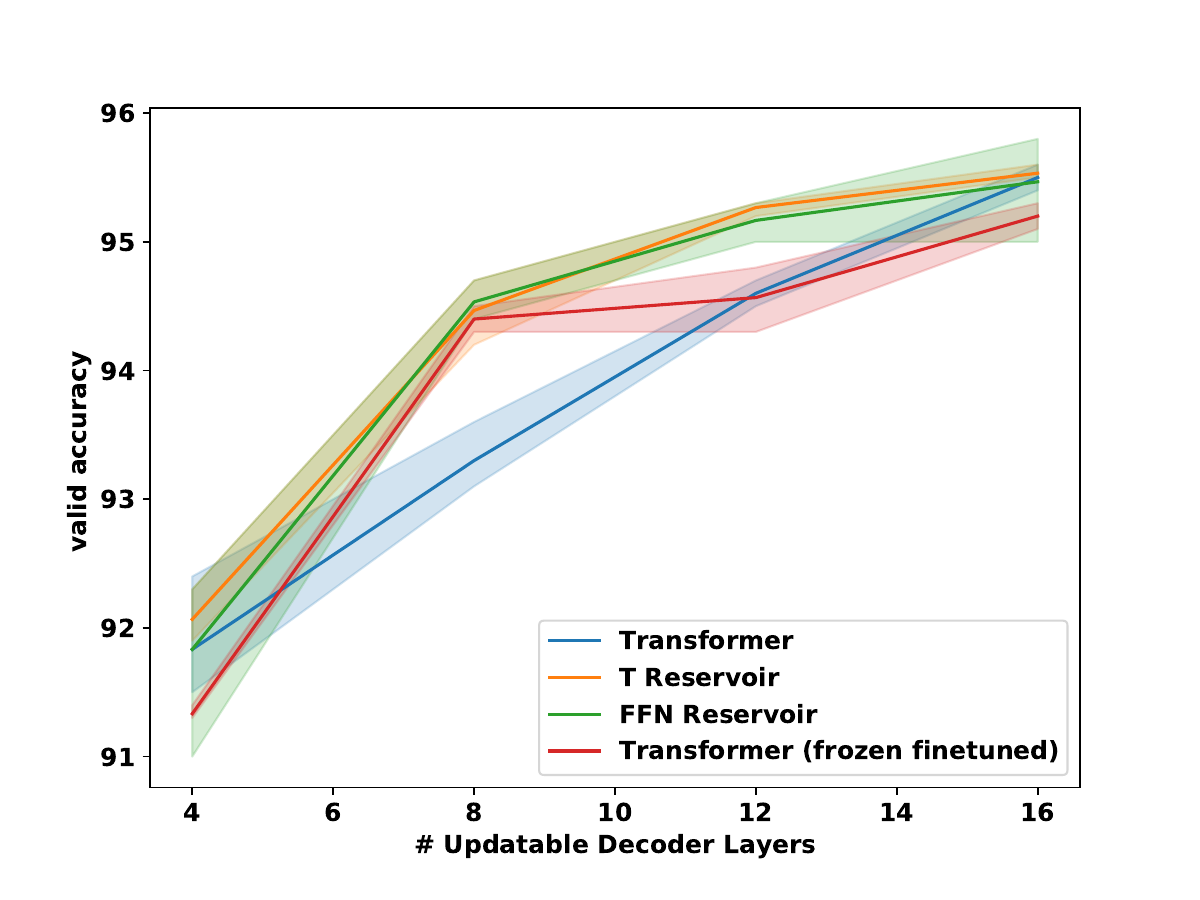}
    \end{subfigure}%
    \begin{subfigure}{.3\textwidth}
    \end{subfigure}%
    \begin{subfigure}{.4\textwidth}
        \centering
        \includegraphics[width=\textwidth]{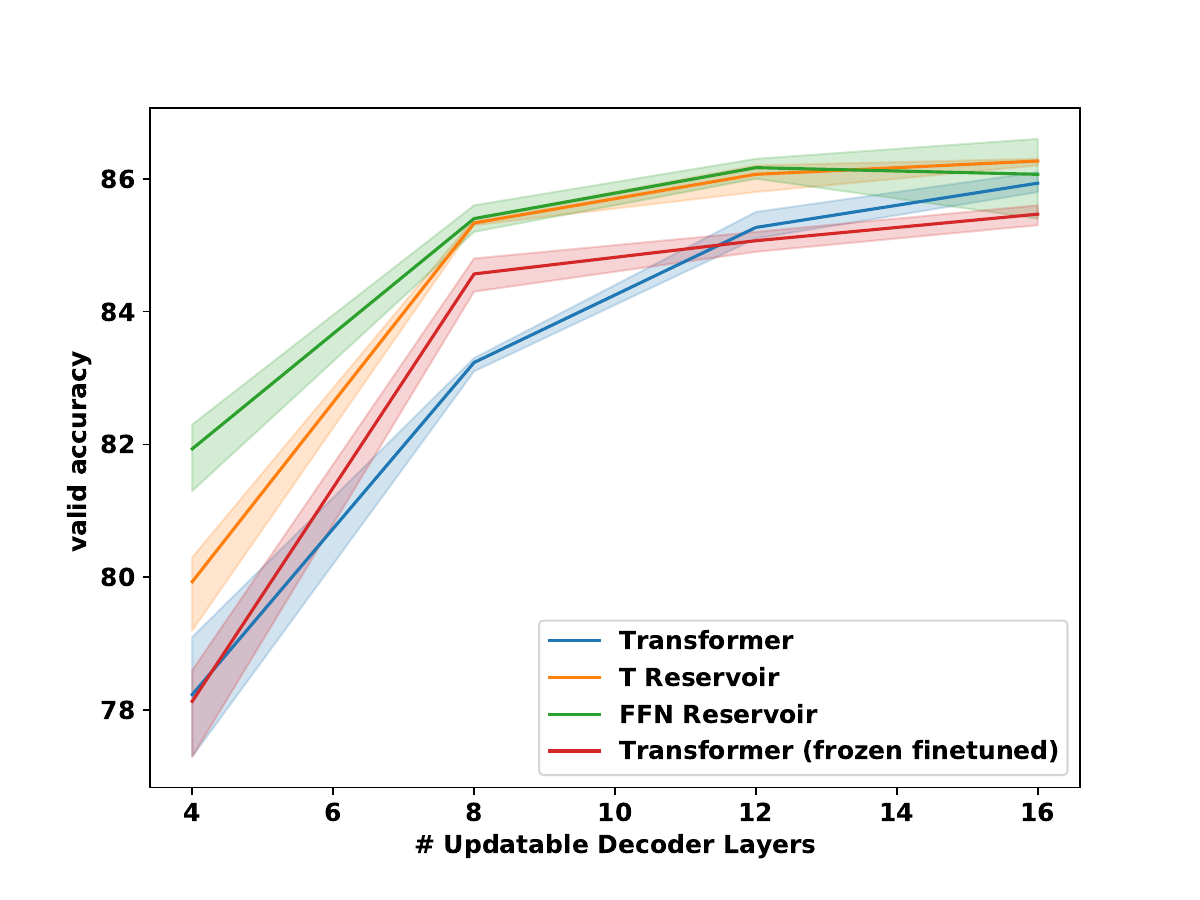}
    \end{subfigure}
    \caption{Downstream RoBERTa performance on SST-2 (left) and MultiNLI-matched (right).}
    \label{fig:finetune}
\end{figure*}

To examine whether the same findings hold for other tasks, we evaluate on the enwiki8 \citep{LLC:2009long} language modelling task. We examine the BPC (bits per character) rate for a variety of network depths (since the task is language modelling, these layers are in the decoder). The results show that except for the $64$-layer regular transformer, which appears to be particularly optimal for this task, we obtain consistently better BPC for all depths. We observe similar trends during test time.

\begin{table}[t]
    \centering
    \resizebox{0.47\textwidth}{!}{
    \begin{tabular}{lccccccc}
    \toprule
        \multirow{2}{*} \textbf{Model} & \textbf{Max BLEU}  & \textbf{AUCC} & \textbf{Train time}\\\midrule
        \multirow{1}{*}{Transformer} 
        & 34.59 $\pm$ 0.11 & 114.57 $\pm$ 0.08 & 142.28 $\pm$ 1.87\\\midrule
        \multirow{1}{*}{T Reservoir} 
        & 34.80 $\pm$ 0.07 & 115.26 $\pm$ 0.26 & 134.49 $\pm$ 1.70\\\midrule
        \multirow{1}{*}{Backskip Reservoir} 
        & 34.75 $\pm$ 0.05 & 115.99 $\pm$ 0.23 & 119.54 $\pm$ 1.78\\
    \bottomrule
    \end{tabular}}
    \caption{Validation max BLEU, AUCC at 4h and wall-clock time per epoch (averaged over multiple runs, in seconds) on IWSLT comparing backskipping with regular and reservoir transformers.}
    \label{tab:backskipped-time}
\end{table}

\subsection{Masked Language Model Pretraining}

We train RoBERTa \citep{Liu:2019roberta} models from scratch at a variety of depths, both in the normal and reservoir setting. We find that these networks show minor differences in their best perplexity and similar AUCC perplexity~(see Appendix \ref{appendix:roberta_result}). We then examine the performance of these models when fine-tuned on downstream tasks, specifically the well known SST-2~\citep{Socher:2013recursive} and MultiNLI-matched ~\citep{Williams2017mnli} tasks. When fine-tuning the reservoir models, we keep the reservoir layers fixed (also fine-tuning them did not work very well, see Appendix \ref{appendix:roberta_result}).

Figure \ref{fig:finetune} shows the results of fine-tuning. We observe that the reservoir transformer outperforms normal RoBERTa at all depths in both tasks. At lower depth, the improvements are substantial. As a sanity check, we also experiment with freezing some of the layers in a regular pre-trained RoBERTa model during fine-tuning only (Transformer ``frozen finetuned'' in the Figure) and show that this helps a little but is still outperformed by the reservoir transformer.

These findings suggest that we can train a RoBERTa model without updating all of the layers, achieving similar perplexity at a similar computational cost, but with \emph{better} downstream performance. This strategy could prove to be beneficial in a wide variety of pre-training scenarios.

We follow \newcite{jawahar-etal-2019-bert} and investigate what the frozen layers in the Reservoir Transformer have actually ``learned'' (while being frozen) as measured by probing tasks, reported in Table \ref{tab:roberta-probing}. 
The set of tasks comprises one surface task, three syntactic tasks, and five semantic tasks. 
From the table, we can see that generally probing performance is quite similar between Transformer and the T Reservoir model. 
We also noticed that the representations collected after the reservoir layer (3, 5, 7, 9) in the T Reservoir actually have significantly better performance over the regular Transformer representations across all the probing tasks. 
Related to our findings, \newcite{voita2020information} show that the wholly-randomly-initialized model representations can still have reasonable probing accuracy if they are contextualized, though the accuracy is strictly worse than a trained one. 
These findings raise interesting repercussions for the study of ``BERTology'', as it clearly shows that even completely random and frozen layers can represent linguistic phenomena. 

\begin{table*}[t]
    \centering
    \small
    \resizebox{\textwidth}{!}{
    \begin{tabular}{lccccccccccc}
    \toprule
        \multirow{2}{*} \textbf{Model} & \textbf{Layer} & \textbf{SentLen} & \textbf{TreeDepth} & \textbf{TopConst}  & \textbf{BShift} & \textbf{Tense}  & \textbf{SubjNum} & \textbf{ObjNum} & \textbf{SOMO} & \textbf{CoordInv} \\
        & & (Surface) & (Syntactic) & (Syntactic) & (Syntactic) & (Semantic) & (Semantic) & (Semantic)& (Semantic) & (Semantic) \\\midrule
        \multirow{12}{*}{Transformer} 
        & 1 & 84.56 $\pm$ 0.54 &	32.30 $\pm$ 0.41 &	54.40 $\pm$ 0.33 &	49.99 $\pm$ 0.01 &	80.98 $\pm$ 0.32 &	76.26 $\pm$ 0.09 &	50.01 $\pm$ 0.19 &	76.38 $\pm$ 0.61 &	54.33 $\pm$ 0.47 \\
& 2 & 87.22 $\pm$ 0.07 &	33.63 $\pm$ 0.57 &	58.38 $\pm$ 0.20 &	50.12 $\pm$ 0.17 &	82.84 $\pm$ 0.68 &	 78.65 $\pm$ 0.19 &	51.47 $\pm$ 0.53 &	78.00 $\pm$ 1.12 &	54.66 $\pm$ 0.55 \\
& 3 & 84.25 $\pm$ 0.16 &	32.60 $\pm$ 0.17 &	54.41 $\pm$ 0.10 &	50.02 $\pm$ 0.01 &	81.72 $\pm$ 0.59 &	77.00 $\pm$ 0.13 &	51.32 $\pm$ 0.64 &	76.57 $\pm$ 1.13 &	54.13 $\pm$ 0.51 \\ 
& 4 & 87.37 $\pm$ 0.20 &	32.59 $\pm$ 0.29 &	50.06 $\pm$ 0.21 &	69.76 $\pm$ 0.26 &	81.63 $\pm$ 1.17 &	76.47 $\pm$ 0.09 &	52.41 $\pm$ 1.49 &	76.15 $\pm$ 0.84 &	52.62 $\pm$ 1.34 \\ 
& 5 & 84.61 $\pm$ 0.24 &	31.14 $\pm$ 0.48 &	44.76 $\pm$ 0.38 &	74.82 $\pm$ 0.11 &	80.16 $\pm$ 0.19 &	73.66 $\pm$ 0.16 &	52.95 $\pm$ 1.77 &	72.90 $\pm$ 0.21 &	51.26 $\pm$ 1.14 \\ 
& 6 & 82.56 $\pm$ 0.25 &	30.31 $\pm$ 0.40 &	39.30 $\pm$ 0.40 &	78.80 $\pm$ 0.38 &	81.88 $\pm$ 0.47 &	75.30 $\pm$ 0.07 &	56.21 $\pm$ 1.26 &	74.37 $\pm$ 0.16 &	51.44 $\pm$ 1.04 \\ 
& 7 & 70.85 $\pm$ 0.13 &	26.65 $\pm$ 0.72 &	40.70 $\pm$ 0.13 &	78.98 $\pm$ 0.32 &	85.11 $\pm$ 0.31 &	72.03 $\pm$ 0.46 &	58.15 $\pm$ 0.46 &	68.71 $\pm$ 0.91 &	55.39 $\pm$ 0.27 \\ 
& 8 & 66.23 $\pm$ 1.33 &	23.46 $\pm$ 0.44 &	25.19 $\pm$ 1.02 &	77.42 $\pm$ 0.27 &	80.35 $\pm$ 0.45 &	67.55 $\pm$ 0.99 &	54.94 $\pm$ 2.04 &	63.69 $\pm$ 2.32 &	50.58 $\pm$ 0.83 \\ 
& 9 & 71.17 $\pm$ 0.29 &	31.21 $\pm$ 0.31 &	58.42 $\pm$ 0.29 &	85.55 $\pm$ 0.44 &	86.77 $\pm$ 0.19 &	80.30 $\pm$ 0.08 &	64.36 $\pm$ 1.20 &	81.68 $\pm$ 0.45 &	66.90 $\pm$ 0.49 \\ 
& 10 & 73.19 $\pm$ 0.50 &	27.74 $\pm$ 0.53 &	41.01 $\pm$ 0.22 &	83.56 $\pm$ 0.96 &	86.13 $\pm$ 0.35 &	83.04 $\pm$ 0.04 &	62.01 $\pm$ 0.59 &	79.73 $\pm$ 0.21 &	62.60 $\pm$ 1.04 \\ 
& 11 & 71.37 $\pm$ 0.42 &	30.22 $\pm$ 0.28 &	48.58 $\pm$ 0.35 &	84.40 $\pm$ 0.44 &	87.28 $\pm$ 0.59 &	82.34 $\pm$ 0.15 &	61.10 $\pm$ 0.14 &	80.00 $\pm$ 0.40 &	64.44 $\pm$ 0.38 \\ 
& 12 & 71.66 $\pm$ 0.12 &	33.43 $\pm$ 0.18 &	64.38 $\pm$ 0.20 &	87.38 $\pm$ 0.02 &	88.41 $\pm$ 0.09 &	84.46 $\pm$ 0.25 &	63.01 $\pm$ 0.05 &	81.80 $\pm$ 0.27 &	65.72 $\pm$ 0.16 \\
        \midrule
        \multirow{12}{*}{T Reservoir} 
&  1 & 87.75 $\pm$ 0.10 &	31.60 $\pm$ 0.21 &	50.38 $\pm$ 0.23 &	50.00 $\pm$ 0.00 &	80.40 $\pm$ 0.18 &	76.47 $\pm$ 0.20 &	50.53 $\pm$ 0.14 &	73.48 $\pm$ 0.15 &	53.55 $\pm$ 0.70 \\ 
 & 2 & 81.28 $\pm$ 0.23 &	34.20 $\pm$ 0.41 &	61.41 $\pm$ 0.42 &	60.64 $\pm$ 0.65 &	81.50 $\pm$ 0.77 &	76.33 $\pm$ 0.08 &	50.73 $\pm$ 0.34 &	74.28 $\pm$ 0.67 &	56.82 $\pm$ 0.10 \\ 
 & \bfseries 3 & \bfseries 89.28 $\pm$ 0.09 &\bfseries 36.42 $\pm$ 0.11 & \bfseries	67.36 $\pm$ 0.45 &	\bfseries 75.64 $\pm$ 0.52 &	\bfseries 85.42 $\pm$ 0.18 & \bfseries	80.53 $\pm$ 0.02 & \bfseries	52.50 $\pm$ 1.80 & \bfseries	78.47 $\pm$ 1.81 & \bfseries	57.16 $\pm$ 0.27 \\ 
 & 4 & 74.31 $\pm$ 0.32 &	32.42 $\pm$ 0.83 &	55.19 $\pm$ 0.33 &	73.41 $\pm$ 0.00 &	79.56 $\pm$ 0.00 &	75.15 $\pm$ 0.08 &	53.68 $\pm$ 0.66 &	75.02 $\pm$ 0.19 &	56.89 $\pm$ 0.08 \\ 
 &  \bfseries 5 & \bfseries 88.03 $\pm$ 0.22 & \bfseries	38.34 $\pm$ 0.64 & \bfseries	68.65 $\pm$ 0.29 &	\bfseries 82.25 $\pm$ 0.12 & \bfseries	86.80 $\pm$ 0.02 & \bfseries	82.27 $\pm$ 0.33 & \bfseries 57.95 $\pm$ 0.24 &	 \bfseries 80.82 $\pm$ 0.91 & \bfseries 58.05 $\pm$ 0.10 \\ 
& 6 & 74.55 $\pm$ 0.37 &	33.13 $\pm$ 0.29 &	52.70 $\pm$ 0.81 &	79.21 $\pm$ 0.13 &	85.70 $\pm$ 0.36 &	77.43 $\pm$ 0.03 &	57.26 $\pm$ 0.19 &	75.38 $\pm$ 0.66 &	51.95 $\pm$ 1.30 \\ 
& \bfseries 7 & \bfseries 85.82 $\pm$ 0.37 &	\bfseries 37.63 $\pm$ 0.13 & \bfseries	70.43 $\pm$ 0.05 &	\bfseries 84.12 $\pm$ 0.35 &	\bfseries 86.88 $\pm$ 0.07 & \bfseries	82.86 $\pm$ 0.30 & \bfseries	61.17 $\pm$ 0.21 & \bfseries	80.79 $\pm$ 0.17 & \bfseries	61.83 $\pm$ 0.95 \\ 
& 8 & 71.69 $\pm$ 0.71 &	30.32 $\pm$ 0.01 &	48.44 $\pm$ 0.30 &	79.12 $\pm$ 0.12 &	84.75 $\pm$ 0.09 &	79.23 $\pm$ 0.11 &	59.53 $\pm$ 0.16 &	76.80 $\pm$ 0.41 &	57.34 $\pm$ 0.14 \\ 
& \bfseries 9 & \bfseries 85.86 $\pm$ 0.12 & \bfseries	37.89 $\pm$ 0.03 & \bfseries	69.53 $\pm$ 0.37 & \bfseries	85.55 $\pm$ 0.12 & \bfseries	87.98 $\pm$ 0.22 & \bfseries	84.13 $\pm$ 0.01 &	\bfseries 63.06 $\pm$ 0.01 & \bfseries	82.55 $\pm$ 0.31 & \bfseries	66.07 $\pm$ 0.05 \\ 
& 10 & 69.22 $\pm$ 0.23 &	25.58 $\pm$ 0.35 &	29.20 $\pm$ 0.58 &	78.57 $\pm$ 0.09 &	85.02 $\pm$ 0.03 &	75.68 $\pm$ 0.16 &	57.55 $\pm$ 1.57 &	74.70 $\pm$ 0.02 &	55.02 $\pm$ 0.64 \\ 
& 11 & 65.70 $\pm$ 0.05 &	30.57 $\pm$ 0.03 &	47.56 $\pm$ 0.02 &	81.20 $\pm$ 0.00 &	86.78 $\pm$ 0.02 &	83.73 $\pm$ 0.05 &	60.38 $\pm$ 0.17 &	80.59 $\pm$ 0.15 &	62.50 $\pm$ 0.11 \\ 
& 12 & 70.61 $\pm$ 0.18 &	34.45 $\pm$ 0.20 &	64.19 $\pm$ 0.10 &	84.53 $\pm$ 0.03 &	87.48 $\pm$ 0.16 &	84.86 $\pm$ 0.14 &	62.75 $\pm$ 0.14 &	82.08 $\pm$ 0.03 &	64.73 $\pm$ 0.06 \\
        \midrule
    \bottomrule
    \end{tabular}}
    \caption{RoBERTa Probing Results. The line in bold text are the the frozen layers in the T Reservoir. Mean accuracy with standard deviation, gathered over 3 random seeds.}
    \label{tab:roberta-probing}
\end{table*}

\subsection{Backskipping}

With the reservoir transformers as described above, we obtain better efficiency by skipping the ``gradient application'' matrix addition step in some of the layers (i.e., updating the weights). One step further would be to investigate skipping the entire backward pass for reservoirs altogether, which would save us from having to do the much more expensive matrix multiplication for these layers that is required for the propagation of gradients through a regular layer.

We report on preliminary experiments where in the backward pass we replace the gradients for the layer $L_i$ \emph{going into} the reservoir $L_{i+1}$ with a noisy estimate~\citep{Jaderberg:2017synthetic,Czarnecki:2017dni}. Promisingly, \newcite{Oktay:2020randomizedautodiff} recently asked ``why spend resources on exact gradients when we’re going to use stochastic optimization?'' and show that we can do randomized auto-differentiation quite successfully.

Here, rather than minimizing the actual gradients $\frac{\partial L_i}{\partial \theta^{L_i}}$, we minimize their expectation and train via continuous-action REINFORCE \citep{Williams1992reinforce}. That is, $L_i$ becomes a policy $\pi_a$: $s \rightarrow \mu$
where we sample actions $a \sim \mathcal{N}(\mu, 1)$. We train to minimize the gradient prediction loss via MSE, i.e., $\frac{1}{n}\sum_{i=0}^{n}(R^i-V^i(a))^2$, and the REINFORCE loss $\mathbb{E}_a\left[\log (a) \left(R - V(a) \right)\right]$, where the value network $V$ acts as the baseline. $R$ is defined as 
the mean of the gradients of the top layer $L_{i+2}$, with the sign flipped. Thus, simply put, we train to minimize the expectation of the true gradients at the layer directly following the reservoir. We employ an annealing scheme where we first train the value network and propagate the true gradients during warmup. Afterwards, we anneal the probability of backskipping instead of doing a true backward pass~(multiplying the probability by $0.99$ every iteration until we only backskip). 
We experimented with setting $R$ to the negation of the total loss but found the mean upstream gradient reward to work better. We call this approach \emph{backskipping}.

As shown in Table \ref{tab:backskipped-time}, the backskip reservoir approach leads to a higher maximum BLEU score than the regular transformer, with a much higher AUCC and much lower training time. The encoder depth is 8 with 2 frozen. Appendix \ref{appendix:backskipping} shows the raw validation BLEU curves over time. We observe that this approach helps especially during the earlier stages of training. This finding opens up intriguing possibilities for having parts of neural networks be completely frozen both in the forward as well as in the backward pass, while still contributing to the overall model computation.

The computational cost is heavily reduced given that we completely bypass the expensive backpropagation computation in the reservoir layers. Backskipping is shown to be a promising approach to further reduce computational costs, and would be even more efficient from a hardware perspective since the circuitry for such layers (which do not need to propagate gradients) can be hardwired.

\section{Related Work}

Recent work has shown that modern NLP models are able to function with different numbers of layers for different examples~\citep{Elbayad:2019depth,Fan:2019reducing,he2021pipetransformer}; that different layers specialize for different purposes~\citep{Zhang:2019alllayersequal}; that layers can be compressed~\citep{Li:2020traincompress,zhu-etal-2019-panlp,shen2020q,sun2020mobilebert}; and, that layers can be reordered~\citep{Press2019improving}. 
There is a growing body of work in efficient self-attention networks \citep{Tay:2020transformersurvey}, such as linear attention~\citep{Wang:2020linformer}, on how to process long context information~\citep{Beltagy:2020longformer,ainslie-etal-2020-etc} and on approximations to make transformers more scalable~\citep{Kitaev:2020reformer,Katharopoulos:2020linearattn}.
BigBIRD \citep{Zaheer2020big} provides random keys as additional inputs to its attention mechanism. Locality sensitive hashing (LSH) as employed e.g. in Reformer \citep{Kitaev:2020reformer} utilizes a fixed random projection. Random Feature Attention~\citep{peng2021random} uses random feature methods to approximate the softmax function. Performer~\citep{Choromanski:2020performer} computes the transformer's multi-head attention weights as a fixed orthogonal random projection. Closely related to this work,~\newcite{Tay:2020synthesizer} showed that randomized alignment matrices in their ``Synthesizer'' architecture are sufficient for many NLP tasks. While these works focus on random attention, we show that \emph{entire} layers can be random and fixed. We also show that entire layers can be replaced by fixed random projections that do not have any attention whatsoever.

Beyond transformers, random features have been extensively explored. Examples of this include FreezeOut \citep{Brock:2017freezeout}, deep reservoir computing networks \citep{Scardapane:2017randomness,Gallicchio:2017echo}, as well as applications in domains as varied as text classification~\citep{Conneau:2017infersent,Zhang:2018language,Wieting:2019notraining} or music classification \citep{Pons2019randomly}. It is well known that randomly initialized networks can display impressive performance on their own~\citep{Ulyanov:2018deepimageprior,Rosenfeld:2019intriguing,Ramanujan:2020cvpr,voita2020information}, which underlies, for example, the recently popularized lottery ticket hypothesis \citep{Frankle:2018lottery,Zhou:2019deconstructing}. We know that learning deep overparameterized networks appears to help in general~\citep{Li:2018overparameterized,Du2019gradient}. Our method constitutes a way to add both depth and parameters to transformer networks without much computational cost.

\section{Conclusion}

This work demonstrated that state-of-the-art transformer architectures can be trained without updating all of the layers. This complements a long history in machine learning of harnessing the power of random features.
We use the ``area under the convergence curve'' (AUCC) metric to demonstrate that on a variety of tasks, and in a variety of settings, ``reservoir transformers'' achieve better performance-efficiency trade-offs. We show that such reservoir transformers show better convergence rates and test-set generalization. We demonstrated that the backward pass can be skipped altogether, opening up exciting vanues for future research. Future work includes further investigating hybrid networks and backskipping strategies, as well as utilizing pruning.

\section*{Acknowledgements}
We thank Eric Wallace, Zhewei Yao, Kevin Lin, Zhiqing Sun, Zhuohan Li, Angela Fan, Shaojie Bai, and anonymous reviewers for their comments and suggestions.
SS and KK were supported by grants from Samsung, Facebook, and the Berkeley Deep Drive Consortium. 
\bibliography{acl2021}
\bibliographystyle{acl2021}

\appendix

\begin{table*}[h]
    \centering
    \small
    \resizebox{\textwidth}{!}{
    \begin{tabular}{lccccccc}
    \toprule
        \multirow{2}{*}\textbf{Model} & \textbf{\# Layers} & \textbf{Frozen} & \textbf{Max BLEU} & \textbf{Train time}  & \textbf{Ratio} &  \textbf{\# Params}  &  \textbf{Train Time each}  \\
        &&&&\textbf{until max }(in hours)& &  \textbf{Trainable} (Total) &  \textbf{epoch} (in seconds)
 \\\midrule
        \multirow{4}{*}{Transformer} 
        & 6 & 0 & 34.97 $\pm$ 0.05 & 1.984 $\pm$ 0.02 & 1 & 39.5M & 177.84 $\pm$ 2.98\\
        & 8 & 0 & 34.99 $\pm$ 0.08 & 2.161 $\pm$ 0.03 & 1 & 43.7M & 206.59 $\pm$ 3.47\\
        & 10 & 0 & 34.98 $\pm$ 0.04 & 2.345 $\pm$ 0.02 & 1 & 47.9M & 236.72 $\pm$ 3.52\\
        & 12 & 0 & 34.78 $\pm$ 0.11 & 2.535 $\pm$ 0.05 & 1 & 52.0M &  265.90 $\pm$ 4.97\\\midrule
        \multirow{4}{*}{T Reservoir} 
        & 6 & 2 & 34.73 $\pm$ 0.11 & 1.838 $\pm$ 0.01 & 0.92 & 35.3M (39.5M) & 166.11 $\pm$ 2.21\\
        & 8 & 2 & 35.07 $\pm$ 0.05 & 1.912 $\pm$ 0.03 & 0.88 & 39.5M (43.7M) & 190.08 $\pm$ 3.73\\
        & 10 & 2 & 35.02 $\pm$ 0.01 & 1.970 $\pm$ 0.04 & 0.84 & 43.7M (47.9M) & 204.42 $\pm$ 2.89\\
        & 12 & 2 & 35.06 $\pm$ 0.02 & 2.429 $\pm$ 0.02 & 0.95 & 47.8M (52.0M) & 236.41 $\pm$ 4.35\\\midrule
        \multirow{4}{*}{FFN Reservoir} 
        & 6 & 2 & 34.85 $\pm$ 0.10 & 1.729 $\pm$ 0.03 & 0.87 & 35.3M (37.4M) & 161.72 $\pm$ 2.32\\
        & 8 & 2 & 34.99 $\pm$ 0.11 & 1.751 $\pm$ 0.02 & 0.81 & 39.5M (41.6M) & 180.21 $\pm$ 2.68\\
        & 10 & 2 & 34.92 $\pm$ 0.03 & 1.907 $\pm$ 0.02 & 0.81 & 43.7M (45.8M) & 191.40 $\pm$ 2.49\\
        & 12 & 2 & 35.16 $\pm$ 0.04 & 2.395 $\pm$ 0.01 & 0.94 & 47.8M (49.9M) & 216.08 $\pm$ 2.57\\\midrule
        \multirow{4}{*}{LayerDrop} 
        &  6 &  2 &  34.51 $\pm$ 0.12 &  1.908 $\pm$ 0.04 &  0.96 &  35.3M (39.5M) &  169.62 $\pm$ 3.16\\
        &  8 &  2 &  34.77 $\pm$ 0.11 &  2.023 $\pm$ 0.02 &  0.94 &  39.5M (43.7M) &  186.71 $\pm$ 2.17\\
        &  10 &  2 &  34.06 $\pm$ 0.05 &  1.912 $\pm$ 0.02 &  0.97 &  43.7M (47.9M) &  205.52 $\pm$ 3.31\\
        &  12 &  2 &  34.08 $\pm$ 0.13 &  2.524 $\pm$ 0.01 &  0.99 &  47.8M (52.0M) &  222.45 $\pm$ 2.21\\
    \bottomrule
    \end{tabular}}
    \caption{Wall-clock time (averaged over multiple runs) for IWSLT for different model types and encoder depths. Max BLEU is for validation. Number of layers is for encoder, decoder depth is kept fixed at 6. Ratio is computed compared to comparable number of layers in the normal case.}
    \label{tab:iwslt-time-dec6}
\end{table*}

\section{Hybrid Networks and Non-Transformer Reservoirs}
\label{appendix:hybrids}

\begin{figure}[h]
    \centering
     \begin{subfigure}{.25\textwidth}
        \centering
        \includegraphics[width=\textwidth]{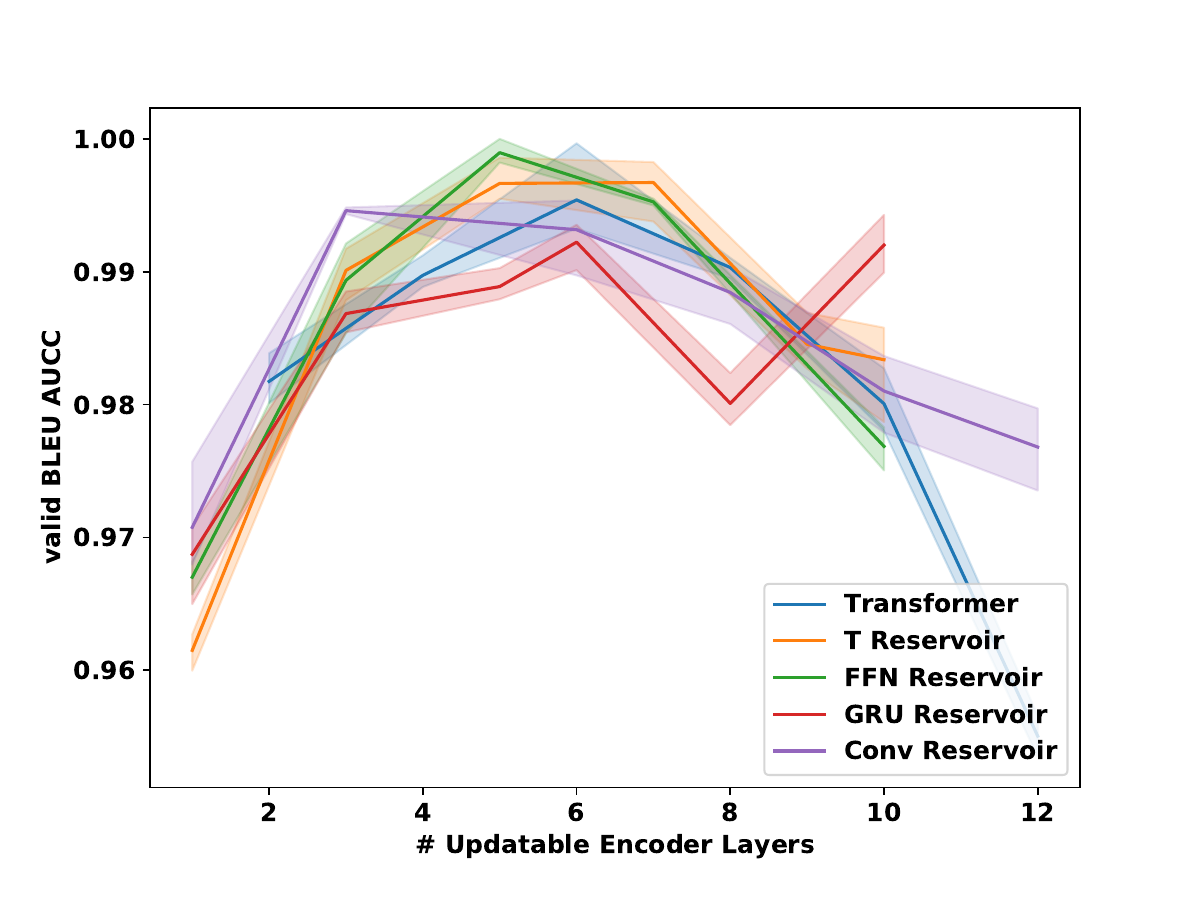}
    \end{subfigure}%
    \begin{subfigure}{.25\textwidth}
        \centering
        \centering
        \includegraphics[width=\textwidth]{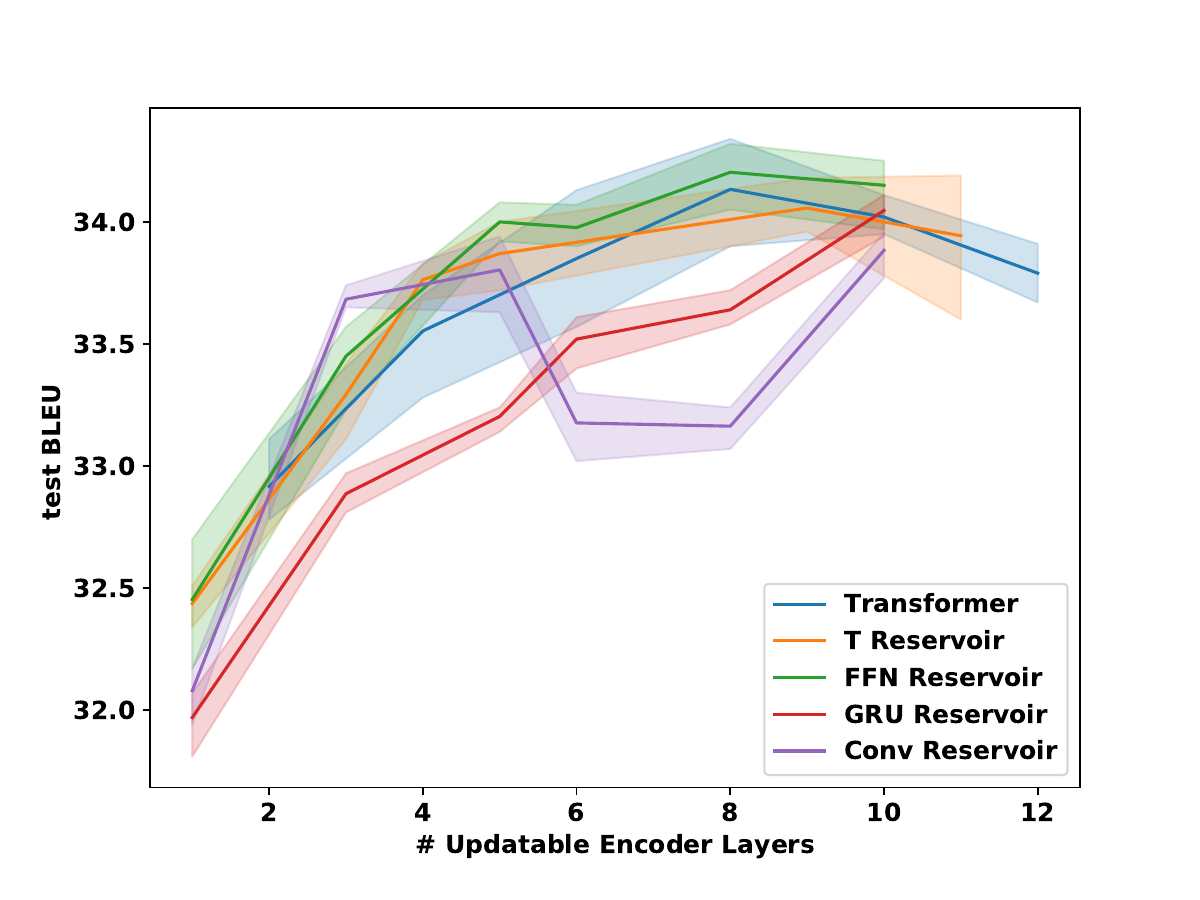}
    \end{subfigure}
    \caption{IWSLT comparison of different hybrid architectures with different reservoir layers.}
    \label{fig:hybrid}
\end{figure}

We investigate whether reservoir layers need to be transformer-based~(or transformers-without-attention, i.e., FFN). We examine two different alternatives: bidirectional Gated Recurrent Units \citep{Cho:2014gru} and Convolutional Neural Networks \citep{Lecun:1998cnn,Kim2014convolutional}, specifically light dynamical convolutions~\citep{Wu2019pay}. Figure \ref{fig:hybrid} shows the results for these hybrids: depending on the setting, they may obtain a better AUCC than the regular transformer, but this is less consistent than with the other reservoir layers, most likely because these layers have different computational properties. It's possible that these hybrids simply require further tuning, as we found e.g. up-projecting to help for BiGRUs, but studying this is outside of the scope of the current work.

\section{Deep Decoders}

\begin{figure}[h]
    \centering
    \begin{subfigure}{.25\textwidth}
        \centering
        \includegraphics[width=\textwidth]{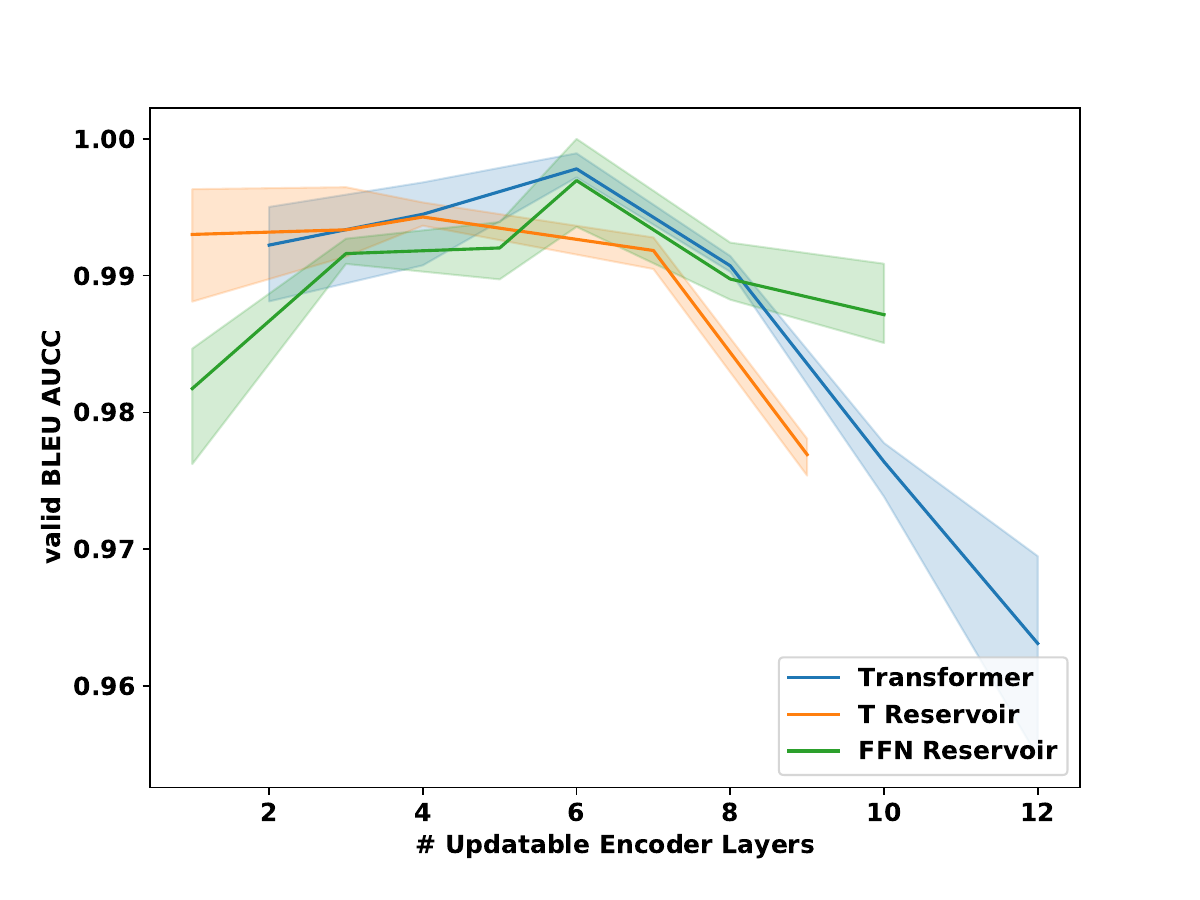}
    \end{subfigure}%
    \begin{subfigure}{.25\textwidth}
        \centering
        \includegraphics[width=\textwidth]{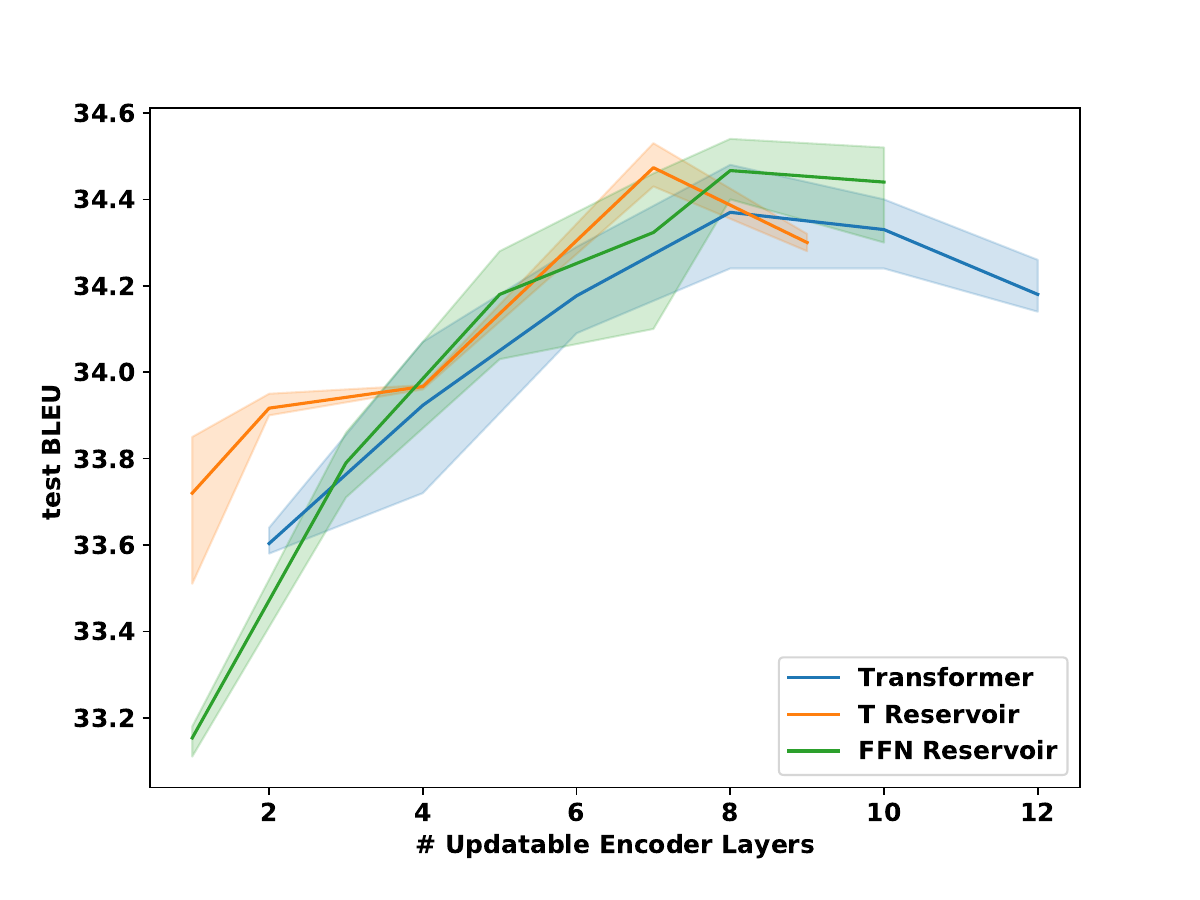}
    \end{subfigure}
    \caption{IWSLT validation AUCC and test BLEU with 6-layer decoder.}
    \label{fig:iwslt-6}
\end{figure}

We show that the same results hold for a 6-layer decoder on IWSLT (although less pronounced for AUCC, probably because the decoder is computationally heavier). See Figure \ref{fig:iwslt-6} and Table \ref{tab:iwslt-time-dec6}.

\begin{table*}[t]
    \centering
    \small
    \resizebox{\textwidth}{!}{
    \begin{tabular}{lcccccccccc}
    \toprule
        \multirow{3}{*}\textbf{Model} & & \textbf{IWSLT-Dec2} & & & \textbf{IWSLT-Dec6} & & & \textbf{WMT-Dec1} & &  
        \\ & \textbf{\# Layers} & \textbf{Train time} & \textbf{ Max BLEU}  & \textbf{\# Layers} & \textbf{Train time} & \textbf{Max BLEU} & \textbf{\# Layers} & \textbf{Train time} & \textbf{Max BLEU}  \\
        &&\textbf{until 95\% max }(in hours)&(95\%)&&\textbf{until 95\% max }(in hours) &(95\%) & & \textbf{until 95\% max }(in hours) & (95\%)
 \\\midrule
        \multirow{4}{*}{Transformer} 
        &  6 & 0.647 $\pm$ 0.03 & 32.89 $\pm$ 0.04 &  6 & 0.642 $\pm$ 0.02 & 33.36 $\pm$ 0.03 & 12 & 3.788 $\pm$ 0.053 & 23.36 $\pm$ 0.06\\
        &  8 & 0.711 $\pm$ 0.05 & 33.04 $\pm$ 0.03 &  8 & 0.765 $\pm$ 0.03 & 33.41 $\pm$ 0.08 & 16 & 3.820 $\pm$ 0.072 & 23.41 $\pm$ 0.05\\
        & 10 & 0.808 $\pm$ 0.02 & 33.96 $\pm$ 0.08 & 10 & 0.898 $\pm$ 0.04 & 33.32 $\pm$ 0.07 & 24 & 5.262 $\pm$ 0.607 & 23.50 $\pm$ 0.03\\
        & 12 & 1.037 $\pm$ 0.03 & 33.07 $\pm$ 0.09 & 12 & 1.037 $\pm$ 0.03 & 33.07 $\pm$ 0.11 & 32 & 6.212 $\pm$ 0.232 & 23.81 $\pm$ 0.04\\\midrule
        \multirow{4}{*}{T Reservoir} 
        &  6 & 0.569 $\pm$ 0.02 & 32.78 $\pm$ 0.03 &  6 & 0.599 $\pm$ 0.01 & 33.09 $\pm$ 0.05 & 12 & 3.563 $\pm$ 0.061 & 23.21 $\pm$ 0.04\\
        &  8 & 0.619 $\pm$ 0.04 & 33.12 $\pm$ 0.05 &  8 & 0.726 $\pm$ 0.02 & 33.38 $\pm$ 0.09 & 16 & 3.603 $\pm$ 0.056 & 23.80 $\pm$ 0.06\\
        & 10 & 0.729 $\pm$ 0.04 & 33.13 $\pm$ 0.07 & 10 & 0.738 $\pm$ 0.03 & 33.37 $\pm$ 0.04 & 24 & 4.923 $\pm$ 0.771 & 23.75 $\pm$ 0.02\\
        & 12 & 0.982 $\pm$ 0.02 & 33.03 $\pm$ 0.11 & 12 & 0.958 $\pm$ 0.01 & 33.46 $\pm$ 0.09 & 32 & 5.780 $\pm$ 0.214 & 23.71 $\pm$ 0.03\\\midrule
        \multirow{4}{*}{FFN Reservoir} 
        &  6 & 0.521 $\pm$ 0.05 & 32.85 $\pm$ 0.02 &  6 & 0.594 $\pm$ 0.03 & 33.13 $\pm$ 0.04 & 12 & 3.417 $\pm$ 0.046 & 23.22 $\pm$ 0.07\\
        &  8 & 0.533 $\pm$ 0.03 & 33.84 $\pm$ 0.04 &  8 & 0.651 $\pm$ 0.04 & 33.36 $\pm$ 0.06 & 16 & 3.527 $\pm$ 0.063 & 23.54 $\pm$ 0.05\\
        & 10 & 0.614 $\pm$ 0.01 & 33.05 $\pm$ 0.08 & 10 & 0.627 $\pm$ 0.05 & 33.26 $\pm$ 0.03 & 24 & 4.197 $\pm$ 0.697 & 23.74 $\pm$ 0.06\\
        & 12 & 0.811 $\pm$ 0.02 & 33.26 $\pm$ 0.10 & 12 & 0.780 $\pm$ 0.02 & 33.46 $\pm$ 0.08 & 32 & 4.984 $\pm$ 0.321 & 23.82 $\pm$ 0.02\\\midrule
        \multirow{4}{*}{LayerDrop} 
        &  6 & 0.837 $\pm$ 0.08 & 32.87 $\pm$ 0.05 &  6 & 0.706 $\pm$ 0.01 & 33.08 $\pm$ 0.03 & 12 & 3.912 $\pm$ 0.068 & 23.33 $\pm$ 0.08\\
        &  8 & 0.934 $\pm$ 0.07 & 33.12 $\pm$ 0.03 &  8 & 0.753 $\pm$ 0.04 & 33.14 $\pm$ 0.05 & 16 & 3.581 $\pm$ 0.076 & 23.17 $\pm$ 0.04\\
        & 10 & 0.901 $\pm$ 0.06 & 33.18 $\pm$ 0.02 & 10 & 0.691 $\pm$ 0.03 & 32.39 $\pm$ 0.05 & 24 & 4.875 $\pm$ 0.728 & 23.43 $\pm$ 0.07\\
        & 12 & 0.914 $\pm$ 0.01 & 32.33 $\pm$ 0.06 & 12 & 0.803 $\pm$ 0.02 & 32.94 $\pm$ 0.10 & 32 & 5.980 $\pm$ 0.219 & 22.97 $\pm$ 0.08\\
    \bottomrule
    \end{tabular}}
    \caption{Wall-clock time (averaged over multiple runs) for IWSLT/WMT for different model types and encoder depths. 95\% Max BLEU is for validation.}
    \label{tab:wt-time-95}
\end{table*}

\begin{table*}[t]
    \centering
    \small
    \resizebox{\textwidth}{!}{
    \begin{tabular}{lcccccccccc}
    \toprule
        \multirow{3}{*}\textbf{Model} & & \textbf{IWSLT-Dec2} & & & \textbf{IWSLT-Dec6} & & & \textbf{WMT-Dec1} & &  
        \\ & \textbf{\# Layers} & \textbf{Train time} & \textbf{ Max BLEU}  & \textbf{\# Layers} & \textbf{Train time} & \textbf{Max BLEU} & \textbf{\# Layers} & \textbf{Train time} & \textbf{Max BLEU}  \\
        &&\textbf{until 99\% max }(in hours)&(99\%)&&\textbf{until 99\% max }(in hours) &(99\%) & & \textbf{until 99\% max }(in hours) & (99\%)
 \\\midrule
        \multirow{4}{*}{Transformer} 
        & 6  & 1.454 $\pm$ 0.06 & 34.24 $\pm$ 0.05 & 6  & 1.297 $\pm$ 0.03 & 34.69 $\pm$ 0.05 & 12 &  9.961 $\pm$ 0.053 & 24.27 $\pm$ 0.04\\
        & 8  & 1.475 $\pm$ 0.09 & 34.32 $\pm$ 0.09 & 8  & 1.390 $\pm$ 0.02 & 34.75 $\pm$ 0.09 & 16 & 12.623 $\pm$ 0.072 & 24.35 $\pm$ 0.06\\
        & 10 & 1.526 $\pm$ 0.04 & 34.25 $\pm$ 0.04 & 10 & 1.622 $\pm$ 0.05 & 34.64 $\pm$ 0.03 & 24 & 13.412 $\pm$ 0.837 & 24.49 $\pm$ 0.07\\
        & 12 & 2.259 $\pm$ 0.07 & 34.24 $\pm$ 0.11 & 12 & 1.748 $\pm$ 0.01 & 34.66 $\pm$ 0.08 & 32 & 15.117 $\pm$ 0.232 & 24.56 $\pm$ 0.02\\\midrule
        \multirow{4}{*}{T Reservoir} 
        & 6  & 1.257 $\pm$ 0.04 & 34.05 $\pm$ 0.09 & 6  & 1.291 $\pm$ 0.03 & 34.51 $\pm$ 0.10 & 12 &  8.314 $\pm$ 0.062 & 24.15 $\pm$ 0.06\\
        & 8  & 1.472 $\pm$ 0.06 & 34.47 $\pm$ 0.05 & 8  & 1.339 $\pm$ 0.03 & 34.80 $\pm$ 0.04 & 16 &  9.221 $\pm$ 0.073 & 24.41 $\pm$ 0.05\\
        & 10 & 1.530 $\pm$ 0.03 & 34.36 $\pm$ 0.02 & 10 & 1.419 $\pm$ 0.04 & 34.72 $\pm$ 0.03 & 24 & 10.413 $\pm$ 0.580 & 24.56 $\pm$ 0.03\\
        & 12 & 2.043 $\pm$ 0.05 & 34.53 $\pm$ 0.07 & 12 & 1.642 $\pm$ 0.02 & 34.87 $\pm$ 0.02 & 32 & 11.465 $\pm$ 0.227 & 24.49 $\pm$ 0.01\\\midrule
        \multirow{4}{*}{FFN Reservoir} 
        & 6  & 1.138 $\pm$ 0.03 & 34.10 $\pm$ 0.13 & 6  & 1.169 $\pm$ 0.02 & 34.71 $\pm$ 0.09 & 12 &  7.407 $\pm$ 0.087 & 24.33 $\pm$ 0.08\\
        & 8  & 1.101 $\pm$ 0.07 & 34.32 $\pm$ 0.11 & 8  & 1.201 $\pm$ 0.03 & 34.79 $\pm$ 0.08 & 16 &  9.336 $\pm$ 0.036 & 24.42 $\pm$ 0.05\\
        & 10 & 1.281 $\pm$ 0.01 & 34.36 $\pm$ 0.03 & 10 & 1.276 $\pm$ 0.03 & 34.63 $\pm$ 0.03 & 24 &  9.978 $\pm$ 0.546 & 24.91 $\pm$ 0.07\\
        & 12 & 1.785 $\pm$ 0.03 & 34.42 $\pm$ 0.06 & 12 & 1.440 $\pm$ 0.01 & 34.87 $\pm$ 0.02 & 32 & 10.524 $\pm$ 0.341 & 24.96 $\pm$ 0.01\\\midrule
        \multirow{4}{*}{LayerDrop} 
        & 6  & 1.363 $\pm$ 0.05 & 34.58 $\pm$ 0.14 & 6  & 1.253 $\pm$ 0.01 & 34.42 $\pm$ 0.10 & 12 &  8.372 $\pm$ 0.059 & 24.17 $\pm$ 0.04\\
        & 8  & 1.468 $\pm$ 0.03 & 34.50 $\pm$ 0.12 & 8  & 1.244 $\pm$ 0.04 & 34.44 $\pm$ 0.09 & 16 &  9.741 $\pm$ 0.043 & 23.93 $\pm$ 0.08\\
        & 10 & 1.678 $\pm$ 0.04 & 34.52 $\pm$ 0.07 & 10 & 1.343 $\pm$ 0.04 & 33.83 $\pm$ 0.06 & 24 & 10.145 $\pm$ 0.628 & 24.07 $\pm$ 0.09\\
        & 12 & 2.071 $\pm$ 0.02 & 33.45 $\pm$ 0.23 & 12 & 1.423 $\pm$ 0.02 & 33.97 $\pm$ 0.12 & 32 & 10.168 $\pm$ 0.329 & 23.81 $\pm$ 0.03\\
    \bottomrule
    \end{tabular}}
    \caption{Wall-clock time (averaged over multiple runs) saved for IWSLT/WMT for different model types and encoder depths. 99\% Max BLEU is for validation.}
    \label{tab:wt-time-99}
\end{table*}

\begin{figure}[h]
    \centering
    \includegraphics[width=0.5\textwidth]{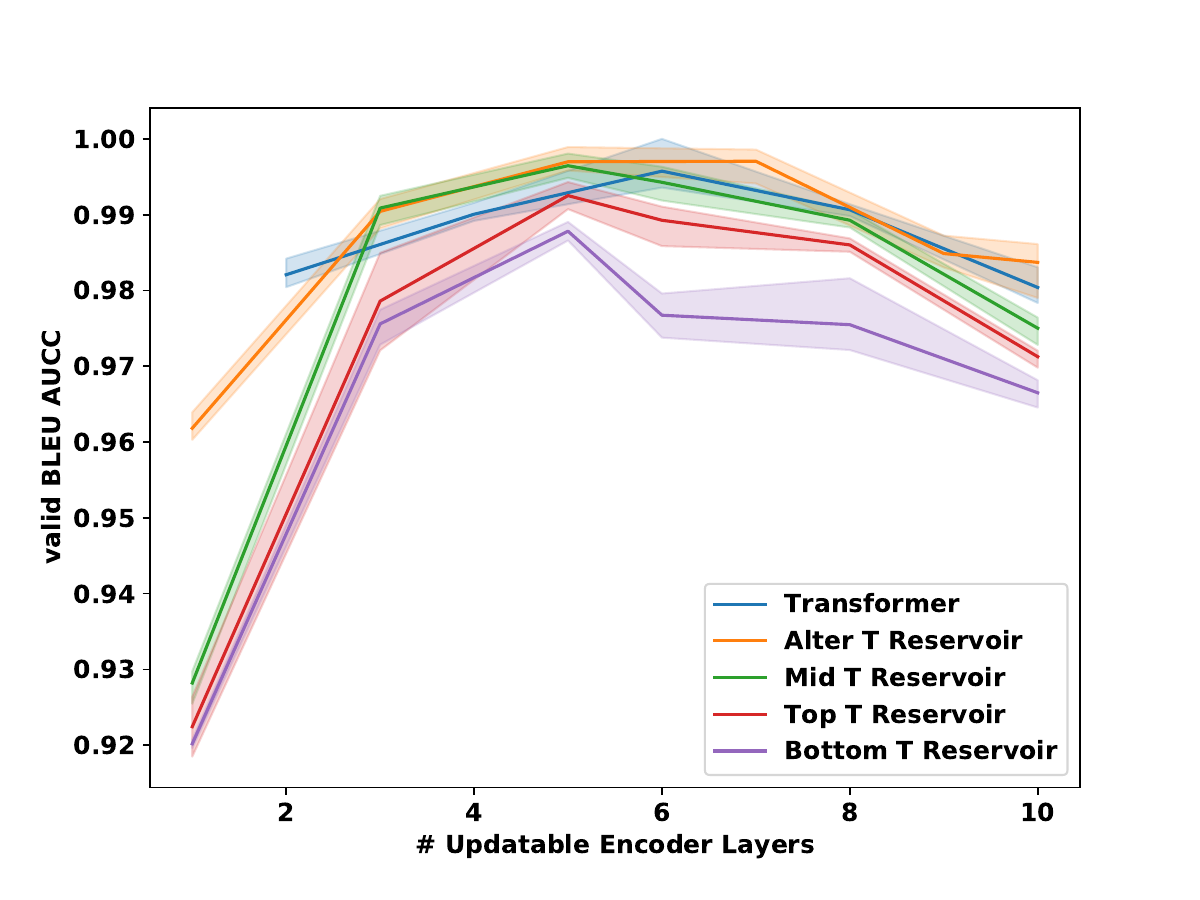}
    \caption{IWSLT with 2-layer decoder using different freezing strategies. }
    \label{fig:iwslt-freezing}
\end{figure}

\section{Freezing Strategy}
\label{appendix:freezing strategy}

We explored different strategies for the placement of reservoir layers and found the ``alternating'' strategy reported in the main body of the paper to work best. Generally, we found repetitive application of reservoirs to yield diminishing returns, as might be expected. See Figure \ref{fig:iwslt-freezing}.

\section{RoBERTa Results}
\begin{figure}[h]
    \centering
    \begin{subfigure}{.25\textwidth}
        \centering
        \includegraphics[width=\textwidth]{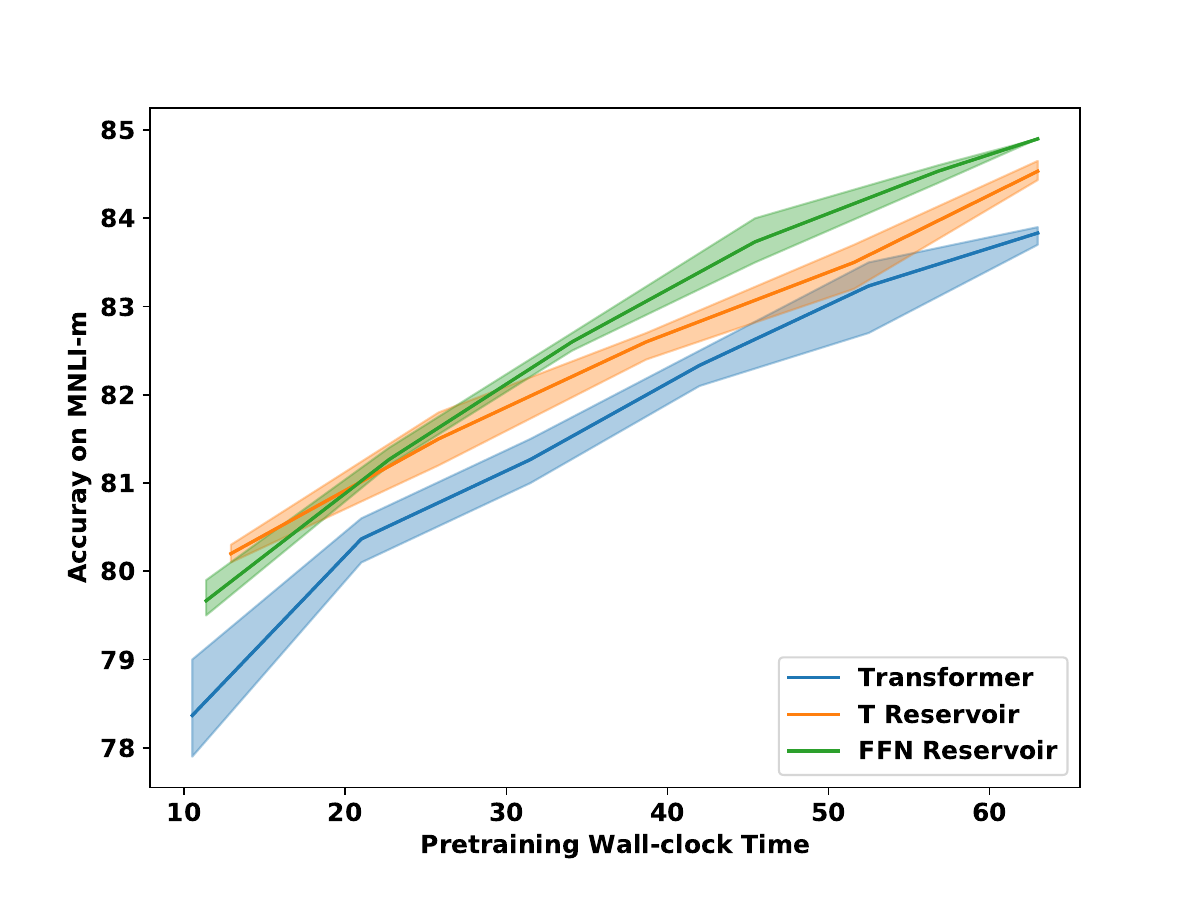}
    \end{subfigure}%
    \begin{subfigure}{.25\textwidth}
        \centering
        \includegraphics[width=\textwidth]{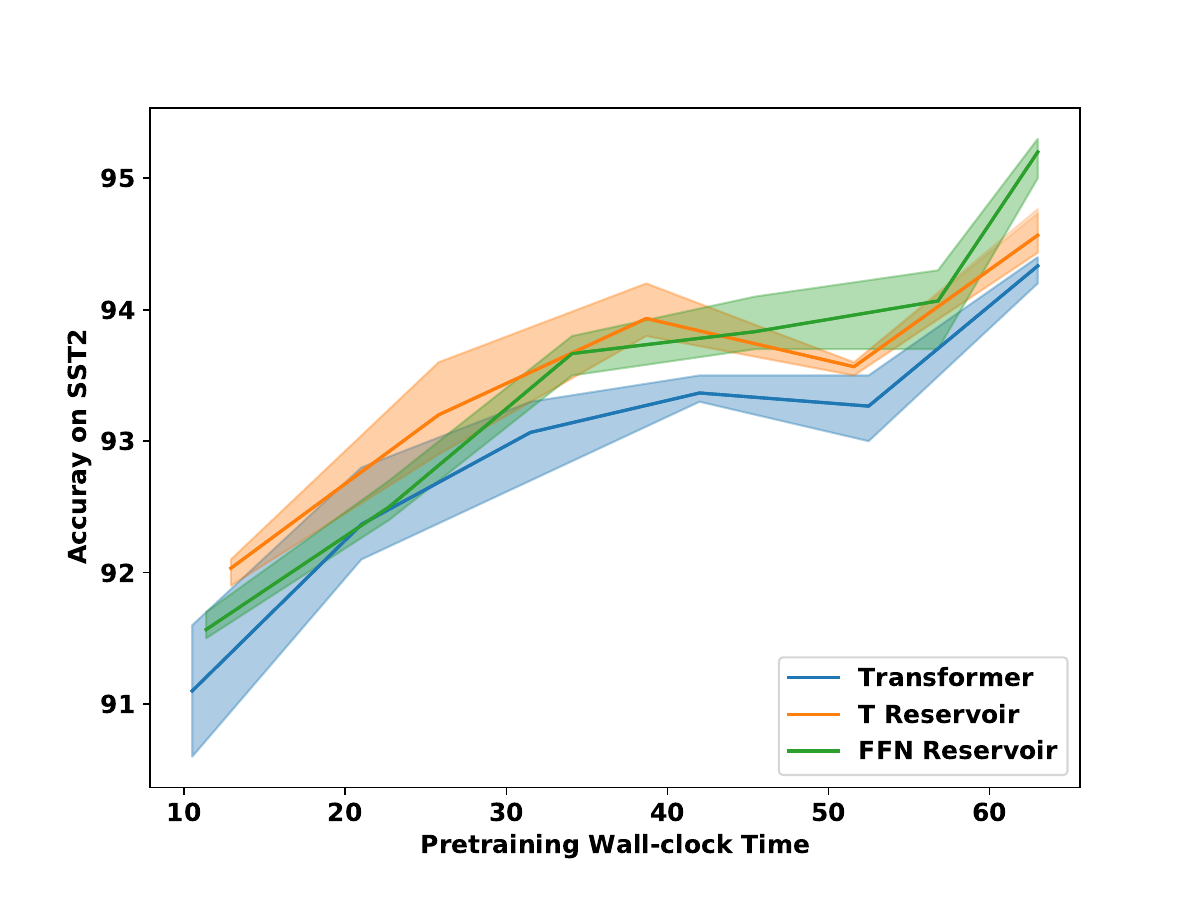}
    \end{subfigure}
    \caption{RoBERTa Reservoir Results, Pre-training versus downstream task plot for 12 layer RoBERTa. MNLI-m (left). SST-2 (right).}
    \label{fig:roberta-pretrain-glue-plot}
\end{figure}

\begin{figure}[h]
    \centering
    \begin{subfigure}{.25\textwidth}
        \centering
        \includegraphics[width=\textwidth]{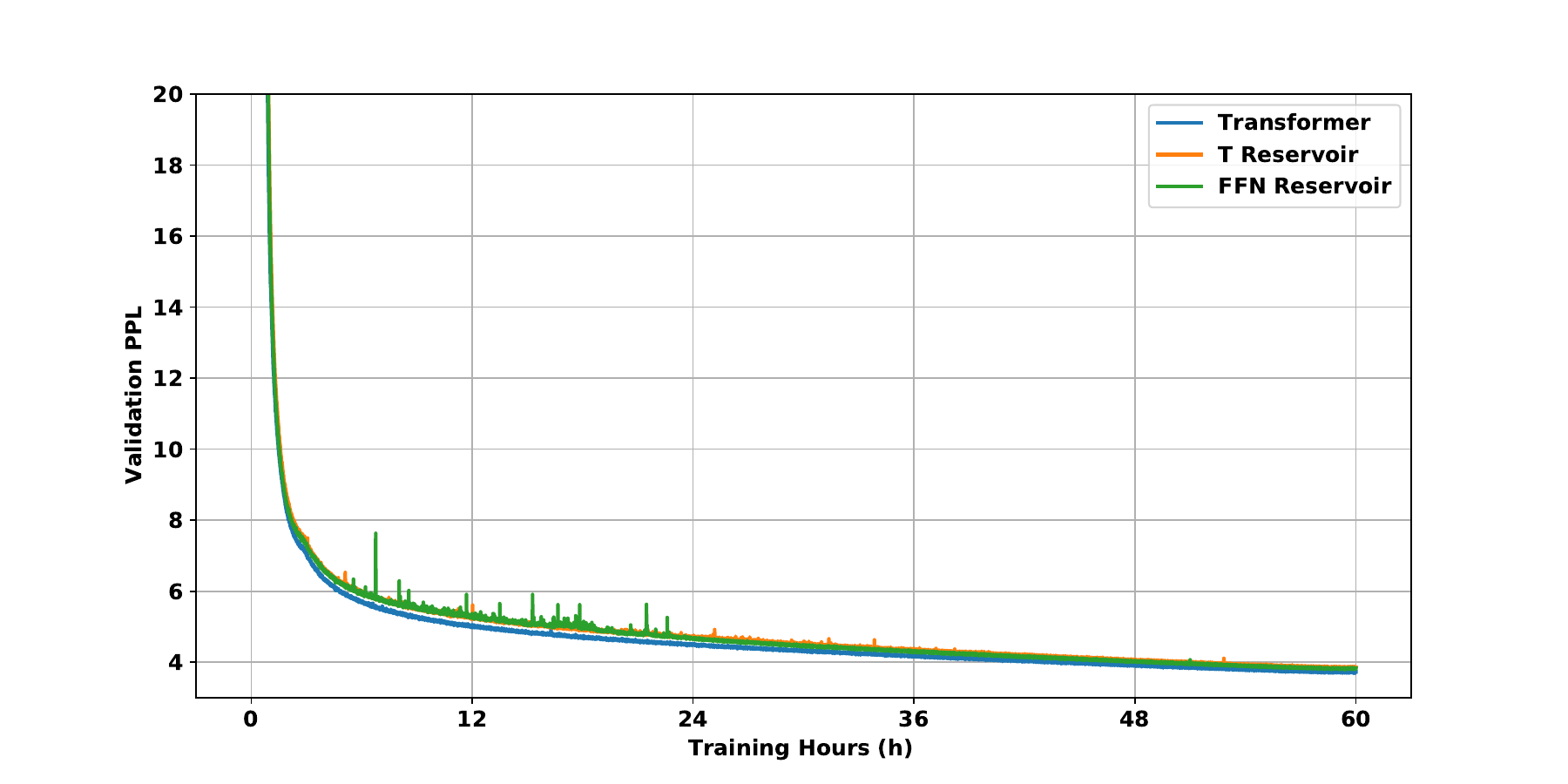}
    \end{subfigure}%
    \begin{subfigure}{.25\textwidth}
        \centering
        \includegraphics[width=\textwidth]{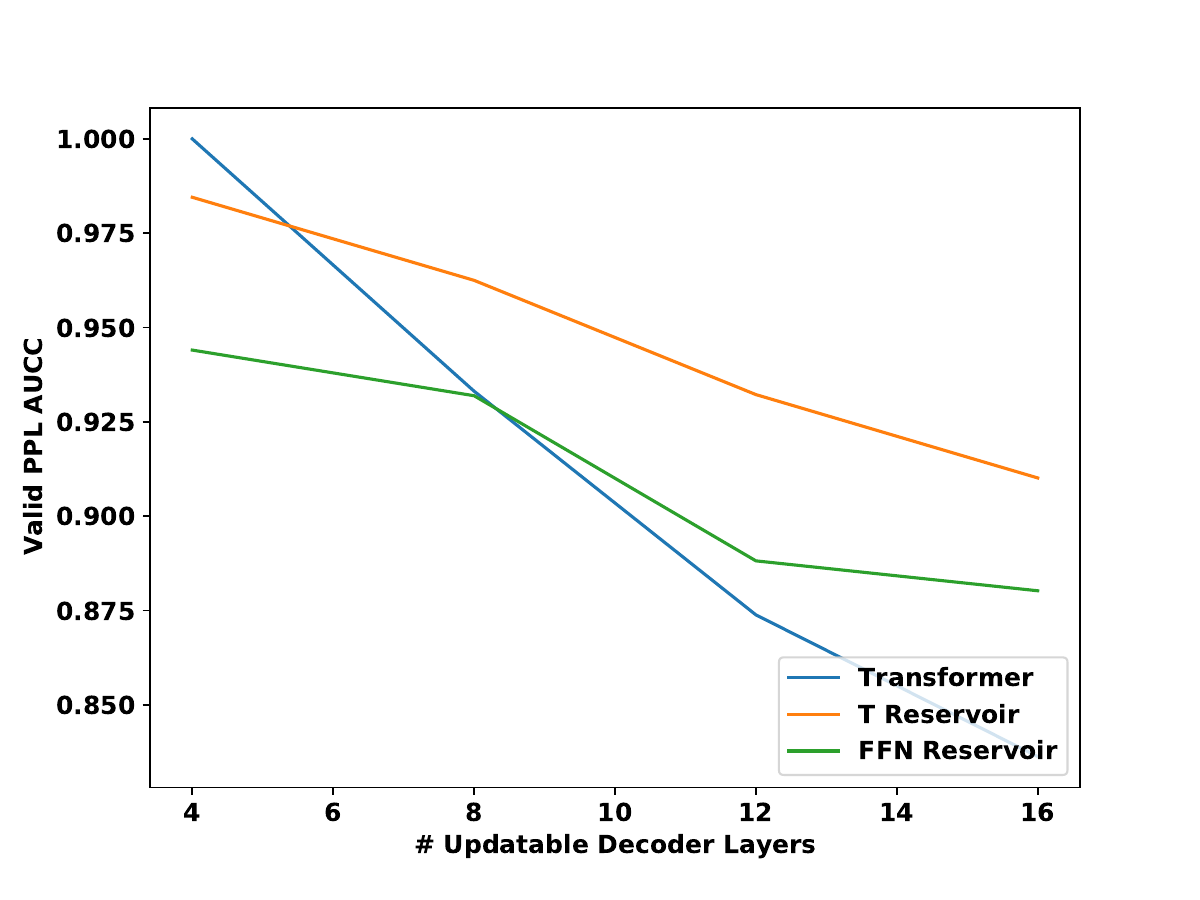}
    \end{subfigure}
    \caption{RoBERTa Reservoir Results, Training plot for 12 layer RoBERTa (left). AUCC result (right).}
    \label{fig:roberta-train-plot}
\end{figure}

Here we present the additional results for RoBERTa , i.e., convergence plots and AUCCs for various depth settings, in Figure \ref{fig:roberta-train-plot}. 
As stated in the main paper, the differences in terms of AUCC and convergence between RoBERTa models with and without reservoir layers are limited. 
Moreover, we plot downstream task performance for SST-2 and MNLI compared to the pretraining wall-clock time in Figure \ref{fig:roberta-pretrain-glue-plot}. It can be seen that the FFN Reservoir can achieve up to 25\% and 10\% pretraining time savings while matching the best performance of vanilla transformers for MNLI-m and SST2, respectively. 
\label{appendix:roberta_result}

\section{Reservoir Results for Total Layers}
\label{appendix:total_layer}
\begin{figure}[t]
    \centering
    \begin{subfigure}{.25\textwidth}
        \centering
        \includegraphics[width=\textwidth]{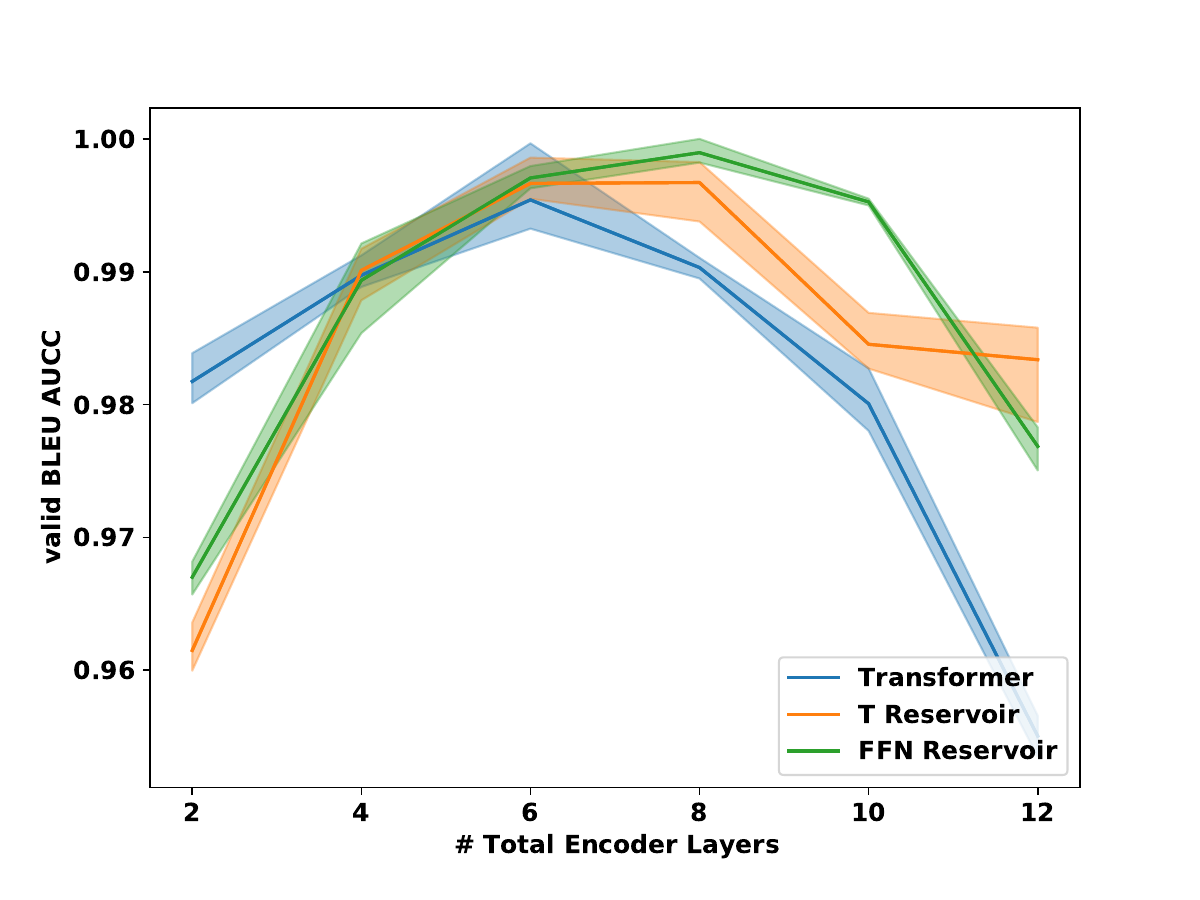}
    \end{subfigure}%
    \begin{subfigure}{.25\textwidth}
        \centering
        \includegraphics[width=\textwidth]{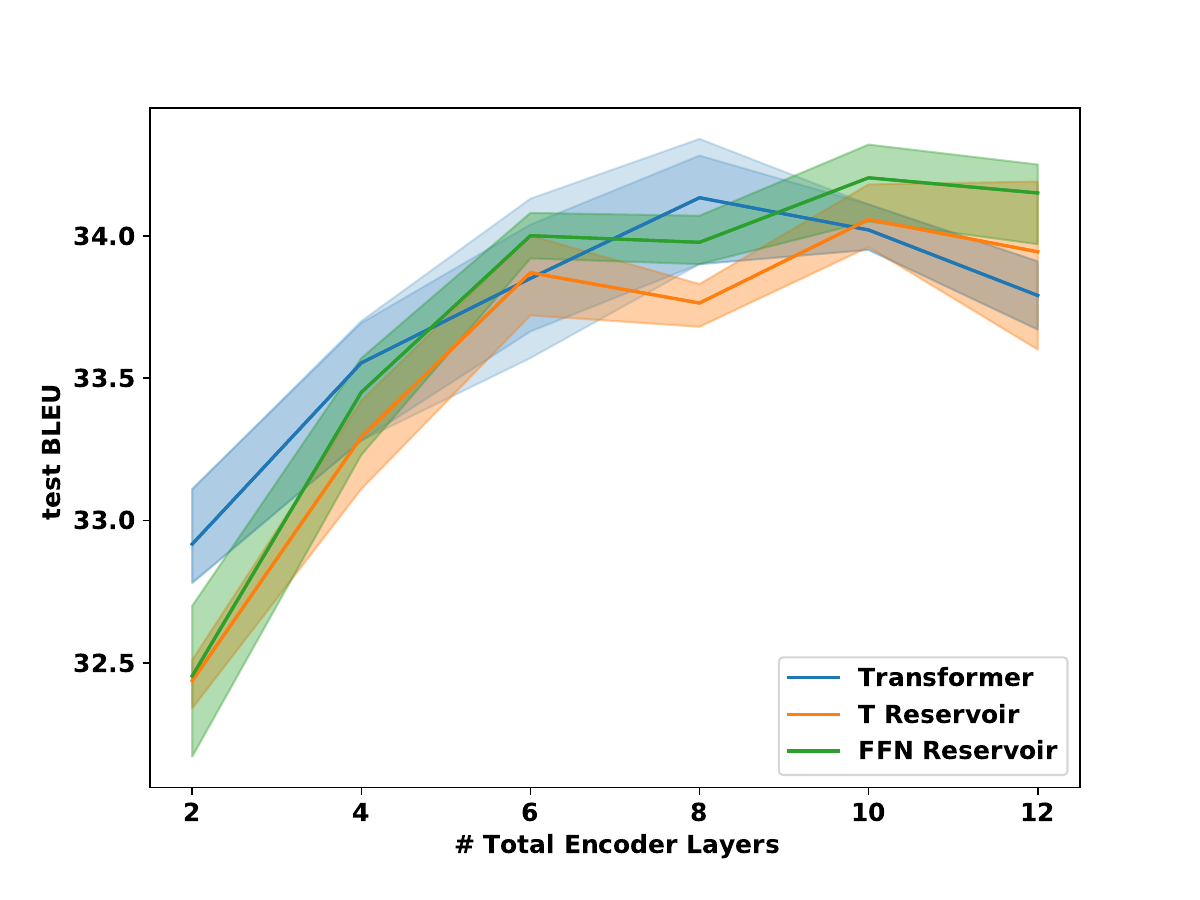}
    \end{subfigure}
    \caption{Validation BLEU AUCC and test BLEU for IWSLT (high is good). Comparison of regular transformer and reservoir transformer with FFN or Transformer reservoir layers added.
    }
    \label{fig:iwslt_total}
\end{figure}

\begin{figure}[t]
    \centering
    \begin{subfigure}{.25\textwidth}
        \centering
        \includegraphics[width=\textwidth]{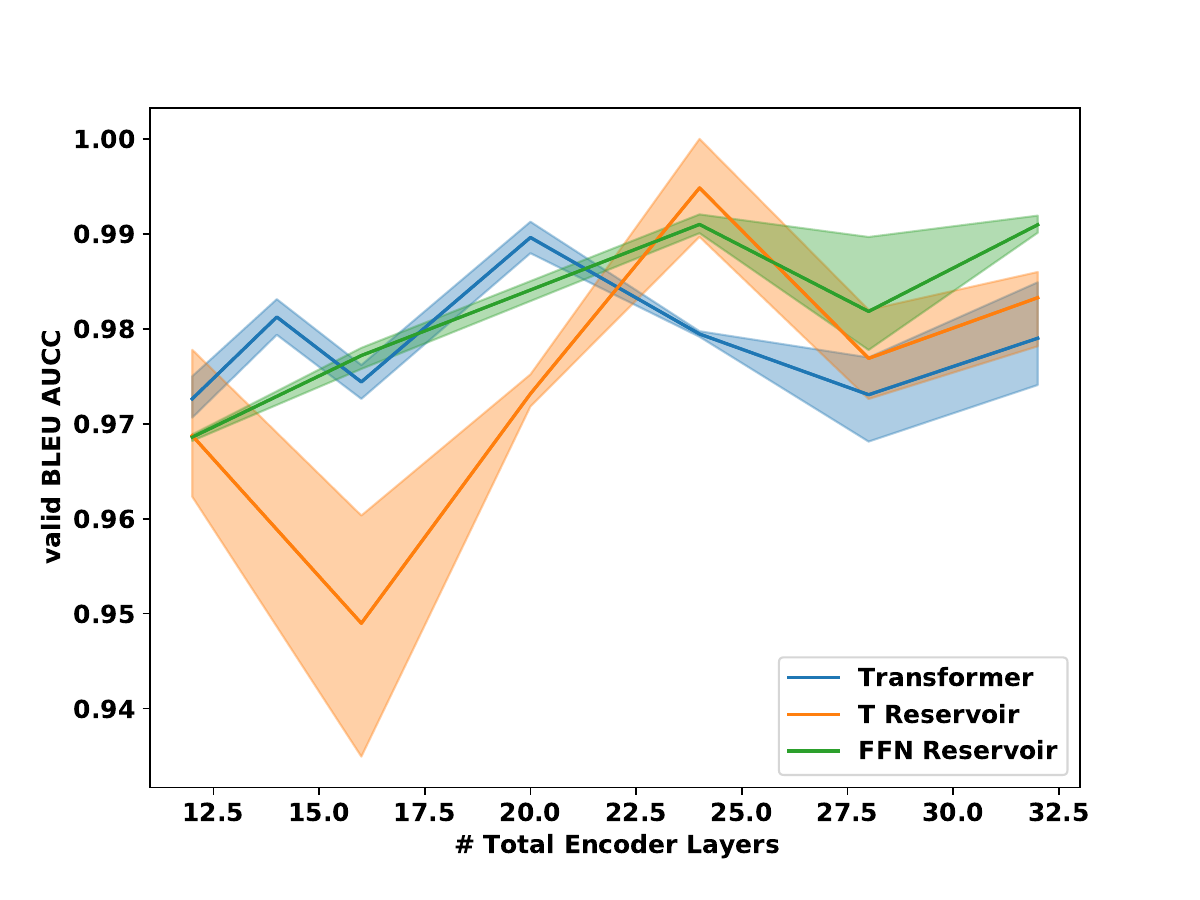}
    \end{subfigure}%
    \begin{subfigure}{.25\textwidth}
        \centering
        \includegraphics[width=\textwidth]{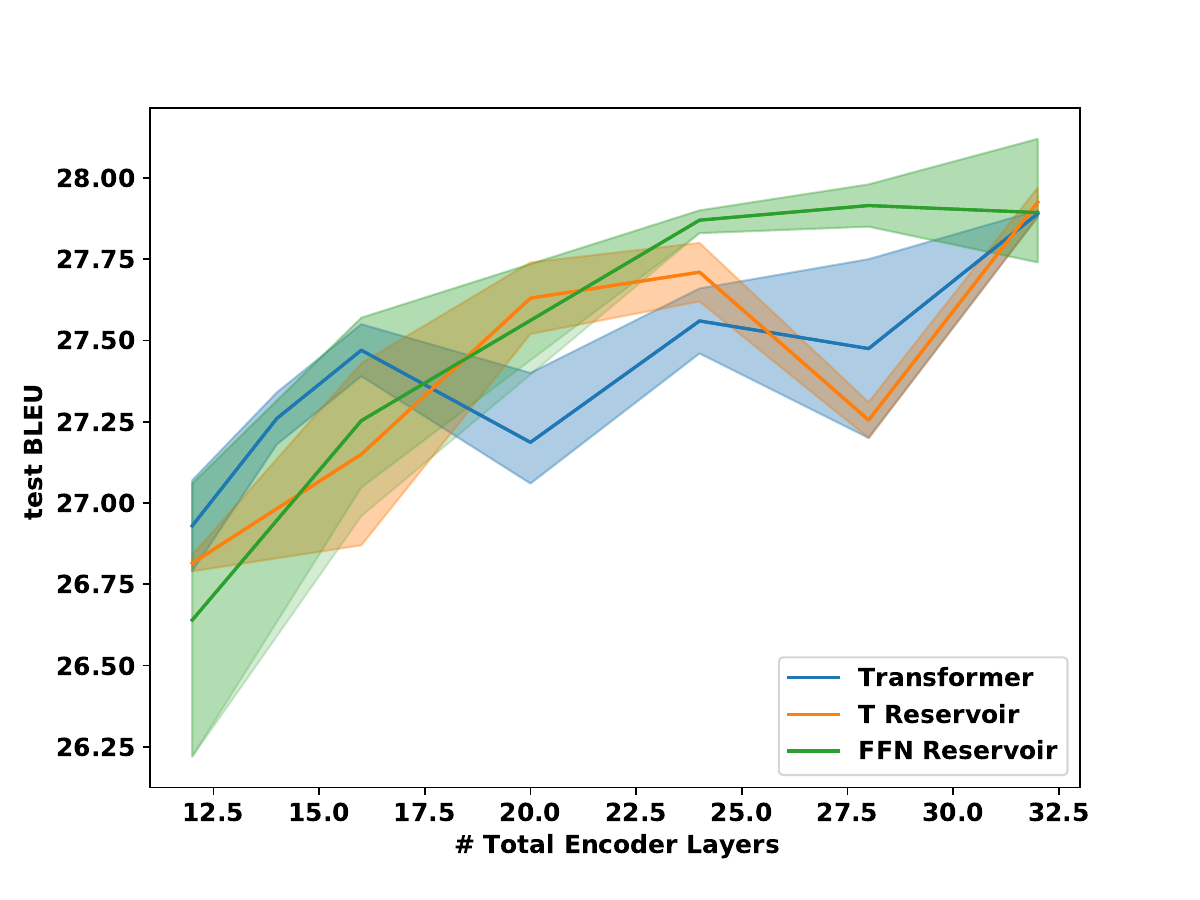}
    \end{subfigure}
    \caption{Validation BLEU AUCC and test BLEU for WMT (high is good). Comparison of regular transformer and reservoir transformer with FFN or Transformer reservoir layers added.}
    \label{fig:wmt_total}
\end{figure}

\begin{figure}[t]
    \centering
    \begin{subfigure}{.25\textwidth}
        \centering
        \includegraphics[width=\textwidth]{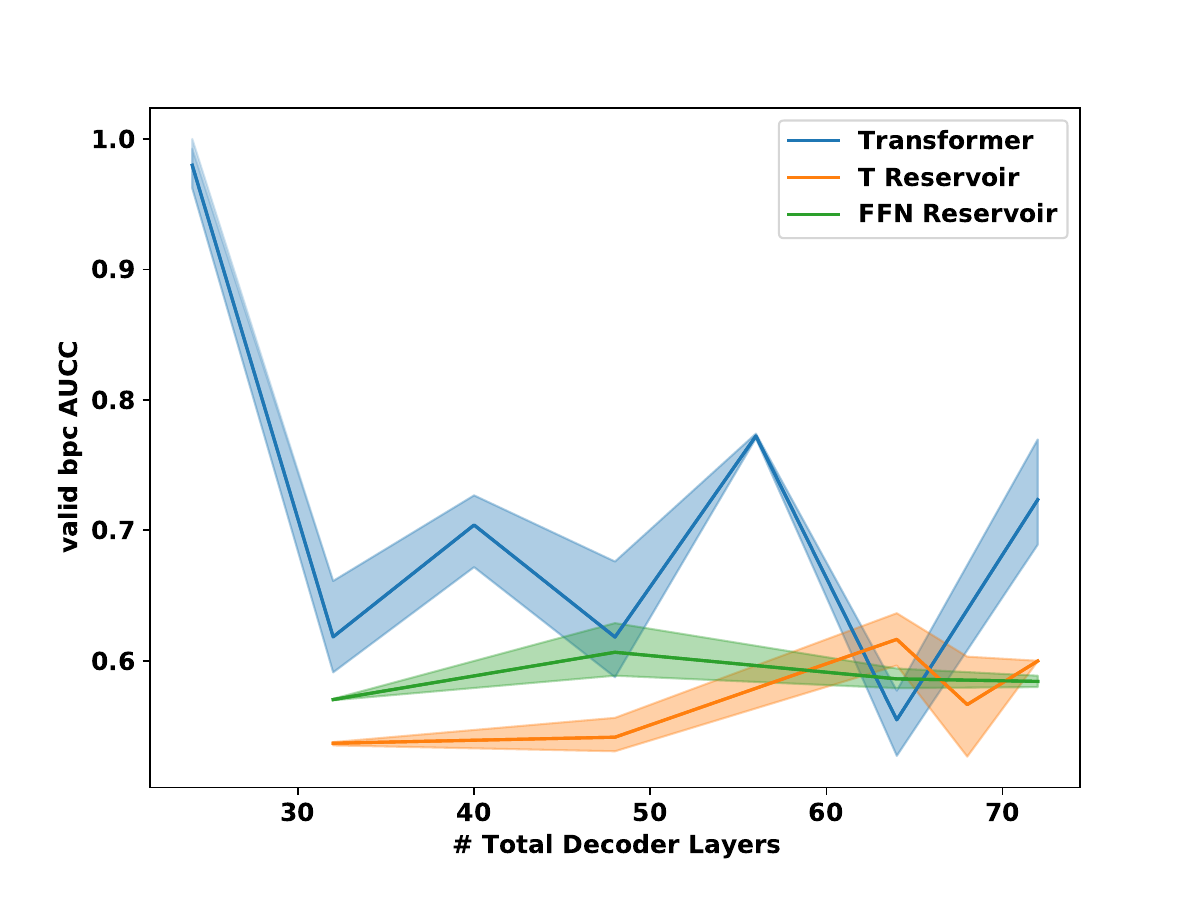}
    \end{subfigure}%
    \begin{subfigure}{.25\textwidth}
        \centering
        \includegraphics[width=\textwidth]{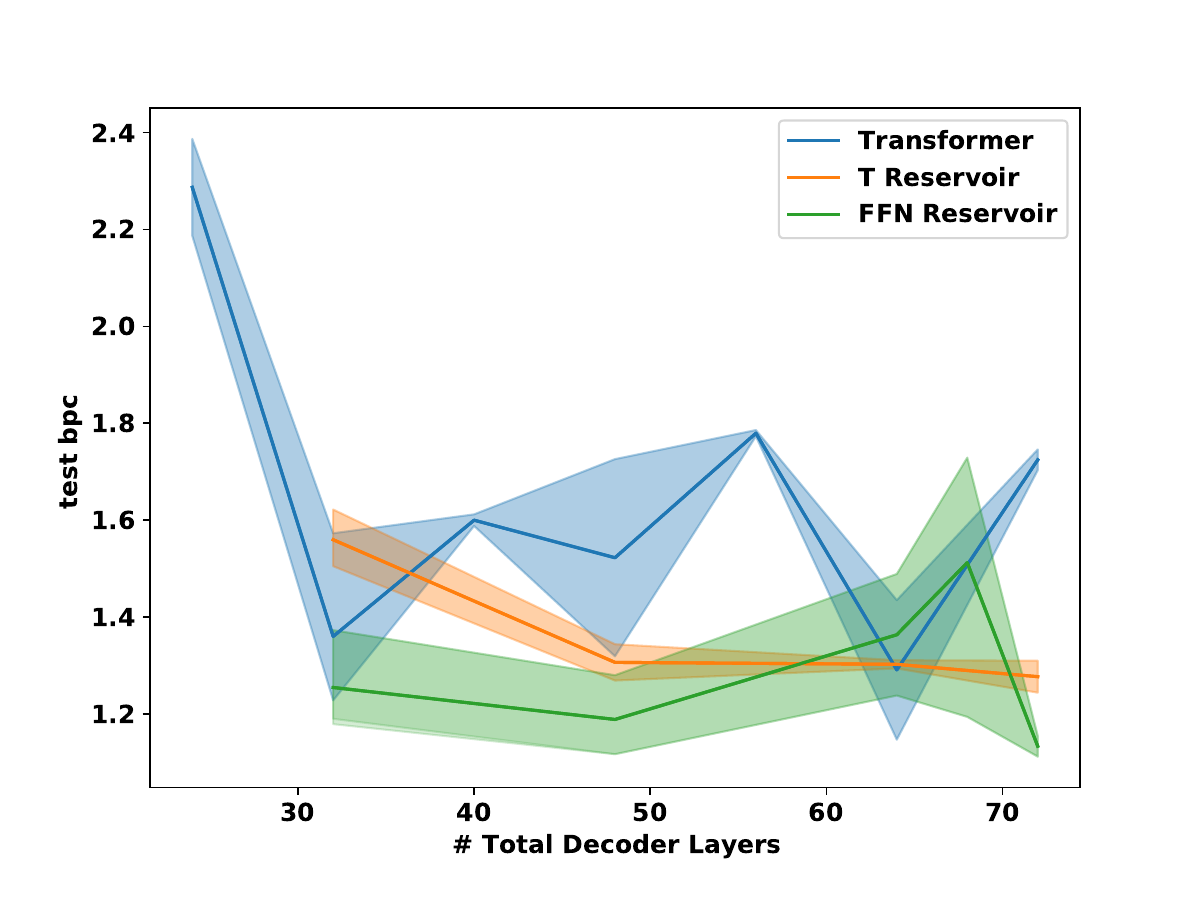}
    \end{subfigure}
    \caption{Validation BPC AUCC and test BPC on the enwik8 language modelling task (low is good). Comparison of regular and reservoir transformers for varying depths.}
    \label{fig:my_label_total}
\end{figure}

\begin{figure}
    \centering
    \begin{subfigure}{.25\textwidth}
        \centering
        \includegraphics[width=\textwidth]{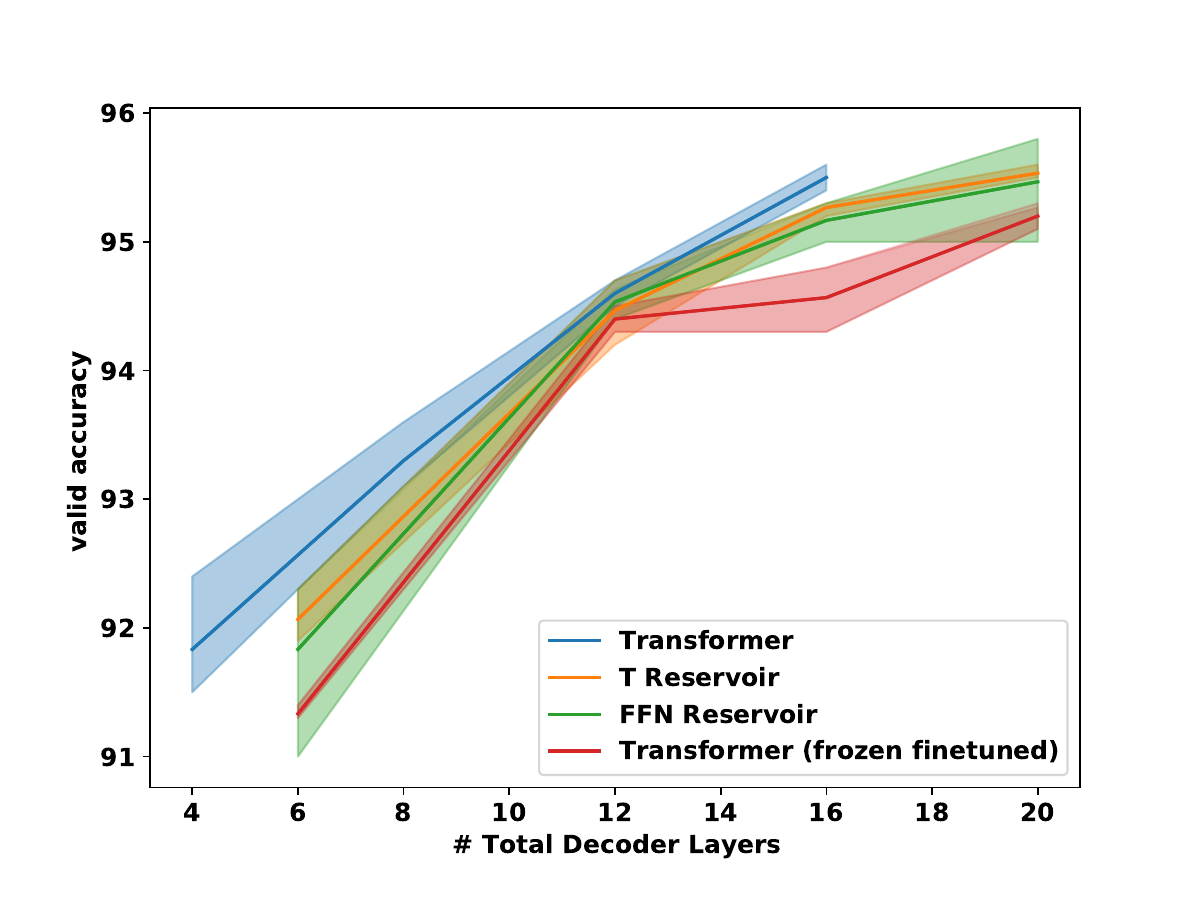}
    \end{subfigure}%
    \begin{subfigure}{.25\textwidth}
        \centering
        \includegraphics[width=\textwidth]{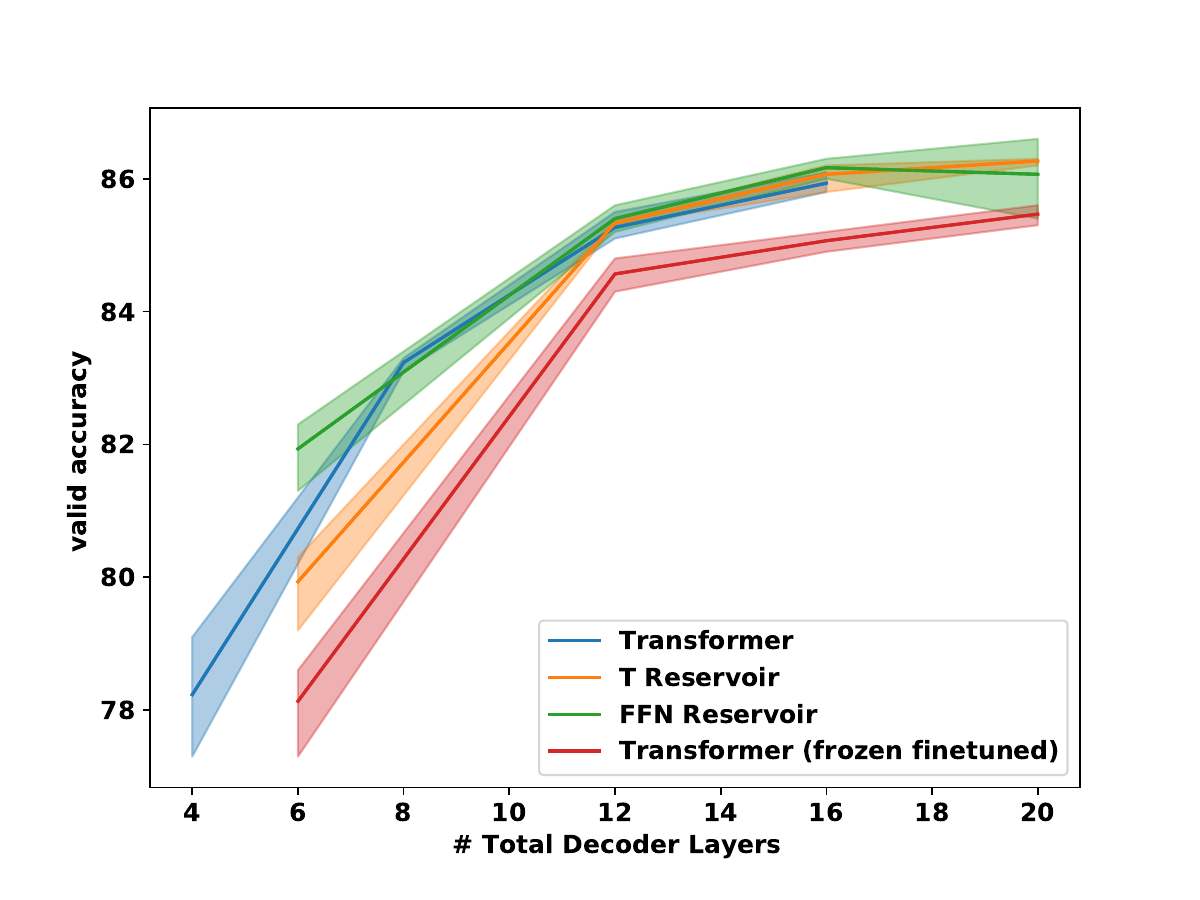}
    \end{subfigure}
    \caption{Downstream RoBERTa performance on SST-2 (left) and MultiNLI-matched (right).}
    \label{fig:finetune_total}
\end{figure}

Here we present the \textit{shifted} Reservoir Results for IWSLT14, WMT16, Enwik8 and RoBERTa finetuning in Figure \ref{fig:iwslt_total}, \ref{fig:wmt_total}, \ref{fig:my_label_total}, \ref{fig:finetune_total}, respectively. We show the same results also hold when it comes to replace normal transformer blocks with Reservoir blocks at least for MT.   

\section{Validation Plots}
\label{appendix:validation_plot}

\begin{figure}[h]
    \centering
    \begin{subfigure}{.24\textwidth}
        \centering
        \includegraphics[width=\textwidth]{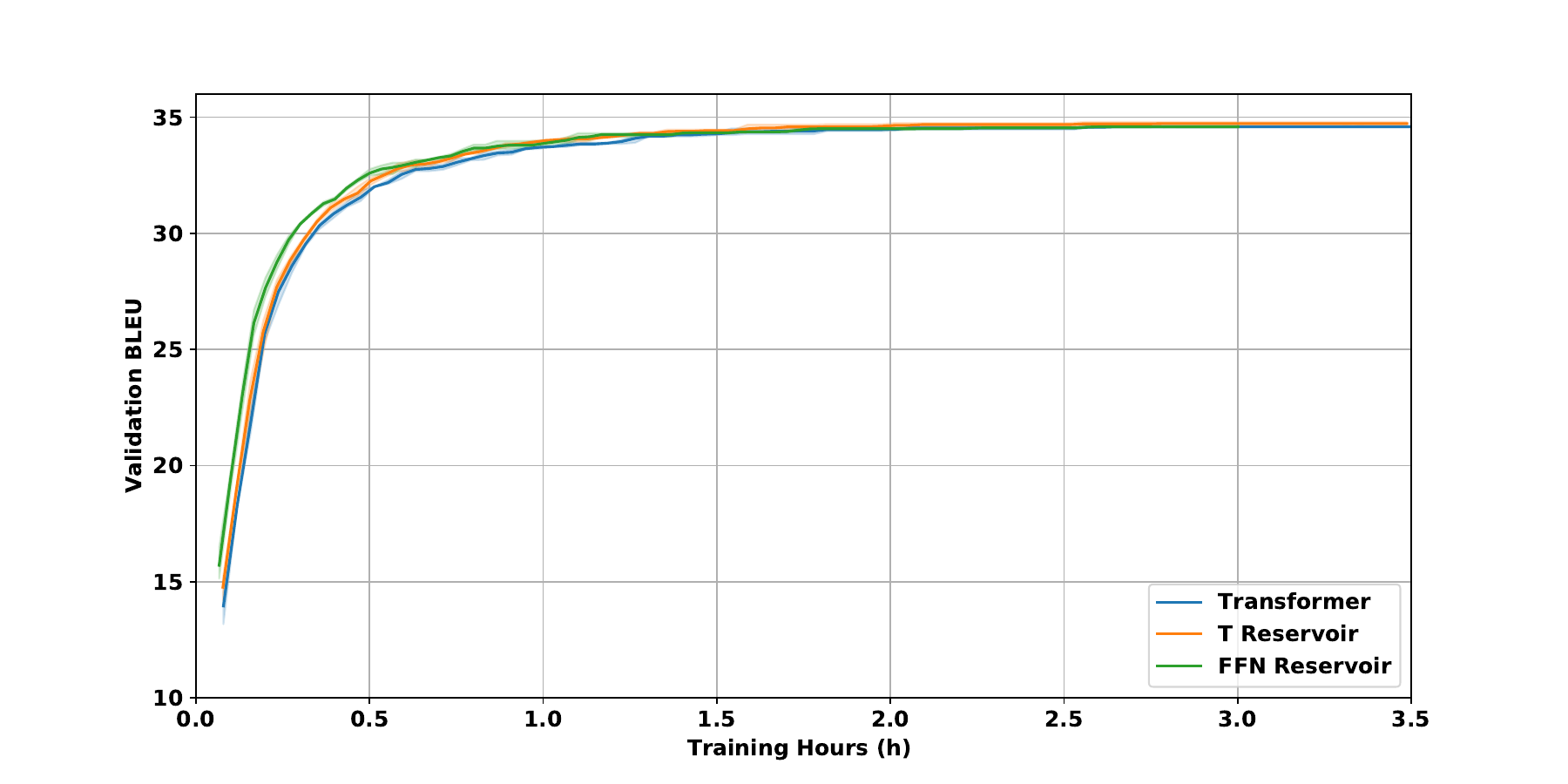}
    \end{subfigure}%
    \begin{subfigure}{.24\textwidth}
        \centering
        \includegraphics[width=\textwidth]{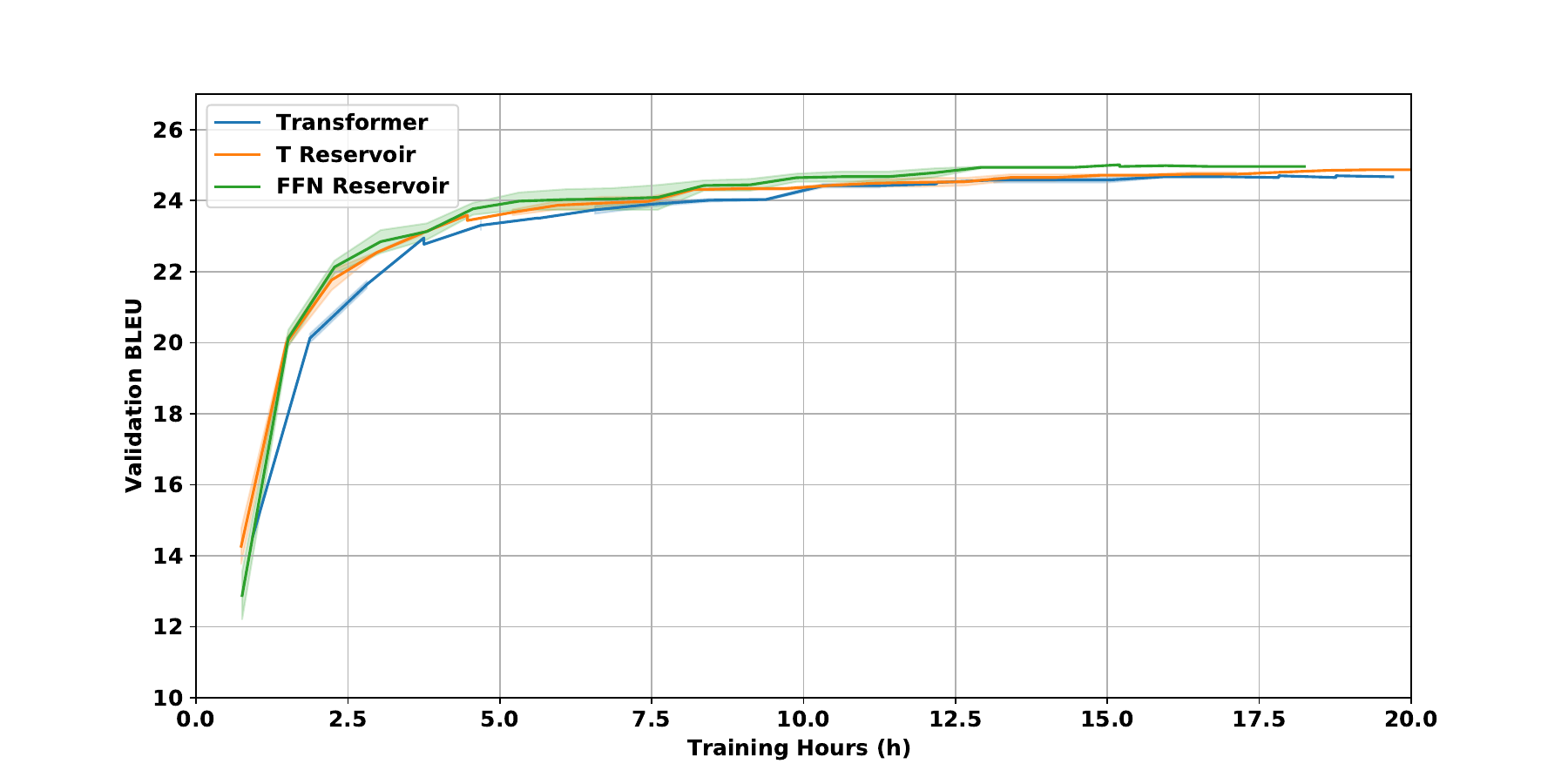}
    \end{subfigure}
    \begin{subfigure}{.23\textwidth}
        \centering
        \includegraphics[width=\textwidth]{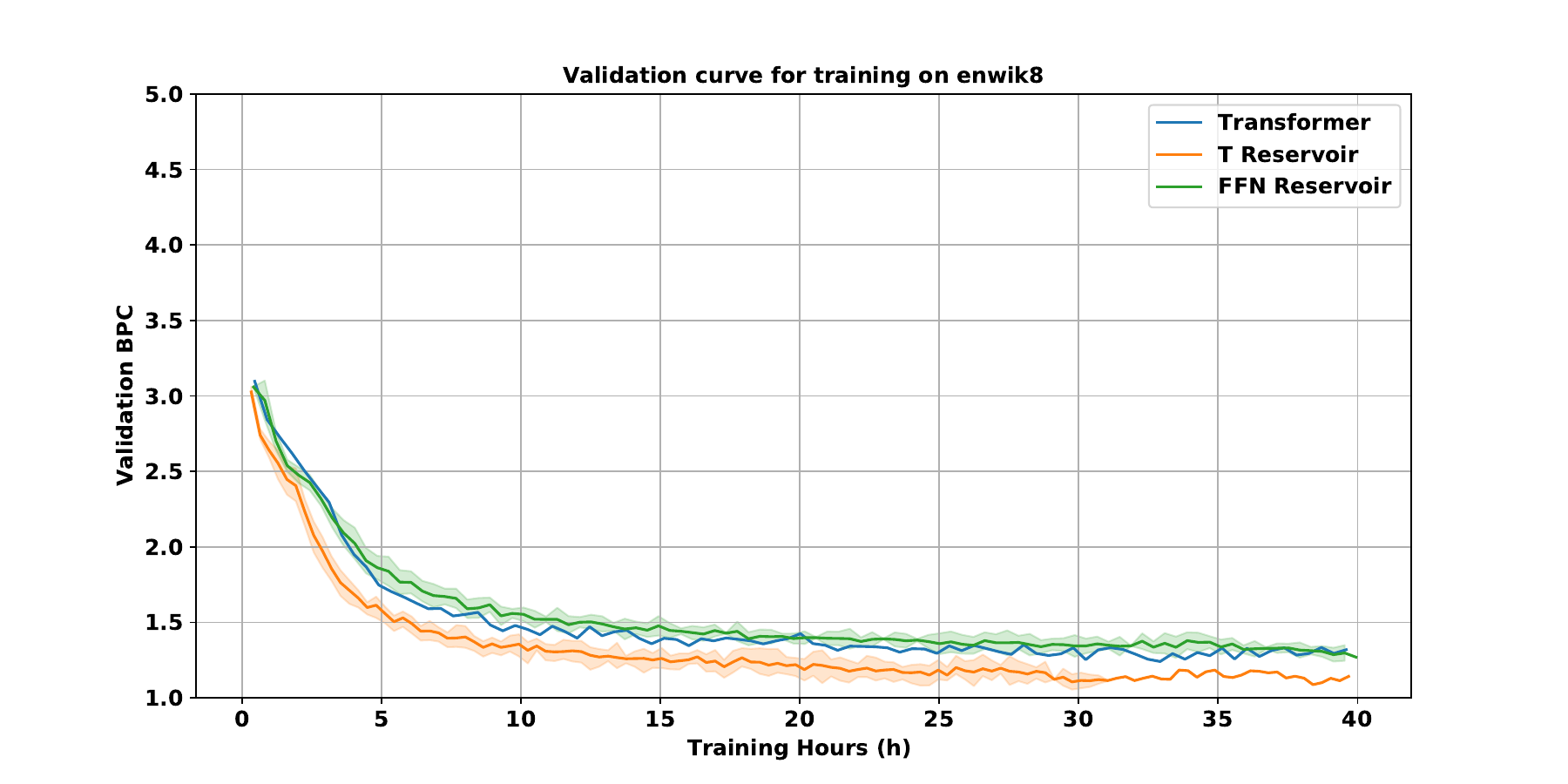}
    \end{subfigure}
    \begin{subfigure}{.23\textwidth}
        \centering
        \includegraphics[width=\textwidth]{figs/final_roberta_train_plot.pdf}
    \end{subfigure}
    \caption{IWSLT with 2-layer decoder validation plot (upper left). WMT with 24-layer decoder validation plot (upper right). Enwik8 with 48-layer decoder validation plot (lower left). RoBERTa with 12-layer decoder validation plot (lower right). }
    \label{fig:validation-train-plot}
\end{figure}

Here we present the validation plots for training a 8-layer encoder, 2-layer decoder model for IWSLT14, a 24-layer encoder, 1-layer decoder model for WMT16, a 48-layer decoder model for enwik8 and a 12-layer decoder model for RoBERTa for detailed steps to calculate the AUCC. 
It can be clearly observed that given the configurations from Section \ref{sec:experimental_details}, all the models have converged. So when we compute the area under the convergence curve, this depicts the training efficiency of the model (basically time x performance) until convergence. Specifically, we set T sufficiently high for computing the AUCC, which is 4h for IWSLT, 20h for WMT, 30h for enwik8 and 60h for RoBERTa pretraning. From the training plot in the appendix, we can see that each model has converged at that point. The Reservoir model in Figure \ref{fig:validation-train-plot} has 2 layers frozen for IWSLT14, 8 layers frozen for enwik8, and 4 layers frozen for WMT16 and RoBERTa.

\section{Backskipping}
\label{appendix:backskipping}

 \begin{figure}[t]
    \centering
    \includegraphics[width=0.5\textwidth]{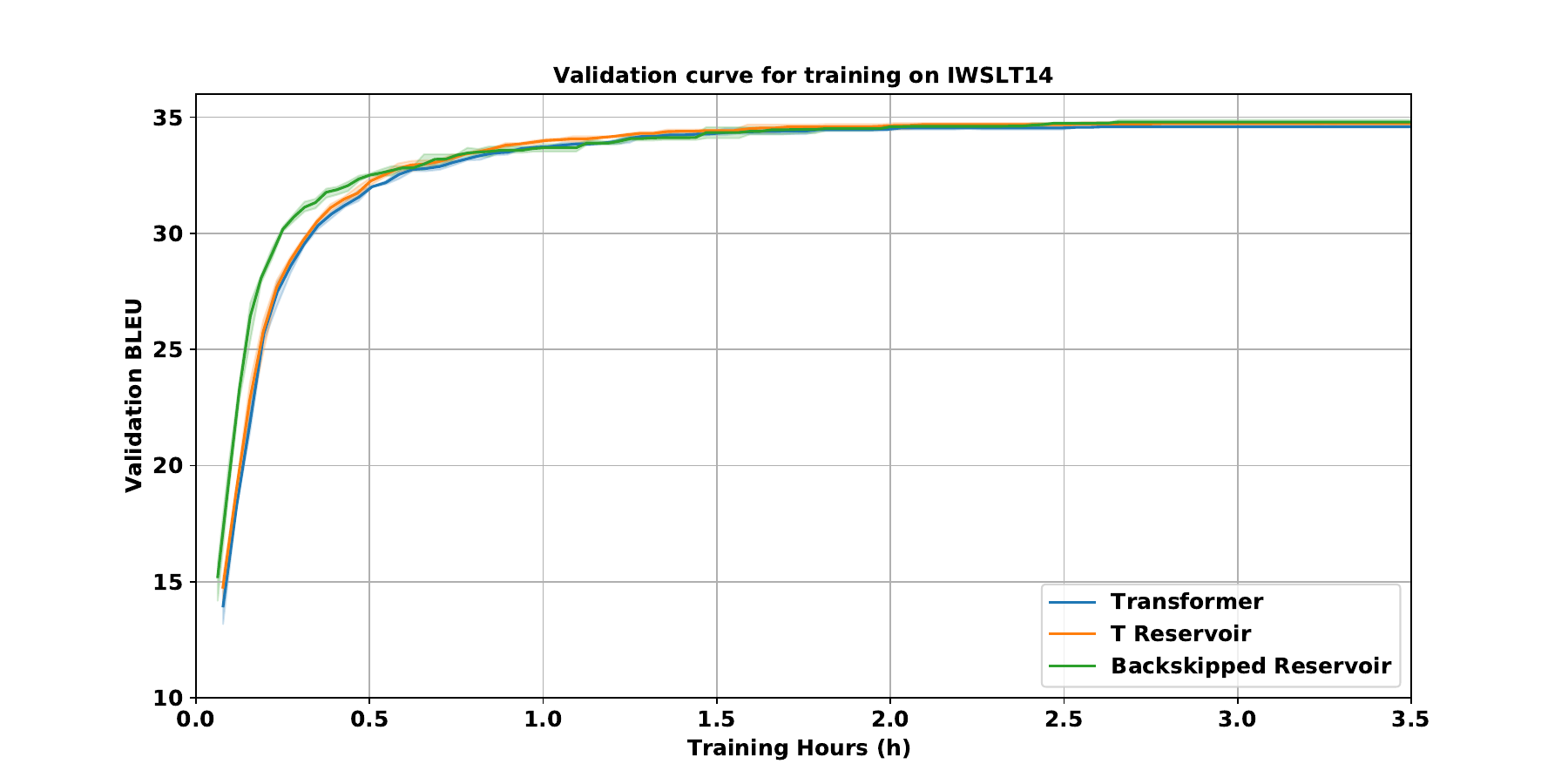}
    \caption{IWSLT comparison of the regular, reservoir and backskipped transformer architectures (encoder has 8 layers with 2 frozen, if any).
    }
    \label{fig:backskip}
\end{figure}

Figure \ref{fig:backskip} shows the BLUE curves for IWSLT comparing regular vs reservoir vs backskipped transformers, with the latter performing surprisingly well.

\end{document}

%% file: math_commands.tex
\usepackage{amsmath,amsfonts,bm}

\def\eqref#1{equation~\ref{#1}}

\def\1{\bm{1}}

\DeclareMathAlphabet{\mathsfit}{\encodingdefault}{\sfdefault}{m}{sl}
\SetMathAlphabet{\mathsfit}{bold}{\encodingdefault}{\sfdefault}{bx}{n}













%% file: acl2021.bbl
\begin{thebibliography}{101}
\expandafter\ifx\csname natexlab\endcsname\relax\def\natexlab#1{#1}\fi

\bibitem[{Ainslie et~al.(2020)Ainslie, Ontanon, Alberti, Cvicek, Fisher, Pham,
  Ravula, Sanghai, Wang, and Yang}]{ainslie-etal-2020-etc}
Joshua Ainslie, Santiago Ontanon, Chris Alberti, Vaclav Cvicek, Zachary Fisher,
  Philip Pham, Anirudh Ravula, Sumit Sanghai, Qifan Wang, and Li~Yang. 2020.
\newblock {ETC}: Encoding long and structured inputs in transformers.
\newblock In \emph{Proceedings of the 2020 Conference on Empirical Methods in
  Natural Language Processing (EMNLP)}.

\bibitem[{Bachlechner et~al.(2020)Bachlechner, Majumder, Mao, Cottrell, and
  McAuley}]{Bachlechner:2020rezero}
Thomas Bachlechner, Bodhisattwa~Prasad Majumder, Huanru~Henry Mao, Garrison~W
  Cottrell, and Julian McAuley. 2020.
\newblock Rezero is all you need: Fast convergence at large depth.
\newblock \emph{arXiv preprint arXiv:2003.04887}.

\bibitem[{Baevski et~al.(2019)Baevski, Schneider, and
  Auli}]{Baevski2019vqwav2vec}
Alexei Baevski, Steffen Schneider, and Michael Auli. 2019.
\newblock vq-wav2vec: Self-supervised learning of discrete speech
  representations.
\newblock \emph{arXiv preprint arXiv:1910.05453}.

\bibitem[{Baum(1988)}]{Baum:1988jc}
Eric~B Baum. 1988.
\newblock On the capabilities of multilayer perceptrons.
\newblock \emph{Journal of complexity}, 4(3):193--215.

\bibitem[{Beltagy et~al.(2020)Beltagy, Peters, and
  Cohan}]{Beltagy:2020longformer}
Iz~Beltagy, Matthew~E Peters, and Arman Cohan. 2020.
\newblock Longformer: The long-document transformer.
\newblock \emph{arXiv preprint arXiv:2004.05150}.

\bibitem[{Block(1962)}]{Block:1962perceptron}
Hans-Dieter Block. 1962.
\newblock The perceptron: A model for brain functioning. i.
\newblock \emph{Reviews of Modern Physics}, 34(1):123.

\bibitem[{Bojar et~al.(2014)Bojar, Buck, Federmann, Haddow, Koehn, Leveling,
  Monz, Pecina, Post, Saint-Amand, Soricut, Specia, and
  Tamchyna}]{bojar:2014-findings}
Ond{\v{r}}ej Bojar, Christian Buck, Christian Federmann, Barry Haddow, Philipp
  Koehn, Johannes Leveling, Christof Monz, Pavel Pecina, Matt Post, Herve
  Saint-Amand, Radu Soricut, Lucia Specia, and Ale{\v{s}} Tamchyna. 2014.
\newblock Findings of the 2014 workshop on statistical machine translation.
\newblock In \emph{Proceedings of the Ninth Workshop on Statistical Machine
  Translation}, Baltimore, Maryland, USA. Association for Computational
  Linguistics.

\bibitem[{Borsellino and Gamba(1961)}]{Borsellino:1961papa}
A~Borsellino and A~Gamba. 1961.
\newblock An outline of a mathematical theory of papa.
\newblock \emph{Il Nuovo Cimento (1955-1965)}, 20(2):221--231.

\bibitem[{Bradley(1997)}]{Bradley:1997roc}
Andrew~P Bradley. 1997.
\newblock The use of the area under the roc curve in the evaluation of machine
  learning algorithms.
\newblock \emph{Pattern recognition}, 30(7):1145--1159.

\bibitem[{Brock et~al.(2017)Brock, Lim, Ritchie, and
  Weston}]{Brock:2017freezeout}
Andrew Brock, Theodore Lim, James~M Ritchie, and Nick Weston. 2017.
\newblock Freezeout: Accelerate training by progressively freezing layers.
\newblock \emph{arXiv preprint arXiv:1706.04983}.

\bibitem[{Brown et~al.(2020)Brown, Mann, Ryder, Subbiah, Kaplan, Dhariwal,
  Neelakantan, Shyam, Sastry, Askell et~al.}]{Brown:2020gpt3}
Tom~B Brown, Benjamin Mann, Nick Ryder, Melanie Subbiah, Jared Kaplan, Prafulla
  Dhariwal, Arvind Neelakantan, Pranav Shyam, Girish Sastry, Amanda Askell,
  et~al. 2020.
\newblock Language models are few-shot learners.
\newblock \emph{arXiv preprint arXiv:2005.14165}.

\bibitem[{Carion et~al.(2020)Carion, Massa, Synnaeve, Usunier, Kirillov, and
  Zagoruyko}]{Carion:2020detr}
Nicolas Carion, Francisco Massa, Gabriel Synnaeve, Nicolas Usunier, Alexander
  Kirillov, and Sergey Zagoruyko. 2020.
\newblock End-to-end object detection with transformers.
\newblock \emph{arXiv preprint arXiv:2005.12872}.

\bibitem[{Cettolo et~al.(2015)Cettolo, Niehues, St{\"u}ker, Bentivogli, and
  Federico}]{Cettolo:2015ReportOT}
M.~Cettolo, J.~Niehues, S.~St{\"u}ker, L.~Bentivogli, and Marcello Federico.
  2015.
\newblock Report on the 11 th iwslt evaluation campaign , iwslt 2014.
\newblock In \emph{Proceedings of IWSLT}.

\bibitem[{Cho et~al.(2014)Cho, Van~Merri{\"e}nboer, Gulcehre, Bahdanau,
  Bougares, Schwenk, and Bengio}]{Cho:2014gru}
Kyunghyun Cho, Bart Van~Merri{\"e}nboer, Caglar Gulcehre, Dzmitry Bahdanau,
  Fethi Bougares, Holger Schwenk, and Yoshua Bengio. 2014.
\newblock Learning phrase representations using rnn encoder-decoder for
  statistical machine translation.
\newblock \emph{arXiv preprint arXiv:1406.1078}.

\bibitem[{Choromanski et~al.(2020)Choromanski, Likhosherstov, Dohan, Song,
  Davis, Sarlos, Belanger, Colwell, and Weller}]{Choromanski:2020performer}
Krzysztof Choromanski, Valerii Likhosherstov, David Dohan, Xingyou Song, Jared
  Davis, Tamas Sarlos, David Belanger, Lucy Colwell, and Adrian Weller. 2020.
\newblock Masked language modeling for proteins via linearly scalable
  long-context transformers.
\newblock \emph{arXiv preprint arXiv:2006.03555}.

\bibitem[{Conneau et~al.(2017)Conneau, Kiela, Schwenk, Barrault, and
  Bordes}]{Conneau:2017infersent}
Alexis Conneau, Douwe Kiela, Holger Schwenk, Loic Barrault, and Antoine Bordes.
  2017.
\newblock Supervised learning of universal sentence representations from
  natural language inference data.
\newblock \emph{arXiv preprint arXiv:1705.02364}.

\bibitem[{Cover(1965)}]{Cover:1965theorem}
Thomas~M Cover. 1965.
\newblock Geometrical and statistical properties of systems of linear
  inequalities with applications in pattern recognition.
\newblock \emph{IEEE transactions on electronic computers}, (3):326--334.

\bibitem[{Czarnecki et~al.(2017)Czarnecki, {\'S}wirszcz, Jaderberg, Osindero,
  Vinyals, and Kavukcuoglu}]{Czarnecki:2017dni}
Wojciech~Marian Czarnecki, Grzegorz {\'S}wirszcz, Max Jaderberg, Simon
  Osindero, Oriol Vinyals, and Koray Kavukcuoglu. 2017.
\newblock Understanding synthetic gradients and decoupled neural interfaces.
\newblock \emph{arXiv preprint arXiv:1703.00522}.

\bibitem[{Dai et~al.(2019)Dai, Yang, Yang, Carbonell, Le, and
  Salakhutdinov}]{dai:2019-transformer}
Zihang Dai, Zhilin Yang, Yiming Yang, Jaime Carbonell, Quoc Le, and Ruslan
  Salakhutdinov. 2019.
\newblock Transformer-{XL}: Attentive language models beyond a fixed-length
  context.
\newblock In \emph{Proceedings of the 57th Annual Meeting of the Association
  for Computational Linguistics}, Florence, Italy. Association for
  Computational Linguistics.

\bibitem[{Daniely et~al.(2016)Daniely, Frostig, and
  Singer}]{Daniely:2016randominit}
Amit Daniely, Roy Frostig, and Yoram Singer. 2016.
\newblock Toward deeper understanding of neural networks: The power of
  initialization and a dual view on expressivity.
\newblock In \emph{Advances In Neural Information Processing Systems}, pages
  2253--2261.

\bibitem[{Devlin et~al.(2018)Devlin, Chang, Lee, and
  Toutanova}]{Devlin:2018bert}
Jacob Devlin, Ming-Wei Chang, Kenton Lee, and Kristina Toutanova. 2018.
\newblock Bert: Pre-training of deep bidirectional transformers for language
  understanding.
\newblock \emph{arXiv preprint arXiv:1810.04805}.

\bibitem[{Du et~al.(2019)Du, Lee, Li, Wang, and Zhai}]{Du2019gradient}
Simon Du, Jason Lee, Haochuan Li, Liwei Wang, and Xiyu Zhai. 2019.
\newblock Gradient descent finds global minima of deep neural networks.
\newblock In \emph{International Conference on Machine Learning}, pages
  1675--1685.

\bibitem[{Edunov et~al.(2018)Edunov, Ott, Auli, Grangier, and
  Ranzato}]{edunov:2018classical}
Sergey Edunov, Myle Ott, Michael Auli, David Grangier, and Marc{'}Aurelio
  Ranzato. 2018.
\newblock Classical structured prediction losses for sequence to sequence
  learning.
\newblock In \emph{Proceedings of the 2018 Conference of the North {A}merican
  Chapter of the Association for Computational Linguistics: Human Language
  Technologies, Volume 1 (Long Papers)}, New Orleans, Louisiana. Association
  for Computational Linguistics.

\bibitem[{Elbayad et~al.(2019)Elbayad, Gu, Grave, and Auli}]{Elbayad:2019depth}
Maha Elbayad, Jiatao Gu, Edouard Grave, and Michael Auli. 2019.
\newblock Depth-adaptive transformer.
\newblock \emph{arXiv preprint arXiv:1910.10073}.

\bibitem[{Enguehard et~al.(2019)Enguehard, Busbridge, Zhelezniak, and
  Hammerla}]{Enguehard:2019neurallanguagepriors}
Joseph Enguehard, Dan Busbridge, Vitalii Zhelezniak, and Nils Hammerla. 2019.
\newblock Neural language priors.
\newblock \emph{arXiv preprint arXiv:1910.03492}.

\bibitem[{Fan et~al.(2019)Fan, Grave, and Joulin}]{Fan:2019reducing}
Angela Fan, Edouard Grave, and Armand Joulin. 2019.
\newblock Reducing transformer depth on demand with structured dropout.
\newblock \emph{arXiv preprint arXiv:1909.11556}.

\bibitem[{Frankle and Carbin(2018)}]{Frankle:2018lottery}
Jonathan Frankle and Michael Carbin. 2018.
\newblock The lottery ticket hypothesis: Finding sparse, trainable neural
  networks.
\newblock \emph{arXiv preprint arXiv:1803.03635}.

\bibitem[{Frankle et~al.(2020)Frankle, Schwab, and
  Morcos}]{Frankle:2020batchnorm}
Jonathan Frankle, David~J Schwab, and Ari~S Morcos. 2020.
\newblock Training batchnorm and only batchnorm: On the expressive power of
  random features in cnns.
\newblock \emph{arXiv preprint arXiv:2003.00152}.

\bibitem[{Gallicchio and Micheli(2017)}]{Gallicchio:2017echo}
Claudio Gallicchio and Alessio Micheli. 2017.
\newblock Echo state property of deep reservoir computing networks.
\newblock \emph{Cognitive Computation}, 9(3):337--350.

\bibitem[{Gallicchio and Scardapane(2020)}]{Gallicchio:2020survey}
Claudio Gallicchio and Simone Scardapane. 2020.
\newblock Deep randomized neural networks.
\newblock In \emph{Recent Trends in Learning From Data}, pages 43--68.
  Springer.

\bibitem[{Gamba et~al.(1961)Gamba, Gamberini, Palmieri, and
  Sanna}]{Gamba:1961papa}
A.~Gamba, L.~Gamberini, G.~Palmieri, and R.~Sanna. 1961.
\newblock Further experiments with papa.
\newblock \emph{Il Nuovo Cimento (1955-1965)}, 20(2):112--115.

\bibitem[{Garg et~al.(2020)Garg, Cao, and Ge}]{Garg:2020echostatenmt}
Ankush Garg, Yuan Cao, and Qi~Ge. 2020.
\newblock Echo state neural machine translation.
\newblock \emph{arXiv preprint arXiv:2002.11847}.

\bibitem[{Giryes et~al.(2016)Giryes, Sapiro, and
  Bronstein}]{Giryes:2016metriclearning}
Raja Giryes, Guillermo Sapiro, and Alex~M Bronstein. 2016.
\newblock Deep neural networks with random gaussian weights: A universal
  classification strategy?
\newblock \emph{IEEE Transactions on Signal Processing}, 64(13):3444--3457.

\bibitem[{Glorot and Bengio(2010)}]{Glorot:2010init}
Xavier Glorot and Yoshua Bengio. 2010.
\newblock Understanding the difficulty of training deep feedforward neural
  networks.
\newblock In \emph{Proceedings of the thirteenth international conference on
  artificial intelligence and statistics}, pages 249--256.

\bibitem[{Gulcehre et~al.(2016)Gulcehre, Moczulski, Denil, and
  Bengio}]{Gulcehre:2016noisy}
Caglar Gulcehre, Marcin Moczulski, Misha Denil, and Yoshua Bengio. 2016.
\newblock Noisy activation functions.
\newblock In \emph{International conference on machine learning}, pages
  3059--3068.

\bibitem[{Hadaeghi et~al.(2017)Hadaeghi, He, and
  Jaeger}]{Hadaeghi:2017hardware}
Fatemeh Hadaeghi, Xu~He, and Herbert Jaeger. 2017.
\newblock \emph{Unconventional Information Processing Systems, Novel Hardware:
  A Tour D'Horizon}.

\bibitem[{He et~al.(2021)He, Li, Soltanolkotabi, and
  Avestimehr}]{he2021pipetransformer}
Chaoyang He, Shen Li, Mahdi Soltanolkotabi, and Salman Avestimehr. 2021.
\newblock Pipetransformer: Automated elastic pipelining for distributed
  training of transformers.
\newblock In \emph{ICML}.

\bibitem[{Hicke et~al.(2013)Hicke, Escalona-Moran, Brunner, Soriano, Fischer,
  and Mirasso}]{Hicke:2013semiconductor}
Konstantin Hicke, Miguel Escalona-Moran, Daniel Brunner, Miguel Soriano, Ingo
  Fischer, and Claudio Mirasso. 2013.
\newblock \href {https://doi.org/10.1109/JSTQE.2013.2241738} {Information
  processing using transient dynamics of semiconductor lasers subject to
  delayed feedback}.
\newblock \emph{Selected Topics in Quantum Electronics, IEEE Journal of},
  19:1501610--1501610.

\bibitem[{Huang et~al.(2006)Huang, Zhu, and Siew}]{Huang:2006nc}
Guang-Bin Huang, Qin-Yu Zhu, and Chee-Kheong Siew. 2006.
\newblock Extreme learning machine: theory and applications.
\newblock \emph{Neurocomputing}, 70(1-3):489--501.

\bibitem[{Jaderberg et~al.(2017)Jaderberg, Czarnecki, Osindero, Vinyals,
  Graves, Silver, and Kavukcuoglu}]{Jaderberg:2017synthetic}
Max Jaderberg, Wojciech~Marian Czarnecki, Simon Osindero, Oriol Vinyals, Alex
  Graves, David Silver, and Koray Kavukcuoglu. 2017.
\newblock Decoupled neural interfaces using synthetic gradients.
\newblock In \emph{International Conference on Machine Learning}, pages
  1627--1635. PMLR.

\bibitem[{Jaeger(2003)}]{Jaeger:2003echostate}
Herbert Jaeger. 2003.
\newblock Adaptive nonlinear system identification with echo state networks.
\newblock In \emph{Advances in neural information processing systems}.

\bibitem[{Jawahar et~al.(2019)Jawahar, Sagot, and
  Seddah}]{jawahar-etal-2019-bert}
Ganesh Jawahar, Beno{\^\i}t Sagot, and Djam{\'e} Seddah. 2019.
\newblock What does {BERT} learn about the structure of language?
\newblock In \emph{Proceedings of the 57th Annual Meeting of the Association
  for Computational Linguistics}.

\bibitem[{Jim et~al.(1995)Jim, Horne, and Giles}]{Jim1995effects}
Kam Jim, Bill~G Horne, and C~Lee Giles. 1995.
\newblock Effects of noise on convergence and generalization in recurrent
  networks.
\newblock In \emph{Advances in neural information processing systems}, pages
  649--656.

\bibitem[{Jim et~al.(1996)Jim, Giles, and Horne}]{jim1996analysis}
Kam-Chuen Jim, C~Lee Giles, and Bill~G Horne. 1996.
\newblock An analysis of noise in recurrent neural networks: convergence and
  generalization.
\newblock \emph{IEEE Transactions on neural networks}, 7(6):1424--1438.

\bibitem[{Johnson and Lindenstrauss(1984)}]{Johnson:1984extensions}
William~B Johnson and Joram Lindenstrauss. 1984.
\newblock Extensions of lipschitz mappings into a hilbert space.
\newblock \emph{Contemporary mathematics}, 26(189-206):1.

\bibitem[{Kaplan et~al.(2020)Kaplan, McCandlish, Henighan, Brown, Chess, Child,
  Gray, Radford, Wu, and Amodei}]{Kaplan2020scaling}
Jared Kaplan, Sam McCandlish, Tom Henighan, Tom~B Brown, Benjamin Chess, Rewon
  Child, Scott Gray, Alec Radford, Jeffrey Wu, and Dario Amodei. 2020.
\newblock Scaling laws for neural language models.
\newblock \emph{arXiv preprint arXiv:2001.08361}.

\bibitem[{Kasai et~al.(2020)Kasai, Pappas, Peng, Cross, and
  Smith}]{Kasai:2020deepshallow}
Jungo Kasai, Nikolaos Pappas, Hao Peng, James Cross, and Noah~A Smith. 2020.
\newblock Deep encoder, shallow decoder: Reevaluating the speed-quality
  tradeoff in machine translation.
\newblock \emph{arXiv preprint arXiv:2006.10369}.

\bibitem[{Katharopoulos et~al.(2020)Katharopoulos, Vyas, Pappas, and
  Fleuret}]{Katharopoulos:2020linearattn}
Angelos Katharopoulos, Apoorv Vyas, Nikolaos Pappas, and Fran{\c{c}}ois
  Fleuret. 2020.
\newblock Transformers are rnns: Fast autoregressive transformers with linear
  attention.
\newblock \emph{arXiv preprint arXiv:2006.16236}.

\bibitem[{Kim(2014)}]{Kim2014convolutional}
Yoon Kim. 2014.
\newblock Convolutional neural networks for sentence classification.
\newblock \emph{arXiv preprint arXiv:1408.5882}.

\bibitem[{Kitaev et~al.(2020)Kitaev, Kaiser, and
  Levskaya}]{Kitaev:2020reformer}
Nikita Kitaev, {\L}ukasz Kaiser, and Anselm Levskaya. 2020.
\newblock Reformer: The efficient transformer.
\newblock \emph{arXiv preprint arXiv:2001.04451}.

\bibitem[{LeCun et~al.(1998)LeCun, Bottou, Bengio, and Haffner}]{Lecun:1998cnn}
Yann LeCun, L{\'e}on Bottou, Yoshua Bengio, and Patrick Haffner. 1998.
\newblock Gradient-based learning applied to document recognition.
\newblock \emph{Proceedings of the IEEE}, 86(11):2278--2324.

\bibitem[{Li and Liang(2018)}]{Li:2018overparameterized}
Yuanzhi Li and Yingyu Liang. 2018.
\newblock Learning overparameterized neural networks via stochastic gradient
  descent on structured data.
\newblock In \emph{Advances in Neural Information Processing Systems}, pages
  8157--8166.

\bibitem[{Li et~al.(2020)Li, Wallace, Shen, Lin, Keutzer, Klein, and
  Gonzalez}]{Li:2020traincompress}
Zhuohan Li, Eric Wallace, Sheng Shen, Kevin Lin, Kurt Keutzer, Dan Klein, and
  Joseph~E Gonzalez. 2020.
\newblock Train large, then compress: Rethinking model size for efficient
  training and inference of transformers.
\newblock \emph{arXiv preprint arXiv:2002.11794}.

\bibitem[{Liu et~al.(2019)Liu, Ott, Goyal, Du, Joshi, Chen, Levy, Lewis,
  Zettlemoyer, and Stoyanov}]{Liu:2019roberta}
Yinhan Liu, Myle Ott, Naman Goyal, Jingfei Du, Mandar Joshi, Danqi Chen, Omer
  Levy, Mike Lewis, Luke Zettlemoyer, and Veselin Stoyanov. 2019.
\newblock Roberta: A robustly optimized bert pretraining approach.
\newblock \emph{arXiv preprint arXiv:1907.11692}.

\bibitem[{LLC(2009)}]{LLC:2009long}
MultiMedia LLC. 2009.
\newblock Large text compression benchmark.

\bibitem[{Luko{\v{s}}evi{\v{c}}ius and
  Jaeger(2009)}]{Lukovsevivcius:2009reservoir}
Mantas Luko{\v{s}}evi{\v{c}}ius and Herbert Jaeger. 2009.
\newblock Reservoir computing approaches to recurrent neural network training.
\newblock \emph{Computer Science Review}, 3(3).

\bibitem[{Maass et~al.(2002)Maass, Natschl{\"a}ger, and
  Markram}]{Maass:2002lsm}
Wolfgang Maass, Thomas Natschl{\"a}ger, and Henry Markram. 2002.
\newblock Real-time computing without stable states: A new framework for neural
  computation based on perturbations.
\newblock \emph{Neural computation}, 14(11):2531--2560.

\bibitem[{Minsky and Papert(2017)}]{Minsky:2017book}
Marvin Minsky and Seymour~A Papert. 2017.
\newblock \emph{Perceptrons: An introduction to computational geometry}.
\newblock MIT press.

\bibitem[{Neftci et~al.(2017)Neftci, Augustine, Paul, and
  Detorakis}]{Neftci:2017randombackprop}
Emre~O Neftci, Charles Augustine, Somnath Paul, and Georgios Detorakis. 2017.
\newblock Event-driven random back-propagation: Enabling neuromorphic deep
  learning machines.
\newblock \emph{Frontiers in neuroscience}, 11:324.

\bibitem[{Noh et~al.(2017)Noh, You, Mun, and Han}]{Noh:2017regularizing}
Hyeonwoo Noh, Tackgeun You, Jonghwan Mun, and Bohyung Han. 2017.
\newblock Regularizing deep neural networks by noise: Its interpretation and
  optimization.
\newblock In \emph{Advances in Neural Information Processing Systems}, pages
  5109--5118.

\bibitem[{Oktay et~al.(2020)Oktay, McGreivy, Aduol, Beatson, and
  Adams}]{Oktay:2020randomizedautodiff}
Deniz Oktay, Nick McGreivy, Joshua Aduol, Alex Beatson, and Ryan~P Adams. 2020.
\newblock Randomized automatic differentiation.
\newblock \emph{arXiv preprint arXiv:2007.10412}.

\bibitem[{Ott et~al.(2018)Ott, Edunov, Grangier, and Auli}]{Ott:2018scaling}
Myle Ott, Sergey Edunov, David Grangier, and Michael Auli. 2018.
\newblock Scaling neural machine translation.
\newblock \emph{arXiv preprint arXiv:1806.00187}.

\bibitem[{Pao et~al.(1994)Pao, Park, and Sobajic}]{Pao:1994nc}
Yoh-Han Pao, Gwang-Hoon Park, and Dejan~J Sobajic. 1994.
\newblock Learning and generalization characteristics of the random vector
  functional-link net.
\newblock \emph{Neurocomputing}, 6(2):163--180.

\bibitem[{Peng et~al.(2021)Peng, Pappas, Yogatama, Schwartz, Smith, and
  Kong}]{peng2021random}
Hao Peng, Nikolaos Pappas, Dani Yogatama, Roy Schwartz, Noah Smith, and
  Lingpeng Kong. 2021.
\newblock Random feature attention.
\newblock In \emph{International Conference on Learning Representations}.

\bibitem[{Pilault et~al.(2020)Pilault, Park, and Pal}]{Pilault:2020impressive}
Jonathan Pilault, Jaehong Park, and Christopher Pal. 2020.
\newblock On the impressive performance of randomly weighted encoders in
  summarization tasks.
\newblock \emph{arXiv preprint arXiv:2002.09084}.

\bibitem[{Pons and Serra(2019)}]{Pons2019randomly}
Jordi Pons and Xavier Serra. 2019.
\newblock Randomly weighted cnns for (music) audio classification.
\newblock In \emph{ICASSP 2019-2019 IEEE international conference on acoustics,
  speech and signal processing (ICASSP)}, pages 336--340. IEEE.

\bibitem[{Press et~al.(2019)Press, Smith, and Levy}]{Press2019improving}
Ofir Press, Noah~A Smith, and Omer Levy. 2019.
\newblock Improving transformer models by reordering their sublayers.
\newblock \emph{arXiv preprint arXiv:1911.03864}.

\bibitem[{Radford et~al.(2018)Radford, Wu, Child, Luan, Amodei, and
  Sutskever}]{Radford:2019gpt2}
Alec Radford, Jeffrey Wu, Rewon Child, David Luan, Dario Amodei, and Ilya
  Sutskever. 2018.
\newblock Language models are unsupervised multitask learners.

\bibitem[{Rahimi and Recht(2008)}]{Rahimi:2008random}
Ali Rahimi and Benjamin Recht. 2008.
\newblock Random features for large-scale kernel machines.
\newblock In \emph{Advances in neural information processing systems}, pages
  1177--1184.

\bibitem[{Rahimi and Recht(2009)}]{Rahimi:2009kitchen}
Ali Rahimi and Benjamin Recht. 2009.
\newblock Weighted sums of random kitchen sinks: Replacing minimization with
  randomization in learning.
\newblock In \emph{Advances in neural information processing systems}, pages
  1313--1320.

\bibitem[{Ramanujan et~al.(2020)Ramanujan, Wortsman, Kembhavi, Farhadi, and
  Rastegari}]{Ramanujan:2020cvpr}
Vivek Ramanujan, Mitchell Wortsman, Aniruddha Kembhavi, Ali Farhadi, and
  Mohammad Rastegari. 2020.
\newblock What's hidden in a randomly weighted neural network?
\newblock In \emph{Proceedings of the IEEE/CVF Conference on Computer Vision
  and Pattern Recognition}, pages 11893--11902.

\bibitem[{Rogers et~al.(2020)Rogers, Kovaleva, and
  Rumshisky}]{Rogers2020:primer}
Anna Rogers, Olga Kovaleva, and Anna Rumshisky. 2020.
\newblock A primer in bertology: What we know about how bert works.
\newblock \emph{arXiv preprint arXiv:2002.12327}.

\bibitem[{Rosenfeld and Tsotsos(2019)}]{Rosenfeld:2019intriguing}
Amir Rosenfeld and John~K Tsotsos. 2019.
\newblock Intriguing properties of randomly weighted networks: Generalizing
  while learning next to nothing.
\newblock In \emph{2019 16th Conference on Computer and Robot Vision (CRV)},
  pages 9--16. IEEE.

\bibitem[{Sahlgren(2005)}]{Sahlgren:2005randomindexing}
Magnus Sahlgren. 2005.
\newblock An introduction to random indexing.
\newblock In \emph{Methods and applications of semantic indexing workshop at
  the 7th international conference on terminology and knowledge engineering}.

\bibitem[{Saxe et~al.(2013)Saxe, McClelland, and Ganguli}]{Saxe:2013orthogonal}
Andrew~M Saxe, James~L McClelland, and Surya Ganguli. 2013.
\newblock Exact solutions to the nonlinear dynamics of learning in deep linear
  neural networks.
\newblock \emph{arXiv preprint arXiv:1312.6120}.

\bibitem[{Scardapane and Wang(2017)}]{Scardapane:2017randomness}
Simone Scardapane and Dianhui Wang. 2017.
\newblock Randomness in neural networks: an overview.
\newblock \emph{Wiley Interdisciplinary Reviews: Data Mining and Knowledge
  Discovery}, 7(2):e1200.

\bibitem[{Schmidt et~al.(1992)Schmidt, Kraaijveld, and Duin}]{Schmidt:1992pr}
Wouter~F Schmidt, Martin~A Kraaijveld, and Robert~PW Duin. 1992.
\newblock Feedforward neural networks with random weights.
\newblock In \emph{Proceedings of the 11th International Conference on Pattern
  Recognition, 1992. Vol. II. Conference B: Pattern Recognition Methodology and
  Systems}, pages 1--4.

\bibitem[{Schrauwen et~al.(2007)Schrauwen, D'Haene, Verstraeten, and
  Campenhout}]{Schrauwen:2007lsmhardware}
Benjamin Schrauwen, Michiel D'Haene, David Verstraeten, and Jan Campenhout.
  2007.
\newblock \href {https://doi.org/10.1109/IJCNN.2007.4371111} {Compact hardware
  for real-time speech recognition using a liquid state machine}.
\newblock pages 1097 -- 1102.

\bibitem[{Schwartz et~al.(2019)Schwartz, Dodge, Smith, and
  Etzioni}]{Schwartz:2019greenai}
Roy Schwartz, Jesse Dodge, Noah~A Smith, and Oren Etzioni. 2019.
\newblock Green ai.
\newblock \emph{arXiv preprint arXiv:1907.10597}.

\bibitem[{Shen et~al.(2020)Shen, Dong, Ye, Ma, Yao, Gholami, Mahoney, and
  Keutzer}]{shen2020q}
Sheng Shen, Zhen Dong, Jiayu Ye, Linjian Ma, Zhewei Yao, Amir Gholami,
  Michael~W Mahoney, and Kurt Keutzer. 2020.
\newblock Q-bert: Hessian based ultra low precision quantization of bert.
\newblock In \emph{Proceedings of the AAAI Conference on Artificial
  Intelligence}, volume~34, pages 8815--8821.

\bibitem[{Socher et~al.(2013)Socher, Perelygin, Wu, Chuang, Manning, Ng, and
  Potts}]{Socher:2013recursive}
Richard Socher, Alex Perelygin, Jean Wu, Jason Chuang, Christopher~D Manning,
  Andrew~Y Ng, and Christopher Potts. 2013.
\newblock Recursive deep models for semantic compositionality over a sentiment
  treebank.
\newblock In \emph{Proceedings of the 2013 conference on empirical methods in
  natural language processing}, pages 1631--1642.

\bibitem[{Strubell et~al.(2019)Strubell, Ganesh, and
  McCallum}]{Strubell:2019energy}
Emma Strubell, Ananya Ganesh, and Andrew McCallum. 2019.
\newblock Energy and policy considerations for deep learning in nlp.
\newblock \emph{arXiv preprint arXiv:1906.02243}.

\bibitem[{Sun et~al.(2020)Sun, Yu, Song, Liu, Yang, and
  Zhou}]{sun2020mobilebert}
Zhiqing Sun, Hongkun Yu, Xiaodan Song, Renjie Liu, Yiming Yang, and Denny Zhou.
  2020.
\newblock Mobilebert: a compact task-agnostic bert for resource-limited
  devices.
\newblock In \emph{Proceedings of the 58th Annual Meeting of the Association
  for Computational Linguistics}, pages 2158--2170.

\bibitem[{Tanaka et~al.(2019)Tanaka, Yamane, Héroux, Nakane, Kanazawa, Takeda,
  Numata, Nakano, and Hirose}]{Tanaka:2019review}
Gouhei Tanaka, Toshiyuki Yamane, Jean~Benoit Héroux, Ryosho Nakane, Naoki
  Kanazawa, Seiji Takeda, Hidetoshi Numata, Daiju Nakano, and Akira Hirose.
  2019.
\newblock Recent advances in physical reservoir computing: A review.
\newblock \emph{Neural Networks}, 115:100 -- 123.

\bibitem[{Tay et~al.(2020{\natexlab{a}})Tay, Bahri, Metzler, Juan, Zhao, and
  Zheng}]{Tay:2020synthesizer}
Yi~Tay, Dara Bahri, Donald Metzler, Da-Cheng Juan, Zhe Zhao, and Che Zheng.
  2020{\natexlab{a}}.
\newblock Synthesizer: Rethinking self-attention in transformer models.
\newblock \emph{arXiv preprint arXiv:2005.00743}.

\bibitem[{Tay et~al.(2020{\natexlab{b}})Tay, Dehghani, Bahri, and
  Metzler}]{Tay:2020transformersurvey}
Yi~Tay, Mostafa Dehghani, Dara Bahri, and Donald Metzler. 2020{\natexlab{b}}.
\newblock Efficient transformers: A survey.
\newblock \emph{arXiv preprint arXiv:2009.06732}.

\bibitem[{Tenney et~al.(2019)Tenney, Das, and Pavlick}]{Tenney:2019bert}
Ian Tenney, Dipanjan Das, and Ellie Pavlick. 2019.
\newblock Bert rediscovers the classical nlp pipeline.
\newblock \emph{arXiv preprint arXiv:1905.05950}.

\bibitem[{Ulyanov et~al.(2018)Ulyanov, Vedaldi, and
  Lempitsky}]{Ulyanov:2018deepimageprior}
Dmitry Ulyanov, Andrea Vedaldi, and Victor Lempitsky. 2018.
\newblock Deep image prior.
\newblock In \emph{Proceedings of the IEEE Conference on Computer Vision and
  Pattern Recognition}, pages 9446--9454.

\bibitem[{Vaswani et~al.(2017)Vaswani, Shazeer, Parmar, Uszkoreit, Jones,
  Gomez, Kaiser, and Polosukhin}]{Vaswani:2017attention}
Ashish Vaswani, Noam Shazeer, Niki Parmar, Jakob Uszkoreit, Llion Jones,
  Aidan~N Gomez, {\L}ukasz Kaiser, and Illia Polosukhin. 2017.
\newblock Attention is all you need.
\newblock In \emph{Advances in neural information processing systems}, pages
  5998--6008.

\bibitem[{Vinyals et~al.(2019)Vinyals, Babuschkin, Czarnecki, Mathieu, Dudzik,
  Chung, Choi, Powell, Ewalds, Georgiev, Oh, Horgan, Kroiss, Danihelka, Huang,
  Sifre, Cai, Agapiou, Jaderberg, Vezhnevets, Leblond, Pohlen, Dalibard,
  Budden, Sulsky, Molloy, Paine, Gulcehre, Wang, Pfaff, Wu, Ring, Yogatama,
  Wünsch, McKinney, Smith, Schaul, Lillicrap, Kavukcuoglu, Hassabis, Apps, and
  Silver}]{vinyals2019grandmaster}
Oriol Vinyals, Igor Babuschkin, Wojciech~M. Czarnecki, Michaël Mathieu, Andrew
  Dudzik, Junyoung Chung, David~H. Choi, Richard Powell, Timo Ewalds, Petko
  Georgiev, Junhyuk Oh, Dan Horgan, Manuel Kroiss, Ivo Danihelka, Aja Huang,
  Laurent Sifre, Trevor Cai, John~P. Agapiou, Max Jaderberg, Alexander~S.
  Vezhnevets, Rémi Leblond, Tobias Pohlen, Valentin Dalibard, David Budden,
  Yury Sulsky, James Molloy, Tom~L. Paine, Caglar Gulcehre, Ziyu Wang, Tobias
  Pfaff, Yuhuai Wu, Roman Ring, Dani Yogatama, Dario Wünsch, Katrina McKinney,
  Oliver Smith, Tom Schaul, Timothy Lillicrap, Koray Kavukcuoglu, Demis
  Hassabis, Chris Apps, and David Silver. 2019.
\newblock Grandmaster level in {StarCraft} {II} using multi-agent reinforcement
  learning.
\newblock \emph{Nature}, 575(7782):350--354.

\bibitem[{Voita and Titov(2020)}]{voita2020information}
Elena Voita and Ivan Titov. 2020.
\newblock Information-theoretic probing with minimum description length.
\newblock In \emph{Proceedings of the 2020 Conference on Empirical Methods in
  Natural Language Processing (EMNLP)}, pages 183--196.

\bibitem[{Wang et~al.(2020)Wang, Li, Khabsa, Fang, and Ma}]{Wang:2020linformer}
Sinong Wang, Belinda Li, Madian Khabsa, Han Fang, and Hao Ma. 2020.
\newblock Linformer: Self-attention with linear complexity.
\newblock \emph{arXiv preprint arXiv:2006.04768}.

\bibitem[{Wieting and Kiela(2019)}]{Wieting:2019notraining}
John Wieting and Douwe Kiela. 2019.
\newblock No training required: Exploring random encoders for sentence
  classification.
\newblock \emph{arXiv preprint arXiv:1901.10444}.

\bibitem[{Williams et~al.(2017)Williams, Nangia, and Bowman}]{Williams2017mnli}
Adina Williams, Nikita Nangia, and Samuel~R Bowman. 2017.
\newblock A broad-coverage challenge corpus for sentence understanding through
  inference.
\newblock \emph{arXiv preprint arXiv:1704.05426}.

\bibitem[{Williams(1992)}]{Williams1992reinforce}
Ronald~J Williams. 1992.
\newblock Simple statistical gradient-following algorithms for connectionist
  reinforcement learning.
\newblock \emph{Machine learning}, 8(3-4):229--256.

\bibitem[{Wu et~al.(2019)Wu, Fan, Baevski, Dauphin, and Auli}]{Wu2019pay}
Felix Wu, Angela Fan, Alexei Baevski, Yann~N Dauphin, and Michael Auli. 2019.
\newblock Pay less attention with lightweight and dynamic convolutions.
\newblock \emph{arXiv preprint arXiv:1901.10430}.

\bibitem[{Zaheer et~al.(2020)Zaheer, Guruganesh, Dubey, Ainslie, Alberti,
  Ontanon, Pham, Ravula, Wang, Yang et~al.}]{Zaheer2020big}
Manzil Zaheer, Guru Guruganesh, Avinava Dubey, Joshua Ainslie, Chris Alberti,
  Santiago Ontanon, Philip Pham, Anirudh Ravula, Qifan Wang, Li~Yang, et~al.
  2020.
\newblock Big bird: Transformers for longer sequences.
\newblock \emph{arXiv preprint arXiv:2007.14062}.

\bibitem[{Zhang et~al.(2019)Zhang, Bengio, and
  Singer}]{Zhang:2019alllayersequal}
Chiyuan Zhang, Samy Bengio, and Yoram Singer. 2019.
\newblock Are all layers created equal?
\newblock \emph{arXiv preprint arXiv:1902.01996}.

\bibitem[{Zhang and Bowman(2018)}]{Zhang:2018language}
Kelly Zhang and Samuel Bowman. 2018.
\newblock Language modeling teaches you more than translation does: Lessons
  learned through auxiliary syntactic task analysis.
\newblock In \emph{Proceedings of the 2018 EMNLP Workshop BlackboxNLP:
  Analyzing and Interpreting Neural Networks for NLP}.

\bibitem[{Zhou et~al.(2019)Zhou, Lan, Liu, and
  Yosinski}]{Zhou:2019deconstructing}
Hattie Zhou, Janice Lan, Rosanne Liu, and Jason Yosinski. 2019.
\newblock Deconstructing lottery tickets: Zeros, signs, and the supermask.
\newblock In \emph{Advances in Neural Information Processing Systems}, pages
  3597--3607.

\bibitem[{Zhu et~al.(2019)Zhu, Zhou, Wang, Luo, Li, Ni, and
  Xie}]{zhu-etal-2019-panlp}
Wei Zhu, Xiaofeng Zhou, Keqiang Wang, Xun Luo, Xiepeng Li, Yuan Ni, and Guotong
  Xie. 2019.
\newblock {PANLP} at {MEDIQA} 2019: Pre-trained language models, transfer
  learning and knowledge distillation.
\newblock In \emph{Proceedings of the 18th BioNLP Workshop and Shared Task},
  pages 380--388, Florence, Italy. Association for Computational Linguistics.

\end{thebibliography}
